\RequirePackage{rotating}
\documentclass[acmsmall, screen]{acmart} 

\acmJournal{CSUR}

\usepackage{booktabs}
\usepackage{subcaption}
\usepackage[utf8]{inputenc}
\usepackage[T1]{fontenc}
\usepackage{microtype}
\usepackage{graphicx}
\usepackage{color,xcolor}
\usepackage{rotating}
\usepackage{soul}
\usepackage[shortlabels]{enumitem}
\usepackage[export]{adjustbox}
\let\subparagraph\relax

\usepackage{etoolbox}
\makeatletter
\patchcmd{\ttlh@hang}{\parindent\z@}{\parindent\z@\leavevmode}{}{}
\patchcmd{\ttlh@hang}{\noindent}{}{}{}
\makeatother

\definecolor{dkgreen}{rgb}{0,0.6,0}
\definecolor{gray}{rgb}{0.5,0.5,0.5}
\definecolor{mauve}{rgb}{0.58,0,0.82}
\definecolor{red}{rgb}{0.9,0,0}
\definecolor{jcolor}{rgb}{0.6,0,0.5}
\definecolor{darktangerine}{rgb}{0.7, 0.4, 0.0}
\definecolor{ltblue}{RGB}{30,150,255}

\newcommand{\mnist}{\textsc{MNIST}\xspace}

\newcommand{\cifarten}{\textsc{CIFAR-10}\xspace}
\newcommand{\cifarhundred}{\textsc{CIFAR-100}\xspace}
\newcommand{\svhn}{\textsc{SVHN}\xspace}
\newcommand{\tinyset}{\textsc{Tiny Images Dataset}\xspace}

\newcommand{\halfmoon}{\textsc{Halfmoon}\xspace}
\newcommand{\mnistv}{\textsc{MNIST 1v7}\xspace}
\newcommand{\abalone}{\textsc{abalone}\xspace}

\usepackage[normalem]{ulem} % for strikethrough \sout, [normalem] ensures that \emph is NOT changed to underlined

\usepackage{filecontents,lipsum}  % http://ctan.org/pkg/lipsum
\usepackage{longtable}  %%% It is not working in a two-column environment. Don't know what to do. Stupid package!!!
\usepackage{balance}
\usepackage{boxedminipage}
\usepackage[linesnumbered]{algorithm2e}
\usepackage{url}
\usepackage{wrapfig}
\usepackage{rotating}
\usepackage{multicol, multirow, makecell}% http://ctan.org/pkg/multirow
\usepackage{tabularx, tabulary}
\usepackage{hhline}     % http://ctan.org/pkg/hhline
\usepackage{amsmath}
\usepackage{amsfonts}
\usepackage{xspace}
\usepackage{array}

\usepackage{courier}
\usepackage{ragged2e}   % for '\RaggedRight' macro (allows hyphenation)
\usepackage{newfloat}
\DeclareFloatingEnvironment[name={Algorithm}]{algfigure}

\usepackage{pifont}

\renewcommand{\checkmark}{\ding{51}}
\newcommand{\checkmarkwithcross}{\textcolor{red}{\ding{51}}\textsuperscript{\textcolor{red}{\kern-0.5em\tiny\ding{55}}}}

\usepackage{tikz}
\usetikzlibrary{shapes,arrows}
\usetikzlibrary{backgrounds,positioning}

\newcommand*\roundrect[1]{\tikz[baseline=(char.base)]{
            \node[shape=rectangle,draw,rounded corners=1.5pt, inner sep=1.5pt, minimum width=0.35cm, minimum height = 0.3cm] (char) {\footnotesize #1};}}

\DeclareMathAlphabet\mathbfcal{OMS}{cmsy}{b}{n}

\usepackage{graphicx}
\newcolumntype{Y}{>{\RaggedRight\arraybackslash}X} 

\newcolumntype{P}[1]{>{\raggedright\arraybackslash\hspace{0pt}}p{#1}}
\newcolumntype{C}[1]{>{\centering\arraybackslash\hspace{0pt}}p{#1}}

\usepackage{listings}
\usepackage{color}

\usepackage{hyperref}

\let\oldnl\nl% Store \nl in \oldnl
\newcommand{\nonl}{\renewcommand{\nl}{\let\nl\oldnl}}% Remove line number for one line

\usepackage{tcolorbox}

\usepackage{stfloats}
\usepackage{afterpage}

\begin{document}

\title{It Is All About Data: A Survey on the Effects of Data on Adversarial Robustness}

\author{Peiyu Xiong}
\affiliation{%
  \institution{University of British Columbia}
  \city{Vancouver}
  \country{Canada}
}
\email{gbxpeiyu@ece.ubc.ca}

\author{Michael Tegegn}
\affiliation{%
  \institution{University of British Columbia}
  \city{Vancouver}
  \country{Canada}
}
\email{mtegegn@ece.ubc.ca}

\author{Jaskeerat Singh Sarin}
\affiliation{%
  \institution{University of British Columbia}
  \city{Vancouver}
  \country{Canada}
}
\email{jsarin@student.ubc.ca}

\author{Shubhraneel Pal}
\affiliation{%
  \institution{Indian Institute of Technology}
  \city{Kharagpur}
  \country{India}
}
\email{shubhraneel@iitkgp.ac.in}

\author{Julia Rubin}
\affiliation{%
  \institution{University of British Columbia}
  \city{Vancouver}
  \country{Canada}
}
\email{mjulia@ece.ubc.ca}

\begin{abstract}
Adversarial examples are inputs to machine learning models that an attacker has intentionally designed to confuse the model
into making a mistake.
Such examples pose a serious threat to the applicability of machine-learning-based systems, especially in life- and safety-critical domains.
To address this problem, the area of adversarial robustness investigates mechanisms behind adversarial attacks and defenses against these attacks.
This survey reviews a particular subset of this literature that focuses on investigating properties of training data in the context of model robustness under evasion attacks.
It first summarizes the main properties of data leading to adversarial vulnerability. 
It then discusses guidelines and techniques for improving adversarial robustness by enhancing the data representation and learning procedures, 
as well as techniques for estimating robustness guarantees given particular data. 
Finally, it discusses gaps of knowledge and promising future research directions in this area.
\end{abstract}

\begin{CCSXML}
<ccs2012>
   <concept>
       <concept_id>10010147.10010257</concept_id>
       <concept_desc>Computing methodologies~Machine learning</concept_desc>
       <concept_significance>500</concept_significance>
       </concept>
   <concept>
       <concept_id>10002978.10003022</concept_id>
       <concept_desc>Security and privacy~Software and application security</concept_desc>
       <concept_significance>500</concept_significance>
       </concept>
 </ccs2012>
\end{CCSXML}

\ccsdesc[500]{Computing methodologies~Machine learning}
\ccsdesc[500]{Security and privacy~Software and application security}

\keywords{Machine Learning, Adversarial Robustness, Evasion Attack, Data Properties}

\maketitle

\section{Introduction}
\label{sec:introduction}

\begin{figure*}[th!]
    \vspace{-0.05in}
	\centering
	\includegraphics[width=0.45\textwidth]{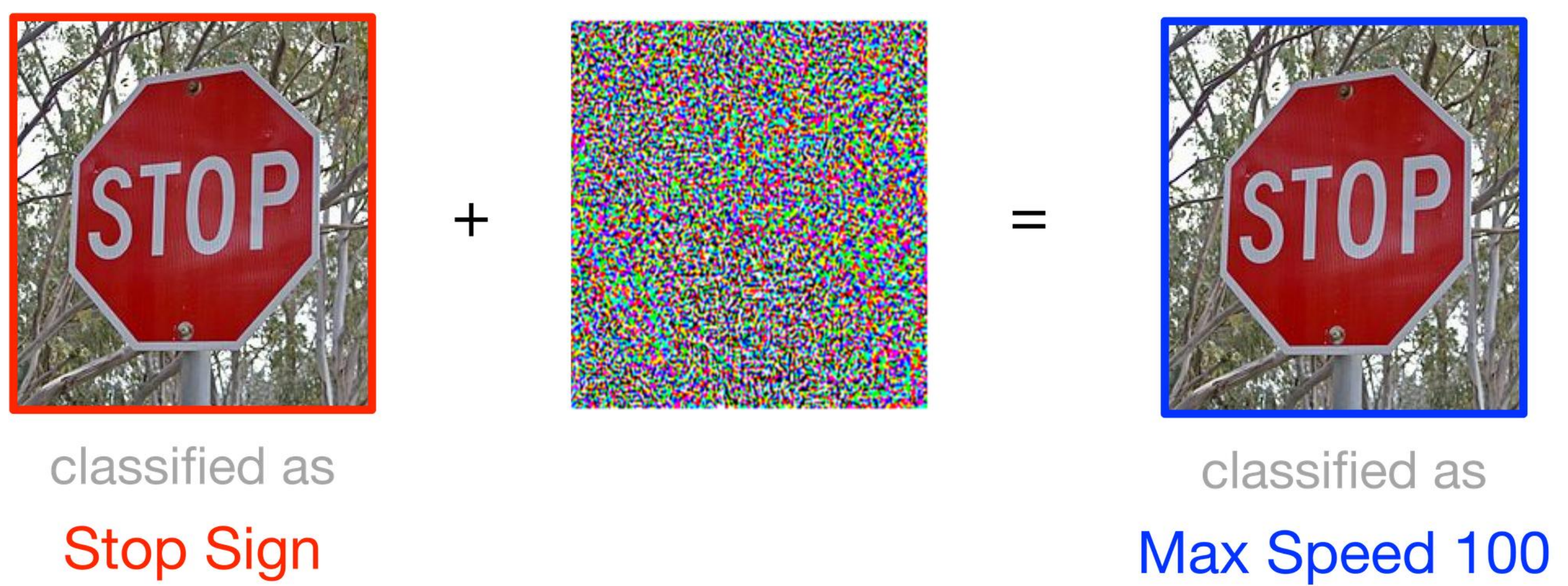}
	\vspace{-0.1in}
	\caption{Adversarial examples for traffic signs (picture by Chen and Wu~\cite{Chen:Wu:Altacognita:Blog:2019}).}
	\label{fig:stop_sign_side}
	\vspace{-0.2in}
\end{figure*}

Recent advances in Machine Learning (ML) led to the development of numerous accurate and scalable ML-based techniques,
which are increasingly used in industry and society.
Yet, concerns related to the safety and security of ML-based systems could substantially impede their widespread adoption, especially in the area
of safety-critical systems, such as autonomous cars.
Examples of fooling ML models into making wrong predictions by adding imperceptible-to-the-human noise to the input
are well known~\cite{OpenAI:report:2017}:
adversarial perturbation to a stop sign may cause a machine learning system to recognize it as a ``max speed'' sign instead,
which might lead to wrong and dangerous actions taken by an autonomous car~\cite{Eykholt:Evtimov:Fernandes:Li:Rahmati:Xiao:Prakash:Kohno:Song:CVPR:2018} (see Fig.~\ref{fig:stop_sign_side}).
Likewise, malicious software can be perturbed to bypass security models while still retaining its malicious behavior~\cite{Demontis:Melis:Biggio:Maiorca:Arp:Konrad:Rieck:Corona:Giacinto:Roli:TDSC:2019}.

ML models are susceptible to such scenarios, known as \emph{adversarial attacks} or \emph{adversarial examples}~\cite{Szegedy:Zaremba:Sutskever:Bruna:Erhan:Goodfellow:Fergus:ICLR:2014, Goodfellow:Shlens:Szegedy:ICLR:2014}.
To address this problem, recent literature investigates mechanisms behind adversarial attacks and proposes defenses against these attacks -- an area commonly referred to as \emph{adversarial robustness}.
The performance of ML models under adversarial attacks, known as \emph{robust accuracy} or \emph{robust generalization}, is often distinguished from the general model accuracy, known as \emph{standard accuracy} or \emph{standard generalization}.

Adversarial attacks that aim to decrease model accuracy can roughly be divided into \emph{evasion} and
\emph{poisoning attacks}~\cite{Biggio:Roli:PR:2018,Li:Li:Ye:Xu:CSUR:2021}.
The stop sign example above is, in fact, an evasion attack, where the attacker carefully modifies
the input to mislead the prediction~\cite{Szegedy:Zaremba:Sutskever:Bruna:Erhan:Goodfellow:Fergus:ICLR:2014,Madry:Makelov:Schmidt:Tspiras:Vladu:ICLR:2018, Kurakin:Goodfellow:Bengio:ICLR:2017, Carlini:Wagner:SP:2017}.
Instead of changing model inputs, poisoning attacks are carried out by injecting corrupt data into the training dataset, 
to compromise the integrity of the model~\cite{Goldblum:Tsipras:Xie:Chen:Schwarzchild:song:Madry:Li:Goldstein:TPAMI:2022,Shafahi:Huang:Najibi:Suciu:Studer:Dumitras:Goldstein:NeurIPS:2018, Tian:Cui:Liang:Yu:CSUR:2022}.
\emph{The focus of this survey is on evasion attacks}, as they are more common, accessible, and
more frequently discussed in the literature~\cite{Machado:Silva:Goldschmidt:CSUR:2021, Chakraborty:Alam:Dey:Chattopadhyay:Mukhopadhyay:ArXiv:2018, Viso:ai:blog:2022, Li:Li:Ye:Xu:CSUR:2021}.
As literature often uses the term \emph{adversarial attacks} to refer to evasion
attacks~\cite{Madry:Makelov:Schmidt:Tspiras:Vladu:ICLR:2018, Kurakin:Goodfellow:Bengio:ICLR:2017, Yuan:He:Zhu:Li:TNNLS:2019},
in this survey we use these two terms interchangeably.

Most techniques that study adversarial evasion attacks attribute adversarial vulnerability to various
aspects of the learning algorithm and/or properties of the data.
Numerous existing surveys on adversarial robustness focus on adversarial attacks and defenses~\cite{Zhou:Liu:Ye:Zhu:Zhou:Yu:CSUR:2022, Akhtar:Mian:Kardan:Shah:IEEEAccess:2021, Li:Li:Ye:Xu:CSUR:2021} and sources of adversarial vulnerability
related to learning algorithms~\cite{Machado:Silva:Goldschmidt:CSUR:2021, Serban:Poll:Visser:CSUR:2020}. 
Yet, to the best of our knowledge, there are no surveys that collect and organize literature 
focusing on the influence of \emph{data} on adversarial robustness.
Our work fills this gap. 
Specifically, {\bf we investigate 
(a) what properties of data influence model robustness and 
(b) how to select, represent, and use data to improve model robustness.}
To the best of our knowledge, this is the first survey to analyze adversarial robustness from the perspective of data properties.

To collect literature relevant to our survey, we used popular digital libraries and search engines, selecting papers that investigate the effect of data on ML adversarial robustness.
We identified more than 4,359 potentially relevant papers published in top scientific venues on Machine Learning, Computer Vision, Computational Linguistics, and Security.
We systematically inspected these papers, identifying 77 papers relevant to our survey. We further analyzed, categorized, and described the selected papers in this manuscript.

\vspace{0.02in}
\noindent
{\bf Main findings} (see Section~\ref{sec:results} for details). The results of our analysis show that producing accurate and robust models requires a larger \emph{number of samples} for training than achieving high accuracy alone.
The required number of samples to learn a robust model also depends on other properties of the data, such as dimensionality and the data distribution itself. Specifically, input data with higher \emph{dimensionality}, i.e., a larger number of features that represent the input dataset, requires a larger number of samples to produce a robust model.
This is consistent with other findings showing that high \emph{dimensionality} is undesirable for robustness.
Moreover, some data \emph{distributions} are inherently more robust than others, e.g., a Gaussian mixture distribution requires more samples to produce a robust model than needed by a Bernoulli mixture distribution.

Another aspect that affects robustness is the \emph{density} of data samples within classes, which measures how far apart samples are from each other.
Papers show that high class density correlates with high robust accuracy and that adversarial examples are commonly found in low-density regions of the data.
This is intuitive as low-density regions imply that there are not enough samples to accurately characterize the region.
A related property, \emph{concentration}, measures how fast the value of a function defined over a data region, e.g., \emph{error rate}, grows as the region expands.
This concept of concentration tightly corresponds to adversarial robustness if we consider the expansion of a data region as the effect of adversarial perturbation, i.e., perturbing samples in all directions causes the region defined by the original samples to expand.
In this case, high concentration implies that the error rate grows as one perturbs all points in a data region and, thus, datasets with high concentration are shown to be inevitably non-robust.
The \emph{separation} between classes of the underlying data distribution also affects robustness, with a large distance between different classes being desirable for adversarial robustness as an attacker would need to use a large perturbation to move samples from one class to another.

Yet another aspect that impacts robust accuracy is the presence of mislabeled samples in a dataset,
referred to as \emph{label noise}.
Furthermore, refining labels to reason about a larger number of classes, e.g., splitting a class ``animal'' into ``cat'' and ``dog'' may improve adversarial robustness as such labels allow learning more compact representations for samples that share stronger similarities.

A number of papers also identify \emph{domain-specific} properties that correlate with adversarial robustness.
For example, image frequency -- the rate of pixel value change -- affects robustness, and it is advisable to use a diverse frequency range in the training dataset to prevent any frequency biases which give rise to adversarial examples.

\vspace{0.02in}
\noindent
{\bf Practical Implications} (see Section~\ref{sec:discussion-practical} for details).
Our survey identifies a number of actionable guidelines and techniques that can be used to improve robustness.
These include data manipulation techniques, such as 
techniques for increasing the number of samples with real or generated data,
feature selection and dimensionality reduction techniques, and 
techniques for learning a latent data representation 
with desirable data properties, such as high density, high separation, and low dimensionality. 
While these techniques aim at changing properties of the underlying data, 
improving robustness can also be achieved by manipulating the learning procedures 
based on the properties of the training data, 
e.g., selecting particular models and/or configuring model parameters based on data dimensionality,
adjusting samples at inference time, and more.
A number of approaches also propose ways to estimate robustness guarantees for particular data, 
making it possible to reason about inherent robustness limitations in a practical setup.

\vspace{0.02in}
\noindent
{\bf Knowledge Gaps and Future Research Directions} (see Section~\ref{sec:discussion-future} for details).
Our literature review shows that, even though most works study data properties from a domain-agnostic perspective, they typically conduct an empirical evaluation on image datasets only.
This constrains the types of attacks and robustness measurements considered, 
and thus the findings may not generalize to other domains or types of datasets.
Furthermore, most works base their formal derivations on quite simple synthetic data models, such as uniform distributions, a mixture of Gaussian distributions, and a mixture of Bernoulli distributions,
which exhibit unrealistic assumptions compared to real datasets used in practice.
We also observed that while most papers only perform a univariate analysis on a specific data property,
most properties are hard to independently optimize, e.g., to decrease dimensionality without decreasing separation,
as decreasing the dimensionality implies that samples have fewer features to be differentiated from each other.
We also found that some properties, e.g., separation, do not have a standard way of measurement for concrete datasets.
We believe future work should look into these directions.

\vspace{0.02in}
\noindent
{\bf Contributions.} The main contributions of this survey are:
\vspace{-0.1in}
\begin{itemize}
\item A collection of literature on the effects of data on adversarial robustness.
\item A categorization and analysis of data properties that affect adversarial robustness.
\item An analysis of practical implications of the finding, 
knowledge gaps, and future \mbox{research directions}.
\end{itemize}

\section{Preliminaries}
\label{sec:background}
We now provide a brief overview of the main concepts related to machine learning,
adversarial robustness, and most commonly studied data distributions.
The goal of this section is to introduce terminology used in the rest of the survey rather than provide an extensive overview of the adversarial robustness research area.
For a more detailed overview, please refer to guides on statistics and machine learning~\cite{Shalev-Shwartz:Ben-David:2014, Giuseppe:2017, Hastie:Tibshirani:Friedman:2009} and adversarial robustness~\cite{Chen:Hsieh:2023, Biggio:Roli:PR:2018, Nicolae:Sinn:Tran:2018:ArXiv}. 
Please also see our online appendix for the list of symbols and acronyms used in this paper~\cite{appendix}.

\subsection{Machine Learning}
Machine learning refers to the automated detection of meaningful patterns in data~\cite{Shalev-Shwartz:Ben-David:2014}
and can be largely divided into supervised, unsupervised, and reinforcement learning.
In supervised learning, a learning model is provided with input-output pairs of data (a.k.a. labeled training data);
based on this data, the model aims to infer a function that maps the inputs to the outputs.
Supervised learning is typically associated with classification and regression problems, which use categorical and continuous labels, respectively. 
In classification, this number of possible labels for an input is also referred to as the number of \emph{classes}.
Datasets with only two classes are called \emph{binary datasets}, on which one can train a \emph{binary classifier}.

Unlike supervised learning, unsupervised learning algorithms are usually concerned with identifying patterns in unlabeled data, e.g.,
grouping similar samples together in the absence of labels (clustering) or
transforming data into a different representation (representation learning).
Reinforcement learning characterizes algorithms that learn from a series of rewards and punishments,
with the goal of maximizing the cumulative reward, e.g., to build robots that learn to take the best sequence of actions according to signals from the environment.

Variations, such as, semi-supervised learning (i.e., learning from partially labeled data) and
self-supervised learning (i.e., learning from labels extracted by the learner itself) have also been proposed for problems where acquiring labeled data may be challenging or expensive.

ML algorithms can also be divided into parametric and non-parametric.
Parametric algorithms have a predetermined, fixed number of parameters defined before the training starts.
For example, for Linear Support Vector Machines (SVMs), these parameters are the coefficients of all features of the
training data and the learned intercept.
For Deep Neural Networks (DNNs), the number of parameters is determined by the architecture of the network.
In non-parametric algorithms, the number of parameters is determined at training time and may vary depending on the number of training samples.
For example, the ``depth'' of Decision Trees can grow (beyond the size of the feature set)
when more decision points are needed to accurately separate training data.
Other commonly used non-parametric models include $k$-Nearest Neighbors ($k$-NN) and Kernel SVMs.

%\vspace{-0.2in}
\subsection{Adversarial Robustness}
Adversarial machine learning studies the arms race between adversarial attacks and defenses.
Attacks aim at degrading model performance while defenses propose algorithms to harden models against the attacks.
Adversarial attacks can be categorized into \emph{evasion} and \emph{poisoning}~\cite{Biggio:Roli:PR:2018,Li:Li:Ye:Xu:CSUR:2021}.
Evasion attacks aim to fool machine learning models by generating inputs that, despite no noticeable difference for a human,
will be incorrectly classified.
Such inputs, known as \emph{adversarial examples} and created by applying non-random perturbations to samples, are carefully designed to change models' predictions~\cite{Szegedy:Zaremba:Sutskever:Bruna:Erhan:Goodfellow:Fergus:ICLR:2014,
Madry:Makelov:Schmidt:Tspiras:Vladu:ICLR:2018, Kurakin:Goodfellow:Bengio:ICLR:2017, Carlini:Wagner:SP:2017}.
Instead, poisoning attacks tampers with model training data, in order to degrade model performance.
In this survey, we focus on evasion attacks; the terms adversarial and evasion attacks are often used interchangeably in the literature as this is the most popular and commonly studied type of attack.

The term \emph{robustness} for machine learning models is often used to refer to different concepts, such as,
stability to distribution shifts,
the ability to identify adversarial examples, and
the ability to make the correct predictions in the face of adversarial examples.
In this survey, we use the latter definition -- the ability to make the correct predictions in the face of adversarial examples.
This is a stronger notion of robustness than merely identifying adversarial examples, as the identification of an adversarial example does not guarantee its correct classification.
The phenomenon of making satisfactory model predictions in the face of adversarial examples is also often referred to as \emph{robust generalization}.
This is different from \emph{standard generalization}, a term used to describe
making satisfactory model predictions for normal, unseen samples.

\vspace{0.05in}
\noindent
{\bf Adversarial (Evasion) Attacks.}
Techniques for generating adversarial examples for evasion attacks can be broadly divided into three categories,
according to the type of information available to the attacker~\cite{Biggio:Roli:PR:2018}.
In \emph{white-box attacks}, the attacker is assumed to be able to leverage all available information about the training data, the model, and the training procedure.
In \emph{grey-box attacks}, the attacker is assumed to have only partial information about the model, such as, the source of training data.
Finally, the most conservative type of attacks are \emph{black-box attacks}, where the attacker has no information about the inner workings of the model except, possibly, for the prediction outcomes.

\emph {Gradient-based attacks} are commonly used in white-box settings. These attacks use the gradient of a differentiable function defined over model weights as a guide when crafting adversarial examples.
The most commonly used differentiable function is the loss function used by the model during training.
A gradient defines the direction of the maximal increase in the local value of a function.
Hence, by using the gradient of the loss function with respect to the input, one can adjust the input to get the maximal increase in the loss of the model, which ultimately leads to a bad prediction.
Fast Gradient Sign Method (FGSM)~\cite{Goodfellow:Shlens:Szegedy:ICLR:2015}, Basic Iterative Method (BIM)~\cite{Kurakin:Goodfellow:Bengio:ICLR:2017}, and Projected Gradient Descent (PGD)~\cite{Madry:Makelov:Schmidt:Tspiras:Vladu:ICLR:2018} are examples of attack algorithms that utilize the gradient of the loss function used for training.
Instead of the loss function, the FAB attack~\cite{Croce:Hein:ICML:2020:1} uses the gradient of a function defined by the difference of model outputs of the penultimate layer of a neural network -- a layer which outputs the probabilities that a given sample belongs to each of the available classes.
By defining the difference of outputs of the penultimate layer as the differentiable function, the FAB attack maximizes the difference in probabilities between the target class and other classes, to increase the chance of misclassification.

\emph{Non-gradient based attacks} are applicable for more diverse types of models that do not use a differentiable functions, e.g., decision trees.
Such attacks can also be used in black-box and grey-box settings, when gradient information is hidden from the attacker.
One example of non-gradient-based attacks is the \emph{mimicry attack}, which involves adding and removing features in the perturbed sample, e.g., based on their popularity in the target class~\cite{Demontis:Melis:Biggio:Maiorca:Arp:Konrad:Rieck:Corona:Giacinto:Roli:TDSC:2019}.

\vspace{0.05in}
\noindent
{\bf Adversarial Defenses.}
Defense mechanisms against adversarial attacks target various stages of the machine learning pipeline.
Specifically, defenses \emph{on raw data} focus on the training data itself, e.g.,
by selecting a subset of ``robust'' features~\cite{Ilyas:Santurkar:Tsipras:Engstrom:Tran:Madry:NeurIPS:2019} or
using representation learning to transform features into a different representation, making sure
a model trained on the new representation is inherently more robust~\cite{Yang:Guo:Wang:Xu:AAAI:2021}.

Defenses \emph{during training} alter the standard training procedure to improve model robustness.
The most common such technique is adversarial training~\cite{Goodfellow:Shlens:Szegedy:ICLR:2015}, which involves continually augmenting the training data with adversarial examples generated by an attack algorithm.
By retraining the model while adding correctly labeled malicious samples to the training dataset, the model learns to capture persistent patterns and becomes more robust against these attacks.
Another common method is \emph{regularization}, where model parameters are constrained so that very small perturbations have little effect on the prediction outcome~\cite{Gouk:Frank:Pfahringer:J.Cree:ML:2021}.

Defenses \emph{during inference} focus on making existing models more robust when the model is being used on new samples.
For example, randomized smoothing~\cite{Cohen:Rosenfeld:Kolter:ICML:2019} involves creating multiple noisy
instances of a sample and aggregating the model's predictions during inference.
Given that adversarial examples are typically close to genuine samples, averaging the results from close neighbors of an input can potentially reduce the chances of the model being misled.
In addition, different variations of ensemble models~-- using multiple models and aggregating their output~--
have been shown to increase the robustness to adversarial attacks~\cite{Pang:Xu:Du:Chen:Zhu:ICML:2019}.

\vspace{0.05in}
\noindent
{\bf Measures of Robustness.}
The strength of an adversary is mostly measured by the size of the perturbation required to create an adversarial example.
That is, adversaries that introduce more perturbations, e.g., change a larger portion of pixel values in the image, are considered to be stronger.
A typical way of measuring the perturbation size, especially in the image domain,
is by using the $L_p$ distance metric, where $p$ can be a whole number or $\infty$.
Specifically,
$L_0$ counts the total number of changed features, regardless of the changes to individual features.
$L_1$ is the Manhattan distance, i.e., the sum of absolute values representing a change in each feature.
$L_2$ measures the Euclidean distance between the feature values of the original and perturbed samples.
The $L_\infty$ metric measures the largest change in any of the features
(while disregarding changes in all other features).

There are two ways to utilize these distance metrics to evaluate the robustness of a model: error-rate-based and radius-based.
The first calculates a pool of adversarial samples generated from a set of real samples with a fixed allowable perturbation size~\cite{Madry:Makelov:Schmidt:Tspiras:Vladu:ICLR:2018}.
The robustness is then defined as the error rate of the model on these adversarial samples.
A related concept, \emph{adversarial risk}, is also defined in a similar manner:
the probability of finding, within a certain predefined distance, an adversarial example for a given real sample.
Radius-based evaluation measures the smallest distance required to generate an adversarial sample from a given real sample~\cite{Szegedy:Zaremba:Sutskever:Bruna:Erhan:Goodfellow:Fergus:ICLR:2014}.
This way is especially useful in robustness certification, which involves learning a classifier that outputs a prediction along with a certified radius within which the prediction is guaranteed to be consistent~\cite{Lee:Yuan:Chang:Jaakkola:NeurIPS:2019}.

\subsection{Data Distributions}
\label{sec:background_distribution}
Numerous works study properties of particular data distributions.
The \emph{uniform distribution} defines a probability distribution in which every possible data point is equally likely.
This implies that for a continuous random variable in the interval $[a, b]$, the probability of seeing a sample from the interval is $\frac{1}{b-a}$ and the probability of seeing a sample from outside of the interval is $0$.
In the discrete case with $n$ possible values, the uniform distribution assigns a probability of $\frac{1}{n}$ to each value.
The \emph{Bernoulli distribution} defines a discrete probability distribution of a random variable with two allowable values, 0 and 1.
Such a random variable takes the value of 1 with probability $p$ and the value of 0 with probability $1-p$.
The \emph{Gaussian (normal) distribution} defines a continuous probability distribution that assigns a probability with its peak at the center of the distribution and decreasing symmetrically outwards.
For a Gaussian distribution, $\mu$ denotes the mean or center of the distribution, and $\sigma^2$ denotes the variance or the spread of the distribution.
Since the mean and variance fully characterize a Gaussian distribution, it is also commonly denoted as $\mathcal{N}(\mu,\sigma^2)$.

\begin{wrapfigure}{r}{0.25\textwidth}
\vspace{-0.15in}
	\centering
	\includegraphics[width=0.2\textwidth]{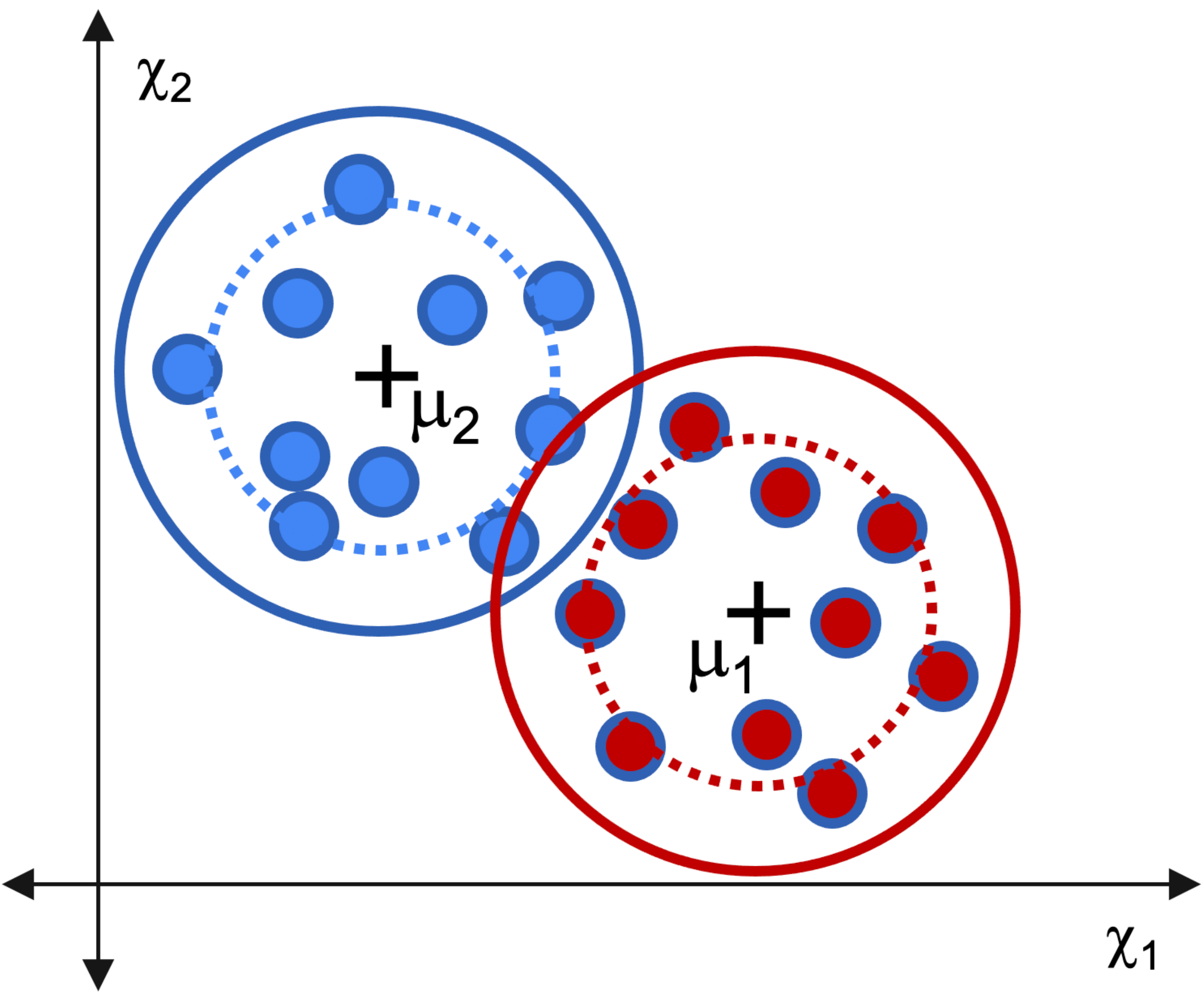}
	\vspace{-0.1in}
	\caption{A two-dimensional Gaussian mixture data.}
	\label{fig:gaussian_mixture}
\vspace{-0.2in}
\end{wrapfigure}
One can also imagine a distribution made up of a \emph{mixture} of multiple distributions.
For example, Fig.~\ref{fig:gaussian_mixture} shows a distribution made up of two Gaussians: one centered at $\mu_1$ and another~-- at $\mu_2$.
This mixture also contains labels associated with each independent Gaussian, shown by the two clusters in the figure.
Furthermore, these two clusters have the same variance,
i.e., the same spread of the  distribution surrounding the center of the class.
While the means of the two classes are separated, the distributions intersect with each other.

\section{Methodology}
\label{sec:methodology}
This section describes our methodology for identifying and categorization relevant papers.

\subsection{Paper Collection}
Papers for this survey were collected in May 2023.
We used the search query schematically described below,
which was designed to identify papers in the area of Machine Learning adversarial robustness that
discuss properties of the underlying data.
We expanded each of these conceptual terms with possible synonyms and specific wording, making sure our query is as comprehensive as possible.

\begin{quote}
  \textbf{Search Query} :=  Machine Learning + Adversarial Robustness + Data + Property  \\
  \textbf{Machine Learning} := classif | ``machine learning'' | ``deep learning'' | ``neural network'' \\
  \textbf{Adversarial Robustness} := ``adversarial robustness''  | ``adversarial vulnerability'' | ``adversarial attack'' | ``adversarial perturbation'' | ``adversarial defense'' | ``evasion attacks''  \\
    \textbf{Data} := data | sample | input\\
  \textbf{Property} := propert | qualit | distribution | characteristic
\end{quote}

In our schematic query representation, the ``+'' and ``|'' signs indicate the AND and OR operators, respectively,
the phases in quotes are matched in full, and
each word is matched with its suffixed versions, e.g., `classif' is matched with both `classifier' and `classification'.
When performing the search, we adapted this schematic query to the requirements and capabilities of each search engine that we used.

An initial search using the query in Google Scholar identified more than 30,000 matches.
To keep the scope of the survey manageable, we thus limited our search to publications from the main track of top-tier conferences and journals in the areas of Machine Learning (ML), Computer Vision (CV), Computational Linguistics (CL), and Security (SEC).
Specifically, we selected all A* conferences from these areas using the most recent, 2021, CORE ranking~\cite{CoreRanking:website:2022};
the top 15 journals according to the Journal Citation Reports (JCR)~\cite{JournalRanking:website:2022}
in the area of Artificial Intelligence (AI),
which includes Machine Learning, Computer Vision, and Computation Linguistics,
and the top five journals in the area of Information System (IS), which includes Security.
Additionally, we included the ACM Computing Surveys Journal~\cite{ACM:Computing:Surveys:Journal} to collect surveys related to our topic.
The first three columns in Table~\ref{tbl:venues} show the final list of publication venues that we selected.

We further identified digital libraries and search engines that host proceedings of our selected venues.
These are shown in the next five columns of Table~\ref{tbl:venues}, for each venue individually.
As some venues are only partially indexed by the digital libraries, 
i.e., the libraries only include proceedings from particular years, we augmented our search for papers in
these venues with a secondary search using Google Scholar.
More specifically, we used Google Scholar's \emph{site} and \emph{source} filtering constraint
to limit our search only to the target venues of interest,
as described in our online appendix~\cite{appendix}.
We also used Google Scholar to search the ArXiv repository and filtered the results using 
\emph{publication venue} information provided by the Semantic Scholar API~\cite{SemanticScholar:api:2022}.

\begin{table}[t!]
\vspace{-0.1in}
\caption{Considered publication venues}
\vspace{-0.1in}
\scalebox{0.61}{
\begin{tabular}{|ccl|ccccc|l|l|}
\hline
\multicolumn{1}{|c|}{}                                                                       & \multicolumn{1}{c|}{}                       & \multicolumn{1}{c|}{}                                                  & \multicolumn{5}{c|}{\textbf{Search Target}}                                                                                                                                                                                                    & \multicolumn{1}{c|}{}                                                                       & \multicolumn{1}{c|}{}                                                                     \\ \cline{4-8}
\multicolumn{1}{|c|}{\multirow{-2}{*}{\begin{tabular}[c]{@{}c@{}}\textbf{Venue}\\ \textbf{Type}\end{tabular}}} & \multicolumn{1}{c|}{\multirow{-2}{*}{\textbf{Area}}} & \multicolumn{1}{c|}{\multirow{-2}{*}{\textbf{Venue Name}}}                      & \multicolumn{1}{c|}{\textbf{ACM}}        & \multicolumn{1}{c|}{\textbf{IEEE}}                          & \multicolumn{1}{c|}{\textbf{Springer}}                      & \multicolumn{1}{c|}{\textbf{Scopus}}     & \begin{tabular}[c]{@{}c@{}}\textbf{Google}\\ \textbf{Scholar}\end{tabular} & \multicolumn{1}{c|}{\multirow{-2}{*}{\begin{tabular}[c]{@{}c@{}}\textbf{Total}\\ \textbf{Hits}\end{tabular}}} & \multicolumn{1}{c|}{\multirow{-2}{*}{\begin{tabular}[c]{@{}c@{}}\textbf{Rele-}\\ \textbf{vant}\end{tabular}}} \\ \hline \hline
\multicolumn{1}{|c|}{}                                                                       & \multicolumn{1}{c|}{}                       & AAAI Conference on Artificial Intelligence (AAAI)                      & \multicolumn{1}{c|}{}           & \multicolumn{1}{c|}{}                              & \multicolumn{1}{c|}{}                              & \multicolumn{1}{c|}{\checkmark} & \checkmark                                               & 320                                                                                         & 2                                                                                         \\ \cline{3-10} 
\multicolumn{1}{|c|}{}                                                                       & \multicolumn{1}{c|}{}                       & ACM International Conference on Web Search and Data Mining (WSDM)      & \multicolumn{1}{c|}{\checkmark} & \multicolumn{1}{c|}{}                              & \multicolumn{1}{c|}{}                              & \multicolumn{1}{c|}{}           &                                                          & 41                                                                                          & 0                                                                                         \\ \cline{3-10} 
\multicolumn{1}{|c|}{}                                                                       & \multicolumn{1}{c|}{}                       & ACM SIGKDD Conference On Knowledge and Data Mining (KDD)               & \multicolumn{1}{c|}{\checkmark} & \multicolumn{1}{c|}{}                              & \multicolumn{1}{c|}{}                              & \multicolumn{1}{c|}{}           &                                                          & 36                                                                                          & 0                                                                                         \\ \cline{3-10} 
\multicolumn{1}{|c|}{}                                                                       & \multicolumn{1}{c|}{}                       & Conference on Neural Information Processing Systems (NeurIPS)          & \multicolumn{1}{c|}{}           & \multicolumn{1}{c|}{}                              & \multicolumn{1}{c|}{}                              & \multicolumn{1}{c|}{\checkmark} & \checkmark                                               & 813                                                                                         & 25                                                                                        \\ \cline{3-10} 
\multicolumn{1}{|c|}{}                                                                       & \multicolumn{1}{c|}{}                       & IEEE International Conference on Data Engineering (ICDE)               & \multicolumn{1}{c|}{}           & \multicolumn{1}{c|}{\checkmark}                    & \multicolumn{1}{c|}{}                              & \multicolumn{1}{c|}{}           &                                                          & 10                                                                                           & 0                                                                                         \\ \cline{3-10} 
\multicolumn{1}{|c|}{}                                                                       & \multicolumn{1}{c|}{}                       & IEEE International Conference on Data Mining (ICDM)                    & \multicolumn{1}{c|}{}           & \multicolumn{1}{c|}{\checkmark}                    & \multicolumn{1}{c|}{}                              & \multicolumn{1}{c|}{}           &                                                          & 48                                                                                          & 1                                                                                         \\ \cline{3-10} 
\multicolumn{1}{|c|}{}                                                                       & \multicolumn{1}{c|}{}                       & International Conference on Learning Representations (ICLR)            & \multicolumn{1}{c|}{}           & \multicolumn{1}{c|}{}                              & \multicolumn{1}{c|}{}                              & \multicolumn{1}{c|}{\checkmark} & \checkmark                                               & 347                                                                                         & 9
 \\ \cline{3-10} 
\multicolumn{1}{|c|}{}                                                                       & \multicolumn{1}{c|}{}                       & International Conference on Learning Theory (COLT)                     & \multicolumn{1}{c|}{}           & \multicolumn{1}{c|}{}                              & \multicolumn{1}{c|}{}                              & \multicolumn{1}{c|}{\checkmark} & \checkmark                                               & 19                                                                                          & 1                                                                                         \\ \cline{3-10} 
\multicolumn{1}{|c|}{}                                                                       & \multicolumn{1}{c|}{}                       & International Conference on Machine Learning (ICML)                    & \multicolumn{1}{c|}{}           & \multicolumn{1}{c|}{}                              & \multicolumn{1}{c|}{}                              & \multicolumn{1}{c|}{\checkmark} & \checkmark                                               & 444                                                                                         & 13                                                                                        \\ \cline{3-10} 
\multicolumn{1}{|c|}{}                                                                       & \multicolumn{1}{c|}{\multirow{-10}{*}{ML}}  & International Joint Conference on Artificial Intelligence (IJCAI)      & \multicolumn{1}{c|}{}           & \multicolumn{1}{c|}{}                              & \multicolumn{1}{c|}{}                              & \multicolumn{1}{c|}{\checkmark} & \checkmark                                               & 115                                                                                          & 1                                                                                         \\ \cline{2-10} 
\multicolumn{1}{|c|}{}                                                                       & \multicolumn{1}{c|}{}                       & European Conference on Computer Vision (ECCV)                          & \multicolumn{1}{c|}{}           & \multicolumn{1}{c|}{}                              & \multicolumn{1}{c|}{\checkmark}                    & \multicolumn{1}{c|}{}           &                                                          & 155                                                                                          & 3                                                                                         \\ \cline{3-10} 
\multicolumn{1}{|c|}{}                                                                       & \multicolumn{1}{c|}{}                       & IEEE / CVF Computer Vision and Pattern Recognition (CVPR)              & \multicolumn{1}{c|}{}           & \multicolumn{1}{c|}{\checkmark}                    & \multicolumn{1}{c|}{}                              & \multicolumn{1}{c|}{}           &                                                          & 339                                                                                         & 3                                                                                         \\ \cline{3-10} 
\multicolumn{1}{|c|}{}                                                                       & \multicolumn{1}{c|}{\multirow{-3}{*}{CV}}   & IEEE International Conference on Computer Vision (ICCV)                & \multicolumn{1}{c|}{}           & \multicolumn{1}{c|}{\checkmark}                    & \multicolumn{1}{c|}{}                              & \multicolumn{1}{c|}{}           &                                                          & 157                                                                                         & 2                                                                                         \\ \cline{2-10} 
\multicolumn{1}{|c|}{}                                                                       & \multicolumn{1}{c|}{CL}                     & Annual Meeting of the Association for Computational Linguistics (ACL)  & \multicolumn{1}{c|}{}           & \multicolumn{1}{c|}{}                              & \multicolumn{1}{c|}{}                              & \multicolumn{1}{c|}{\checkmark} & \checkmark                                               & 87                                                                                          & 0                                                                                         \\ \cline{2-10} 
\multicolumn{1}{|c|}{}                                                                       & \multicolumn{1}{c|}{}                       & ACM Conference on Computer and Communications Security (CCS)           & \multicolumn{1}{c|}{\checkmark} & \multicolumn{1}{c|}{}                              & \multicolumn{1}{c|}{}                              & \multicolumn{1}{c|}{}           &                                                          & 79                                                                                          & 0                                                                                         \\ \cline{3-10} 
\multicolumn{1}{|c|}{}                                                                       & \multicolumn{1}{c|}{}                       & IEEE Symposium on Security and Privacy (S\&P)                         & \multicolumn{1}{c|}{}           & \multicolumn{1}{c|}{\checkmark}                    & \multicolumn{1}{c|}{}                              & \multicolumn{1}{c|}{}           &                                                          & 33                                                                                          & 0                                                                                         \\ \cline{3-10} 
\multicolumn{1}{|c|}{}                                                                       & \multicolumn{1}{c|}{}                       & Network and Distributed System Security Symposium (NDSS)               & \multicolumn{1}{c|}{}           & \multicolumn{1}{c|}{}                              & \multicolumn{1}{c|}{}                              & \multicolumn{1}{c|}{\checkmark} & \checkmark                                               & 34                                                                                          & 0                                                                                         \\ \cline{3-10} 
\multicolumn{1}{|c|}{\multirow{-18}{*}{\begin{turn}{90}Conference\end{turn}}}                                                & \multicolumn{1}{c|}{\multirow{-4}{*}{SEC}}   & USENIX Security Symposium                                              & \multicolumn{1}{c|}{}           & \multicolumn{1}{c|}{}                              & \multicolumn{1}{c|}{}                              & \multicolumn{1}{c|}{\checkmark} & \checkmark                                               & 133                                                                                          & 0                                                                                         \\ \hline
\multicolumn{1}{|c|}{}                                                                       & \multicolumn{1}{c|}{}                       & Artificial Intelligence Journal                                        & \multicolumn{1}{c|}{}           & \multicolumn{1}{c|}{}                              & \multicolumn{1}{c|}{}                              & \multicolumn{1}{c|}{\checkmark} & \checkmark                                               & 13                                                                                           & 0                                                                                         \\ \cline{3-10} 
\multicolumn{1}{|c|}{}                                                                       & \multicolumn{1}{c|}{}                       & Computational Linguistics Journal (CL)                                 & \multicolumn{1}{c|}{}           & \multicolumn{1}{c|}{}                              & \multicolumn{1}{c|}{}                              & \multicolumn{1}{c|}{\checkmark} & \checkmark                                               & 5                                                                                           & 0                                                                                         \\ \cline{3-10} 
\multicolumn{1}{|c|}{}                                                                       & \multicolumn{1}{c|}{}                       & Data Mining and Knowledge Discovery                                                   & \multicolumn{1}{c|}{}           & \multicolumn{1}{c|}{}                              & \multicolumn{1}{c|}{\checkmark}                              & \multicolumn{1}{c|}{} &                                               & 12 & 0                                                                                         \\ \cline{3-10} 
\multicolumn{1}{|c|}{}                                                                       & \multicolumn{1}{c|}{}                       & IEEE Transactions on Knowledge and Data Engineering (TKDE)             & \multicolumn{1}{c|}{}           & \multicolumn{1}{c|}{\checkmark}                    & \multicolumn{1}{c|}{}                              & \multicolumn{1}{c|}{}           &                                                          & 48                                                                                          & 0                                                                                         \\ \cline{3-10} 
\multicolumn{1}{|c|}{}                                                                       & \multicolumn{1}{c|}{}                       & IEEE Transactions on Neural Networks and Learning Systems (TNNLS)      & \multicolumn{1}{c|}{}           & \multicolumn{1}{c|}{\checkmark}                    & \multicolumn{1}{c|}{}                              & \multicolumn{1}{c|}{}           &                                                          & 79                                                                                          & 0                                                                                         \\ \cline{3-10} 
\multicolumn{1}{|c|}{}                                                                       & \multicolumn{1}{c|}{}                       & IEEE Transactions on Pattern Analysis and Machine Intelligence (TPAMI) & \multicolumn{1}{c|}{}           & \multicolumn{1}{c|}{\checkmark}                    & \multicolumn{1}{c|}{}                              & \multicolumn{1}{c|}{}           &                                                          & 138                                                                                          & 1                                                                                         \\ \cline{3-10} 
\multicolumn{1}{|c|}{}                                                                       & \multicolumn{1}{c|}{}                       & IEEE Transactions on Image Processing (TIP)                                                  & \multicolumn{1}{c|}{}           & \multicolumn{1}{c|}{\checkmark}                              & \multicolumn{1}{c|}{}                              & \multicolumn{1}{c|}{} &                                                &  45                                                                                         & 2                                                                                         \\ \cline{3-10} 
\multicolumn{1}{|c|}{}                                                                       & \multicolumn{1}{c|}{}                       & International Journal of Computer Vision (IJCV)                        & \multicolumn{1}{c|}{}           & \multicolumn{1}{c|}{}                              & \multicolumn{1}{c|}{\checkmark}                    & \multicolumn{1}{c|}{}           &                                                          & 33                                                                                          & 0                                                                                         \\ \cline{3-10} 
\multicolumn{1}{|c|}{}                                                                       & \multicolumn{1}{c|}{}                       & Journal of Machine Learning Research (JMLR)                            & \multicolumn{1}{c|}{}           & \multicolumn{1}{c|}{}                              & \multicolumn{1}{c|}{}                              & \multicolumn{1}{c|}{\checkmark} & \checkmark                                               & 42                                                                                          & 2                                                                                         \\ \cline{3-10} 
\multicolumn{1}{|c|}{}                                                                       & \multicolumn{1}{c|}{}                       & Knowledge-Based Systems (KBS)                                          & \multicolumn{1}{c|}{}           & \multicolumn{1}{c|}{}                              & \multicolumn{1}{c|}{}                              & \multicolumn{1}{c|}{\checkmark} & \checkmark                                               & 41                                                                                          & 0                                                                                         \\ \cline{3-10} 
\multicolumn{1}{|c|}{}                                                                       & \multicolumn{1}{c|}{}                       & Machine Learning & \multicolumn{1}{c|}{}           & \multicolumn{1}{c|}{}                              & \multicolumn{1}{c|}{\checkmark}                              & \multicolumn{1}{c|}{} &                                                & 44                                                                                          & 2                                                                                         \\ \cline{3-10} 
\multicolumn{1}{|c|}{}                                                                       & \multicolumn{1}{c|}{}                       & Medical Image Analysis & \multicolumn{1}{c|}{}           & \multicolumn{1}{c|}{}                              & \multicolumn{1}{c|}{}                              & \multicolumn{1}{c|}{\checkmark} & \checkmark                                               & 14                                                                                          & 0                                                                                         \\ \cline{3-10} 
\multicolumn{1}{|c|}{}                                                                       & \multicolumn{1}{c|}{}                       & Nature Machine Intelligence & \multicolumn{1}{c|}{}           & \multicolumn{1}{c|}{}                              & \multicolumn{1}{c|}{\checkmark}                              & \multicolumn{1}{c|}{} &                                                & 10                                                                                           & 0                                                                                         \\ \cline{3-10} 
\multicolumn{1}{|c|}{}                                                                       & \multicolumn{1}{c|}{}                       & Neural Networks                                                        & \multicolumn{1}{c|}{}           & \multicolumn{1}{c|}{}                              & \multicolumn{1}{c|}{}                              & \multicolumn{1}{c|}{\checkmark} & \checkmark                                               & 83                                                                                          & 0                                                                                         \\ \cline{3-10} 
\multicolumn{1}{|c|}{}                                                                       & \multicolumn{1}{c|}{\multirow{-15}{*}{AI}}                      & Pattern Recognition (PR)                                                     & \multicolumn{1}{c|}{}           & \multicolumn{1}{c|}{}                              & \multicolumn{1}{c|}{}                              & \multicolumn{1}{c|}{\checkmark} & \checkmark                                               & 117                                                                                                                                                                                   & 3                                                                                         \\ \cline{2-10} 

\multicolumn{1}{|c|}{}                                                                       & \multicolumn{1}{c|}{}                       & Computer \& Security                                                    & \multicolumn{1}{c|}{}           & \multicolumn{1}{c|}{}                              & \multicolumn{1}{c|}{}                              & \multicolumn{1}{c|}{\checkmark} & \checkmark                                               & 82                                                                                          & 0                                                                                         \\ \cline{3-10} 
\multicolumn{1}{|c|}{}                                                                       & \multicolumn{1}{c|}{}                       & IEEE Security \& Privacy                                                & \multicolumn{1}{c|}{}           & \multicolumn{1}{c|}{\checkmark}                    & \multicolumn{1}{c|}{}                              & \multicolumn{1}{c|}{}           &                                                          & 17                                                                                          & 0                                                                                         \\ \cline{3-10} 
\multicolumn{1}{|c|}{}                                                                       & \multicolumn{1}{c|}{}                       & IEEE Transactions on Dependable and Secure Computing (TDSC)            & \multicolumn{1}{c|}{}           & \multicolumn{1}{c|}{\checkmark}                    & \multicolumn{1}{c|}{}                              & \multicolumn{1}{c|}{}           &                                                          & 103                                                                                          & 0                                                                                         \\ \cline{3-10} 
\multicolumn{1}{|c|}{}                                                                       & \multicolumn{1}{c|}{}                       & IEEE Transactions on Information and Forensic Science (TIFS)           & \multicolumn{1}{c|}{}           & \multicolumn{1}{c|}{\checkmark}                    & \multicolumn{1}{c|}{}                              & \multicolumn{1}{c|}{}           &                                                          & 114                                                                                          & 1                                                                                         \\ \cline{3-10} 
\multicolumn{1}{|c|}{\multirow{-16}{*}{\begin{turn}{90}Journal\end{turn}}}                                                & \multicolumn{1}{c|}{\multirow{-5}{*}{IS}}   & Journal of Information Security and Applications (JISA)                & \multicolumn{1}{c|}{}           & \multicolumn{1}{c|}{} & \multicolumn{1}{c|}{} & \multicolumn{1}{c|}{\checkmark} & \checkmark                                               & 23                                                                                          & 0                                                                                         \\ \cline{2-10} 
\multicolumn{1}{|c|}{}                                                                       & \multicolumn{1}{c|}{Surv.}                       & ACM Computing Surveys (CSUR)          & \multicolumn{1}{c|}{\checkmark} & \multicolumn{1}{c|}{}                           & \multicolumn{1}{c|}{}                              & \multicolumn{1}{c|}{}           &                                                          & 86                                                                                         & 0                                                                                      \\ \hline
%\multicolumn{1}{|c|}{}                                                                                                               & ACM Computing Survey                                                   & \multicolumn{1}{c|}{\checkmark} & \multicolumn{1}{c|}{}                              & \multicolumn{1}{c|}{}                              & \multicolumn{1}{c|}{}           &                                                          & 59                                                                                          & 0                                                                                         \\ \hline
\multicolumn{3}{|c|}{Total}                                                                                                                                                                                         & \multicolumn{5}{c|}{}                                                                                                                                                                                                                  & 4359                                                                                        & 71                                                                                        \\ \hline
\end{tabular}
}
\vspace{-0.2in}
\label{tbl:venues}
\end{table}

The second to last column in Table~\ref{tbl:venues} shows the number of hits for each publication venue identified
by this search. In total, using this procedure, we identified 6,390 papers;
after removing duplicates, we obtained a set of 4,359 potentially relevant papers.

\subsection{Manual Filtering}
Next, we manually classified papers identified in the automated search into relevant and irrelevant for our survey.
To this end, we first randomly selected a set of 40 papers and used them as a pilot for drafting the inclusion and exclusion criteria.
Four authors of this survey independently read the abstract, introduction, and conclusions of each paper and classified it as \emph{relevant}, \emph{non-relevant}, or \emph{unsure}.
Each author also assigned a concise label for each of the \emph{relevant} and \emph{non-relevant} papers, specifying the reasons for inclusion/exclusion.

After completing this phase, all authors of this survey met to cross-validate the decisions,
and to consolidate and refine the inclusion/exclusion labels.
All disagreements between the authors (mostly between \emph{non-relevant} and \emph{unsure} papers) were resolved through a joint discussion.

In the second phase, we randomly selected an additional set of 40 papers, to validate our filtering process.
Four authors of this survey, again, independently read and categorized each of these papers, and all authors met to discuss the results.
While there were disagreements, with a rate of 6.25\%, between the assignment of \emph{non-relevant} and \emph{unsure} papers,
the inclusion/exclusion labels were consistent among the raters.

Specifically, we included papers that:
\vspace{-0.05in}
\begin{enumerate}
\item[(a)] study the correlation between properties of input data and the adversarial robustness of resulting model trained on this data; and/or
\item[(b)] present techniques to improve or disrupt a model's adversarial robustness through explicitly modifying some properties of the input data or its latent representation.
\end{enumerate}

We excluded papers that:
\vspace{-0.05in}
\begin{enumerate}
\item[(a)] discuss adversarial evasion attacks, but focus on features~\cite{Tong:Li:Hajaj:Chen:Xiao:Zhang:Vorobeychik:USENIX:2019}, models~\cite{Ghosh:Losalka:Black:AAAI:2019,  Xu:Chen:Liu:Chen:Weng:Hong:Lin:IJCAI:2019}, and training algorithms~\cite{Guo:Chen:Chen:Zhang:TPAMI:2021, Jeong:Shin:NeurIPS:2020} 
rather than data (44.2\%);
\item[(b)] discuss aspects of ML that are not related to adversarial robustness but rather related to
    accuracy~\cite{Ansuini:Laio:Macke:Zoccolan:NeurIPS:2019,Zhu:Jin:TNNLS:2020,Pope:Zhu:Abdelkader:Goldblum:Goldstein:ICLR:2021},
    robustness to distribution shifts not induced by adversaries~\cite{Mintun:Kirillov:Xie:NeurIPS:2021,Xu:Zhang:Zhang:Wang:Tian:CVPR:2021},
	privacy~\cite{He:Yang:CCS:2021, Hu:Salcic:Sun:Dobbie:Yu:Zhang:CSUR:2022}, and
	interpretability~\cite{Ghorbani:Abid:Zou:AAAI:2019, Guo:Mu:Xu:Su:Wang:Xing:CCS:2018} 
	(41.3\%); 
\item[(c)] propose new robustness evaluation metrics~\cite{Cheng:Deng:Zhao:Cai:Zhang:Feng:ICML:2021}
	or assessment frameworks~\cite{Ling:Ji:Zou:Wang:Wu:Li:Wang:SP:2019} (3.1\%);
\item[(d)] focus on poisoning~\cite{Zhang:Zheng:Gao:Miao:Su:Li:Ren:IJCAI:2019, Liu:Si:Zhu:Li:Hsieh:NeurIPS;2019} rather than evasion attacks (2.7\%); or
\item[(e)] discuss unrelated topics, e.g.,
	literature on blockchain~\cite{Huang:Kong:Zhou:Zheng:Guo:CSUR:2021},
	remote access trojan systems~\cite{Rezaeirad:Farinholt:Dharmdasani:Pearce:Levchenko:Macoy:USENIX:2018},
	or hardware systems~\cite{Tan:Wan:Zhou:Li:SP:2021}.
	These papers appeared in our search results as ``adversarial robustness'' is also desirable for non-ML systems,
	e.g., blockchain systems need to be robust against adversarial selfish miners or Denial-of-Service attacks 
	(5.8\%).
\end{enumerate}

As the inclusion/exclusion labels were consistent among the raters, we decided to proceed to the next phase: distribute the remaining 4,279 papers among the four authors and categorize them using these inclusion and exclusion criteria. In this phase, we instructed each rater to conservatively mark as \emph{unsure} papers with even a slight doubt in categorization.

Following this process, we identified 277 papers as either \emph{relevant} or \emph{unsure}.
We assigned a second reader to papers marked as \emph{unsure}, summarized each such paper in writing, and held a meeting with all the authors of this survey to categorize the paper as either \emph{relevant} or \emph{non-relevant}.
For a select set of papers where we could not confidently reach a decision, we emailed the paper authors to validate our understanding and decide on the relevance of the work (all such papers were excluded in the end).
Our analysis resulted in 71 \emph{relevant} papers. The distribution of these papers by publication venues is shown in the last column of Table~\ref{tbl:venues}.

\begin{figure*}[b]
\centering
    \vspace{-0.15in}
    \begin{minipage}{0.6\textwidth}
		\centering
		\includegraphics[scale=0.4]{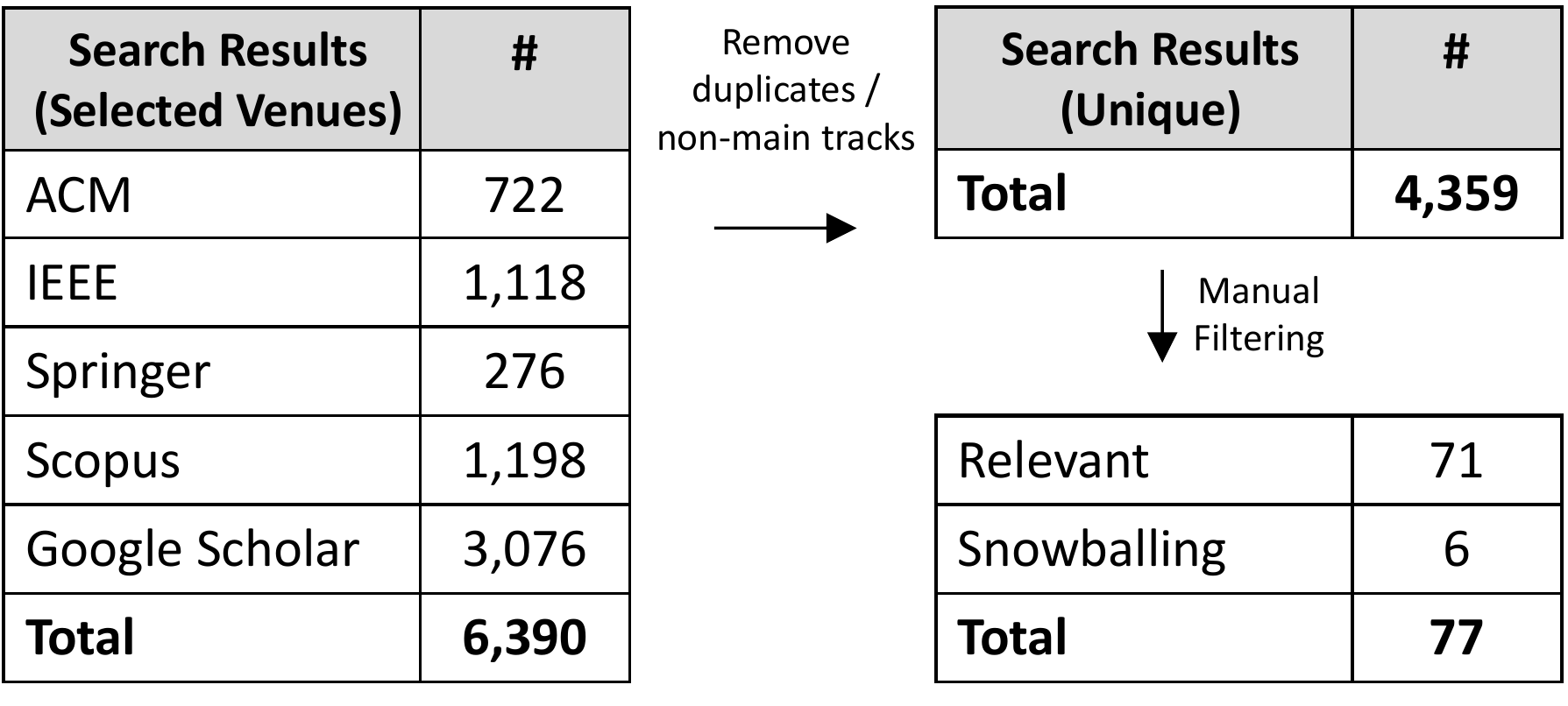}
		\caption{Summary of the collection and selection of papers.}
		\label{fig:collection}
		\vspace{-0.1in}
    \end{minipage}\hfill
    \begin{minipage}{0.38\textwidth}
        \centering
        \includegraphics[width=0.78\textwidth]{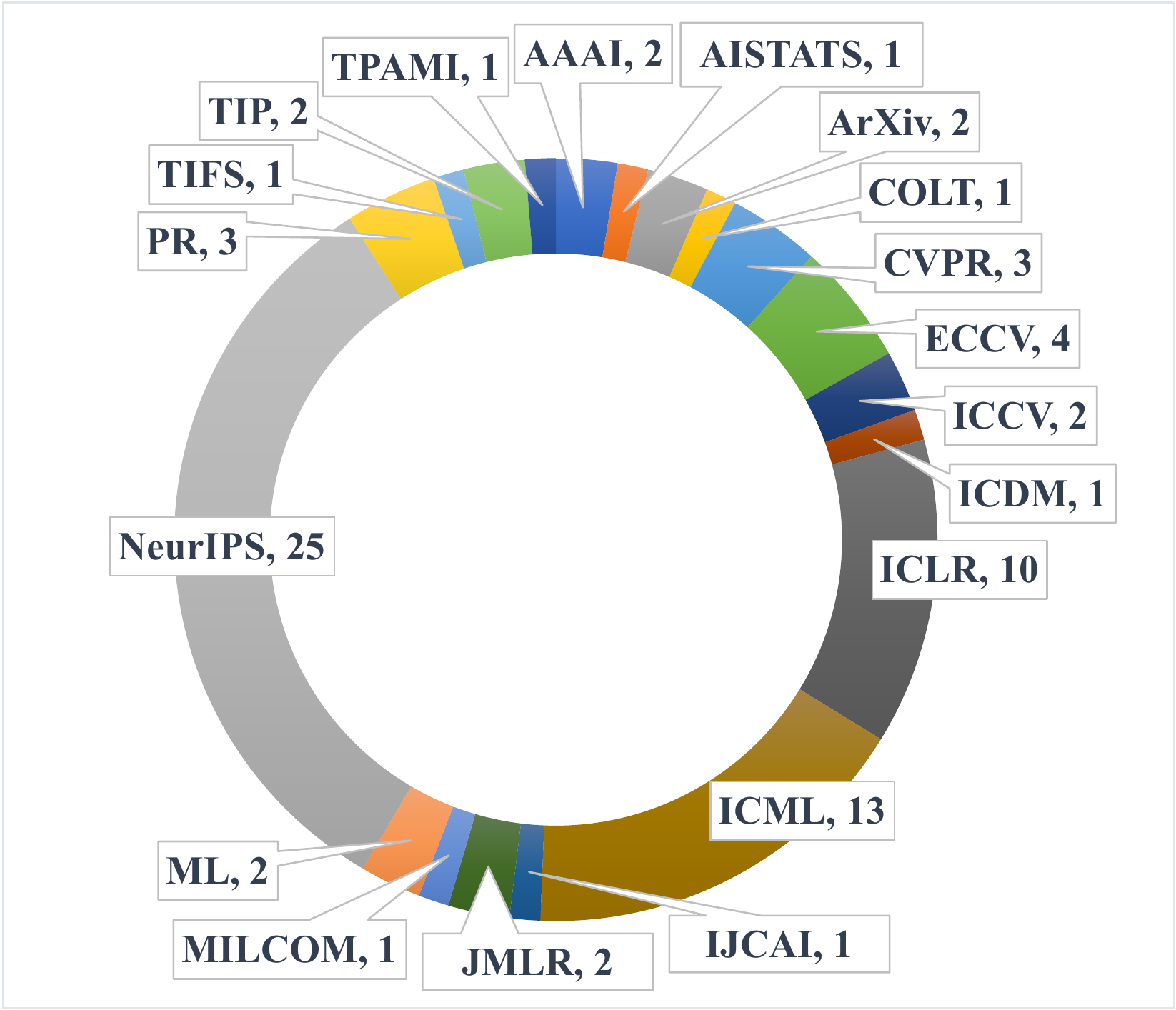} % first figure itself
%        \vspace{-0.1in}
        \caption{Breakdown by publication venues.}
        \label{fig:selection_by_venue}
    \end{minipage}
    \vspace{-0.05in}
\end{figure*}

We further assigned a second reader to each identified relevant paper, to extract and summarize its main findings.
To make sure that we included most of the relevant works on the topic, we also performed backward snowballing 
using the related work sections of the selected papers, which resulted in 6 additional papers, 
bringing our selection to 77 papers in total.
Fig.~\ref{fig:collection} summarizes our paper selection process. A more detailed view is also available online~\cite{appendix}. 

Fig.~\ref{fig:selection_by_venue} shows another view on the distribution of publication venues for all 77 papers included in our survey.
The majority of the papers (64, 83\%) are published in Machine Learning venues.
In fact, only the \emph{Advances in Neural Information Processing Systems (NeurIPS)} conference published 25 (around 32\%)
of all papers.
The distribution of selected papers by their publication year can be found in our online appendix~\cite{appendix}. 

\begin{figure*}[b]
	\centering
	\vspace{-0.2in}
	\includegraphics[width=13.6cm, height=3.6cm] {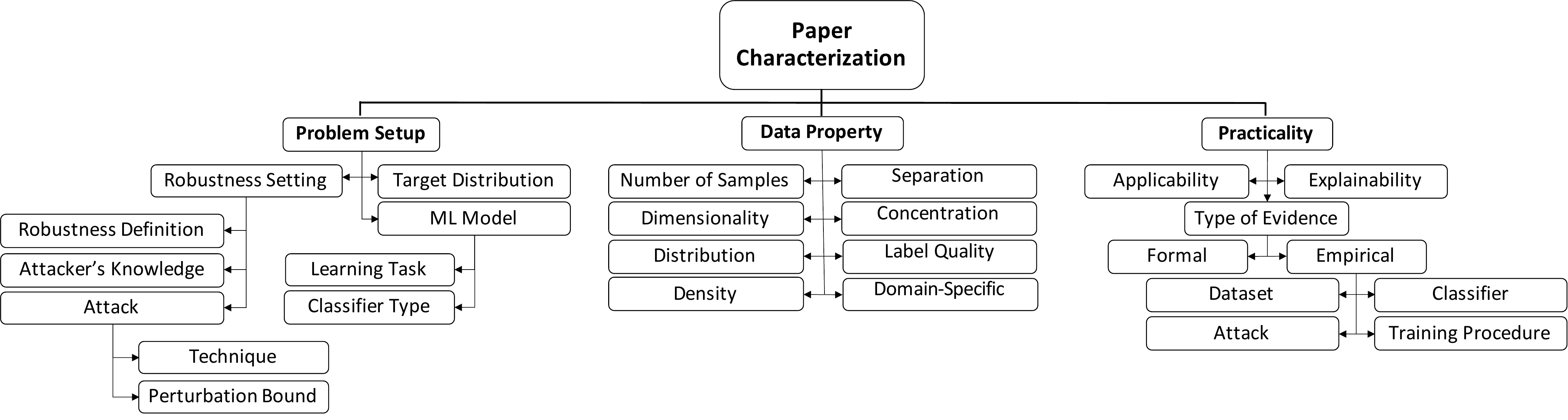}
	\vspace{-0.3in}
	\caption{Paper categorization dimensions.}
	\label{fig:categorization}
\end{figure*}

\subsection{Categorization of Selected Papers}
To better classify, discuss, and compare the papers, we proposed a categorization schema shown in Fig.~\ref{fig:categorization}.
To construct the schema, we followed an iterative process similar to the one we used for paper selection:
we first sampled ten papers from the final collection and each of the four authors independently proposed categorization attributes to describe these ten papers.
The proposed attributes were discussed by all authors while unifying related attributes, removing redundant ones,
and updating labels.
We then verified the applicability of the constructed schema on another set of ten papers and adjusted it based on a joint discussion.
We continued extending the schema after reading the remaining papers, to ensure its inclusiveness.

The resulting categorization schema, which we further use to analyze and describe the papers,
contains three high-level areas described below.
The detailed categorization of each papers along the attributes of this schema is available online~\cite{appendix}.

\vspace{0.02in}
\noindent
\textbf{1. Problem Setup} defines the scope of the proposed approach and the assumptions authors make in their work. Specifically,
\vspace{-0.05in}
\begin{itemize}
	\item \emph{Target Distribution} describes the type of data distributions the papers focus on.
	The common data distributions studied among the collected papers include Gaussian mixtures, Bernoulli mixtures, and Uniform distributions.
	
	\item \emph{ML Model} focuses on the studied model.
	It captures the \emph{learning task}, e.g., binary classification, multi-class classification, and regression, and
	\emph{classifier type}, e.g., parametric models, such as, neural networks and non-parametric models, such as, $k$-NNs.
	
	\item \emph{Robustness Setting} records the paper's definition of adversarial robustness.
	This includes the \emph{robustness definition} sub-category, which refers to how the authors measure robustness,
	e.g., error-rate-based or 
	radius-based. 
	The second sub-category, the \emph{attacker's knowledge}, reflects the level of information about the target system that the attacker can exploit: white-box, grey-box, or black-box.
	The last sub-category, the \emph{attack}, characterizes the \emph{technique} used to construct the attack,
	e.g., gradient- or non-gradient based, and the \emph{perturbation bound} considered, e.g., the type of $L_p$ norm.
\end{itemize}

\vspace{0.02in}
\noindent \textbf{2. Data Property} dimension includes the eight data properties we identified in the collected papers:
the \emph{number of samples}, data \emph{dimensionality}, \emph{distribution}, \emph{density},
\emph{concentration}, \emph{separation}, \emph{label quality}, and \emph{domain-specific} properties,
relevant in context of particular application domains.
We introduce and structure the discussion of the surveyed papers in Section~\ref{sec:results} around these properties.

\vspace{0.02in}
\noindent \textbf{3. Practicality} specifies how to apply the approach or technique introduced in each paper: 
\begin{itemize}
	\item \emph{Applicability} determines whether specific \emph{quantitative} metrics are provided to measure the data properties discussed in the paper or whether there are any concrete techniques proposed to modify these data properties.
	
	\item \emph{Explainability} determines whether the paper focuses on explaining (rather than establishing) the correlation between data property and robustness.
	
	\item \emph{Type of Evidence} records the type of arguments provided by the paper.
	This may be a \emph{formal proof}, an \emph{empirical evaluation}, or a combination of both.
    For cases where an empirical evaluation is performed, we also collect information about \emph{datasets} and \emph{classifiers} used, the applied \emph{training procedures} (standard vs. adversarial training), and the \emph{attack techniques} employed.
\end{itemize}

In what follows, we present the results of our analysis of the surveyed papers (Section~\ref{sec:results}) and
discuss our observations (Section~\ref{sec:discussion}).

\section{Results}
\label{sec:results}
We present the results of our analysis by organizing the papers according to the robustness-related data property they discuss: 
number of samples,
dimensionality,
type of distribution,
density,
concentration,
separation,
label quality, and
domain-specific properties.
Papers that discuss more than one data property are presented in all corresponding sections.
That is, in what follows, a paper can be discussed in more than one section.
Section~\ref{sec:results-summary} summarizes our findings.  
Furthermore, a detailed categorization and comparison of papers collected in this survey is available in our online appendix~\cite{appendix}.

To ease navigation, for each discussed data property,
we also include a map showing how the relevant papers relate to each other via their citation information.
We further annotate each paper with its \emph{applicability} and \emph{explainability} categories.
Specifically,
we annotate with an \roundrect{A} symbol papers that propose an actionable technique to modify or measure a robustness-related property;
we annotate with \roundrect{E} papers that put extra emphasis on explaining the correlation between a data property and robustness rather than establishing such a correlation.

\subsection{Number of Samples}
\label{sec:results-number-of-samples}

\begin{figure*}[b!]
  \centering
  \includegraphics[width=0.5\linewidth]{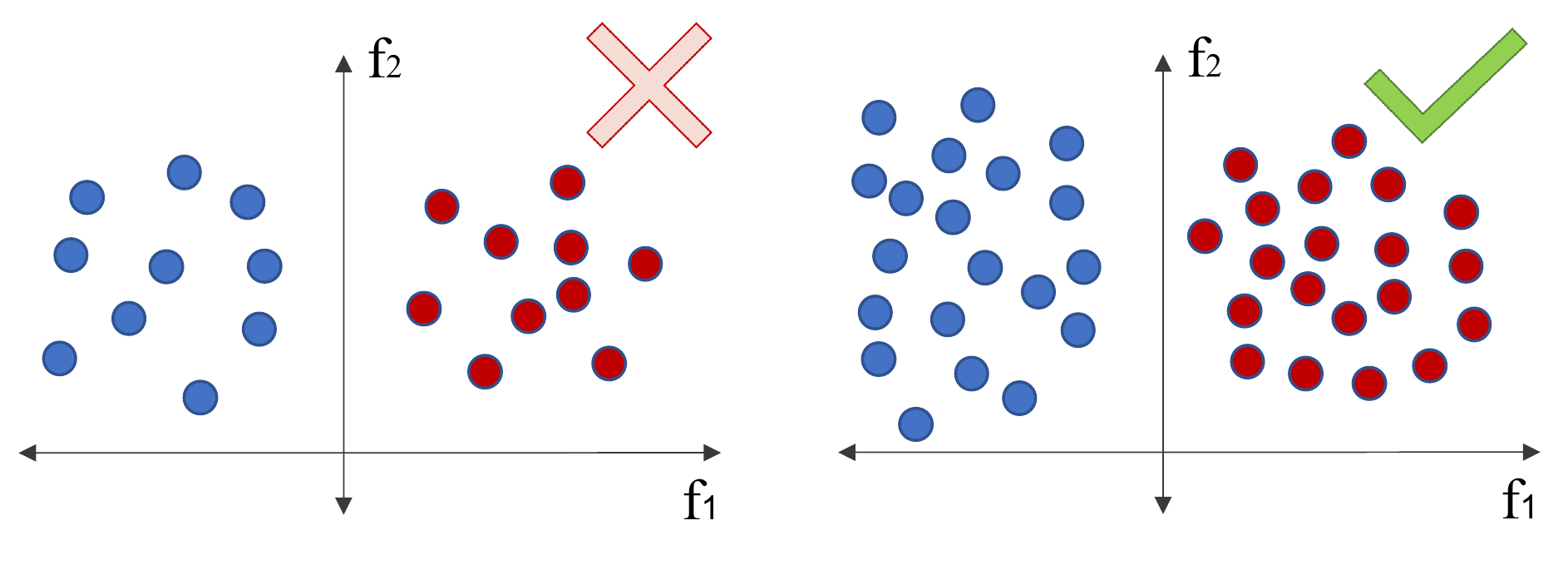}
  \vspace{-0.2in}
  \caption{Number of samples illustration.}
  \label{fig:samples_guideline}
  \vspace{-0.1in}
\end{figure*}

\emph{Number of samples} simply means the quantity of samples available in the training dataset.
For the example in Fig.~\ref{fig:samples_guideline}, where circles represent training samples for a two-class dataset,
the left dataset has fewer samples than the right dataset.

The term \emph{sample complexity} refers to the number of training samples required to achieve a certain model performance,
e.g., 90\%, in terms of either robust or standard generalization.
Then, \emph{sample complexity gap} refers to the difference in the number of samples required to achieve the same model performance for robust generalization as for standard generalization.

\begin{figure*}[t!]
	\centering
	\includegraphics[width=0.9\linewidth]{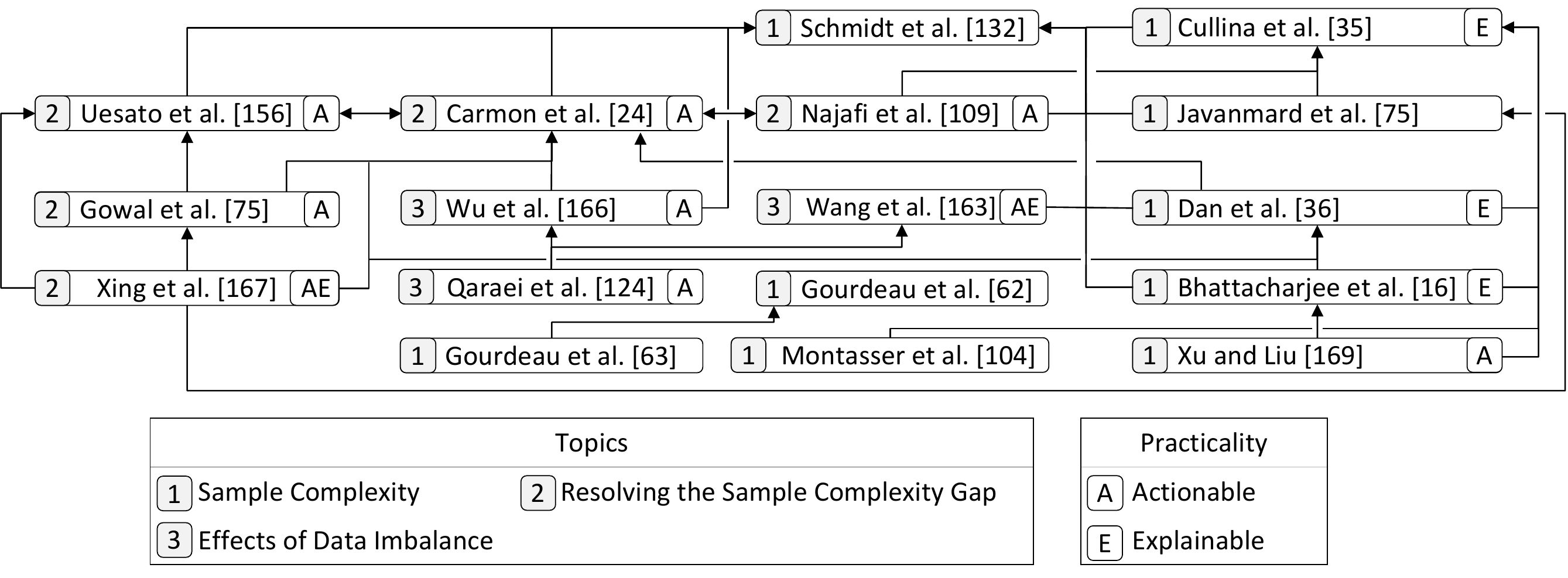} 
	\vspace{-0.1in}
	\caption{Papers discussing the number of samples.}
	\label{fig:samples}
	\vspace{-0.15in}
\end{figure*}

Papers studying the relationship between the number of training samples and the robustness of the resulting model
are shown in Fig.~\ref{fig:samples}. They can roughly be divided into
\roundrect{1} papers discussing sample complexity for robust generalization,
\roundrect{2} papers proposing techniques to resolve the sample complexity gap between the number of samples required to achieve the same level of robust and standard generalization, and
\roundrect{3} papers proposing techniques to deal with data imbalance, i.e., an unequal number of samples in different classes.

\noindent
\roundrect{1} {\bf Sample Complexity}.
Schmidt et al.~\cite{Schmidt:Santurkar:Tsipras:Talwar:Madry:NeurIPS:2018} observe that the number of training samples required for robust generalization is larger than the number of samples required for the equivalent-level standard generalization, i.e., that there exists a \emph{sample complexity gap} between the standard and robust generalization.
Specifically, for linear classifiers trained on a mixture of Gaussian distributions
(referred to as \emph{Schmidt's Gaussian mixture} in the remainder of this paper),
the authors prove that standard generalization requires a constant number of samples while
equivalent-level robust generalization requires a number of samples proportional to the data dimensionality
($O(\sqrt{d})$).
The gap in sample complexity persists for this data distribution in nonlinear classifiers as well.
Yet, the sample complexity gap disappears for nonlinear classifiers trained on a mixture of Bernoulli distributions;
these distributions also need substantially fewer samples than Gaussian mixtures.
The authors conclude that sample complexity for robust generalization depends on the distribution,
even when the same type of classifiers is considered.
Their experimental validation with the \mnist~\cite{MNST:1998}, \cifarten~\cite{CIFAR:ten:hundred:2009}, and \svhn~\cite{SVHN:dataset:2011} image datasets shows that \mnist, which is closer to a Bernoulli mixture, indeed requires a smaller number of training samples to achieve a reasonable robust generalization 
than the \cifarten and \svhn datasets, which are \mbox{closer to a Gaussian mixture}.

In follow-up work, Dan et al.~\cite{Dan:Wei:Ravikumar:ICML:2020} provide reasons for why robust generalization
requires more samples than standard generalization, focusing, again, on Gaussian mixture distributions.
Departing from the Signal-to-Noise ratio (SNR) metric that is based on the distance
between two Gaussian distributions and is known to capture the hardness of standard classification,
the authors propose a new Adversarial SNR (AdvSNR) metric,
defined as the minimum SNR for standard and adversarially perturbed data,
to capture the hardness of robust classification.
They then show that, given a dataset of a particular dimensionality,
the number of samples required to achieve the theoretically optimal, accurate classifier is
inversely proportional to SNR.
Likewise, the number of samples required to achieve the theoretically optimal, robust classifier
is inversely proportional to AdvSNR.
Because AdvSNR is never greater than SNR for a given dataset, it follows that
achieving the same robust generalization as standard generalization requires at least the same amount of samples.

Bhattacharjee et al.~\cite{Bhattacharjee:Jha:Chaudhuri:PMLR:2021} study the sample complexity gap for linear classifiers,
as a factor of data dimensionality (the number of features representing samples ) and 
separation (the distance between samples from different classes).
The authors show that the sample complexity gap is directly proportional to the dimensionality of the data when
the allowed perturbation radius of adversarial samples is similar to the distance between classes.
%That is, in this setup, the higher the dimensionality of the data, the more samples are needed for robust generalization.
However, such a gap no longer exists in well-separated data, when the perturbation radius is much smaller than 
\mbox{the distance between classes}.

Similarly, Gourdeau et al.~\cite{Gourdeau:Kanade:Kwiatkowska:Worrell:JMLR:2021, Gourdeau:Kanade:Kwiatkowska:Worrell:IJCAI:2022} show that,
for simple classifiers based on feature conjunctions and $\alpha$-log-Lipschitz distributions laying on a boolean hyper-cube, 
the sample complexity is a function of the data dimensionality $d$ and the adversarial perturbation budget.
Specifically, when the adversarial perturbation size is bounded by $log (d)$, the sample complexity is polynomial to the dimensionality;
when the perturbation size is at least $log (d)$, the sample complexity becomes superpolynomial to dimensionality.
Javanmard et al.~\cite{Javanmard:Soltanolkotabi:Hassani:COLT:2020} focus on adversarially-trained linear regression models
for standard Gaussian distributions.
The authors show that, when the number of samples is greater than the data dimensionality,
there exists a trade-off between adversarial and standard risks.
Moreover, this trade-off improves as the number of samples per \mbox{dimension increases}.

Cullina et al.~\cite{Cullina:Bhagoji:Mittal:NeurIPS:2018} give an upper bound on the number of samples needed
for robust generalization for the binary classification problem with linear classifiers in a distribution-agnostic setup,
with $L_p$ norm-bounded adversaries.
The authors derive the upper bound using the classifier VC dimension~--
a common measure of the capacity and the expressive power of the classifier, shown earlier
to be useful to determine the upper limit of sample complexity for standard generalization~\cite{Shalev-Shwartz:Ben-David:2014}.
They show that the VC dimension for learning adversarially-robust models remains the same as that for learning accurate models, which means that the upper bound of sample complexity is identical for standard and robust generalization
in this setup.
However, the authors demonstrate that this conclusion does not generalize to other types of classifiers and types of adversaries.

Similarly, Montasser et al.~\cite{Montasser:Hanneke:Srebro:NeurIPS:2022} study binary classifiers
constructed using 
the one-inclusion graph algorithm~\cite{Haussler:Littlestone:Warmuth:IC:1994}. 
They show that both the lower and the upper bound of sample complexity is finite, i.e., 
one can achieve robust generalization using a finite number of samples in this setup.

Xu and Liu~\cite{Xu:Liu:NeurIPS:2022} study sample complexity bounds in a multi-class setup. 
As VC dimension is defined only for the binary case, the authors propose 
Adversarial Graph dimension and Adversarial Natarajan dimension metrics, 
which extend their corresponding counterparts, 
Graph dimension~\cite{BenDavid:CesaBianchi:Haussler:Long:JCSS:1995} and Natarajan dimension~\cite{Natarajan:ML:2004}, 
commonly used in multi-class learning. 
The authors show that sample complexity is upper-bounded by the former and lower-bounded by the latter metric. 
 
\vspace{0.05in}
\noindent
\roundrect{2} {\bf Resolving the Sample Complexity Gap}.
As the number of labeled samples required to achieve robust generalization could be large and not readily available,
researchers explore cheaper alternatives, such as, unlabeled data and generated (fake) data.
Uesato et al.~\cite{Uesato:Alayrac:Huang:Stanforth:Fawzi:Kohli:NeurIPS:2019} and
Carmon et al.~\cite{Carmon:Raghunathan:Schmidt:Duchi:Liang:NeurIPS:2019}
concurrently proposed to use pseudo-labeling~\cite{Scudder:ITIT:1965}~--
a process of assigning labels to unlabeled samples using a classifier trained on a set of labeled samples,
assessing the effectiveness of their approaches on Schmidt's Gaussian mixture.
The main result of both works is that closing the sample complexity gap requires a number of unlabeled samples
proportional to the dimensionality of the data, albeit with a higher quantity than for the labeled samples,
likely due to the ``noise'' in generating labels.
The main difference between the works is that while Uesato et al. show that in their setup (a specific linear classifier)
the quantity of the required unlabeled samples only depends on data dimensionality,
Carmon et al.~\cite{Carmon:Raghunathan:Schmidt:Duchi:Liang:NeurIPS:2019} use a less restrictive setup and show that the quantity of unlabeled samples also depends on the original sample complexity for standard generalization.
Both of these works empirically evaluate the effectiveness of their proposed approaches on the \cifarten and \svhn datasets showing that unlabeled data could be a much cheaper alternative to labeled data for enhancing the robustness of models.

Najafi et al.~\cite{Najafi:Maeda:Koyama:Miyato:NeurIPS:2019}
note that the biggest risk of using a mixture of labeled and unlabeled datasets for learning adversarially robust models is
the uncertainty in sample labels.
Given an estimate of the quality of pseudo-labels,
the authors derive the minimum ratio between labeled and unlabeled samples required to avoid the additional
adversarial risk induced by label uncertainties. 

Instead of using unlabeled data, which might also be hard to find,
Gowal et al.~\cite{Gowal:Rebuffi:Wiles:Stimberg:Calian:Mann:NeurIPS:2021} suggest
using Generative Adversarial Networks (GANs) to generate labeled data.
The authors show that GANs are more effective than other methods, e.g., image cropping, when
producing additional samples.
This is because such models result in a more diverse dataset, which is beneficial for increasing robust accuracy.
Using images from \cifarten, \cifarhundred~\cite{CIFAR:ten:hundred:2009}, \svhn, and \tinyset~\cite{Torralba:Fergus:Freeman:TPAMI:TinyImageSet:2008}, the authors show that their proposed approach can significantly increase robust accuracy without the need for additional real samples.

Xing et al.~\cite{Xing:Song:Cheng:NeurIPS:2022} reason about the effects of real unlabeled and 
generated data on robustness of adversarially trained models. 
As real data is more informative for building decision boundaries, and as 
both real unlabeled and generated data need pseudo-labeling, 
the difference between real unlabeled and generated data boils down to the quality of data generators. 
Following this reasoning, the authors propose a strategy that assigns lower weights to the loss from generated samples compared to real samples during adversarial training, 
with the exact weights (i.e., the representation of the generator quality) being determined through cross-validation.

\vspace{0.05in}
\noindent
\roundrect{3} {\bf Effects of Data Imbalance}.
Wu et al.~\cite{Wu:Liu:Huang:Wang:Lin:CVPR:2021} analyze the adversarial robustness of DNNs on long-tail distributions: setups where the training data contains a large number of classes with few samples.
They show that robust generalization is harder to achieve on such distributions and compare the performance of
multiple adversarially trained classifiers that use learning algorithms specifically designed for such setups.
The comparisons show that scale-invariant classifiers~\cite{Wang:Wang:Zhou:Ji:Gong:Zhou:Li:Liu:CVPR:2018,Pang:Yang:Dong:Xu:Zhu:Su:NeurIPS:2020} result in higher robust accuracy as they
avoid assigning smaller weights to minority classes, which, in turn, promotes robust generalization by reducing bias in the decision boundary.

Both Wang et al.~\cite{Wang:Xu:Han:Xiaorui:Yaxin:Thuraisingham:Tang:ArrXiv:2021} and 
Qaraei et al.~\cite{Qaraei:Babbar:ML:2022}
propose to use re-weighted loss functions to improve robustness of under-represented classes.
Specifically, Wang et al. show that, 
for Gaussian mixture distributions, the robustness gap between classes depends on the amount of imbalance and the overall separation of a dataset. 
The authors thus propose modifying the loss function to assign weights correlated with data imbalance
while also promoting separation. 
Qaraei et al. focus on extreme multilabel text classification, where the output space is extremely large and the data follows a strongly imbalanced distribution. 
They also recommend using re-weighted loss functions to mitigate the robustness issue of the under-represented classes. 

\subsection{Dimensionality}
\label{sec:results-dimensionality}

\begin{figure*}[h]
\centering
  \begin{minipage}{0.55\textwidth}
  \centering
  \includegraphics[width=\linewidth]{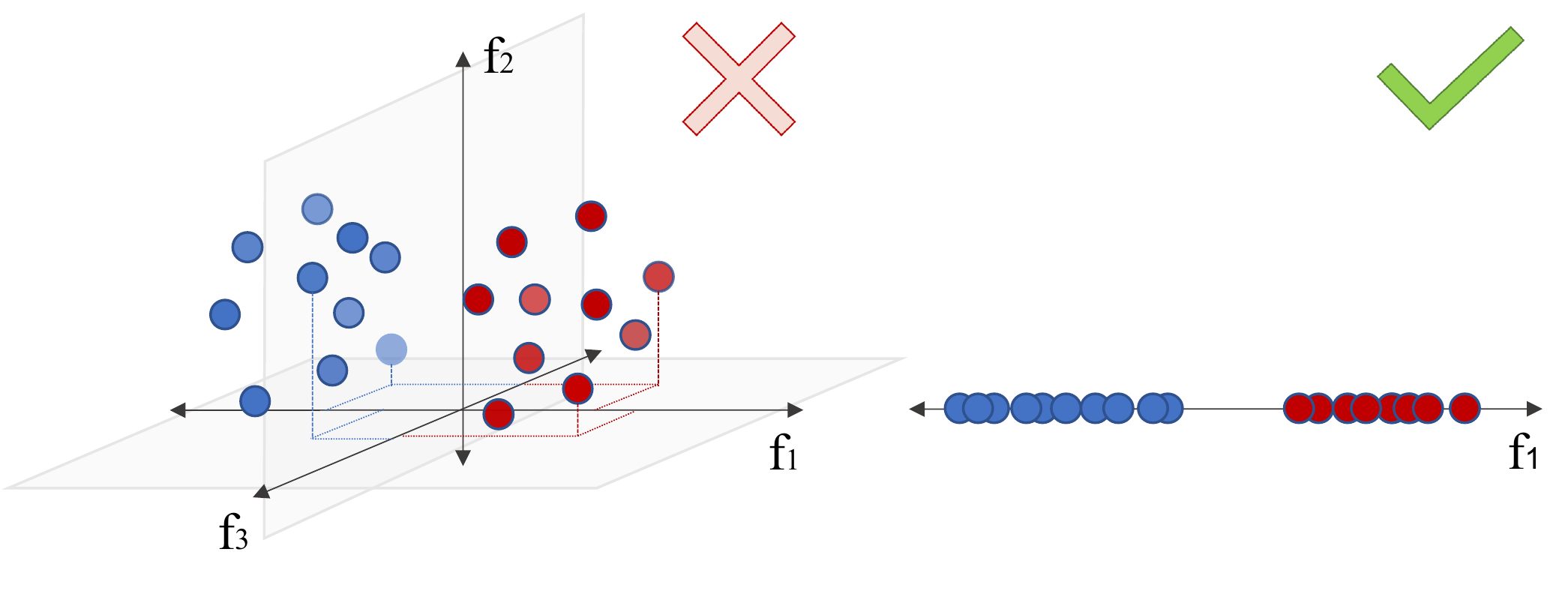}
  \vspace{-0.25in}
  \caption{Dimensionality illustration.}
  \label{fig:dimensionality_illustration}
  \vspace{-0.1in}
  \end{minipage}
  \begin{minipage}{0.44\textwidth}
  \vspace{0.1in}
  \centering
    \subcaptionbox{Actual. \label{fig:intrinsic_dimensionality_original}}{
     \includegraphics[width=0.42\linewidth]{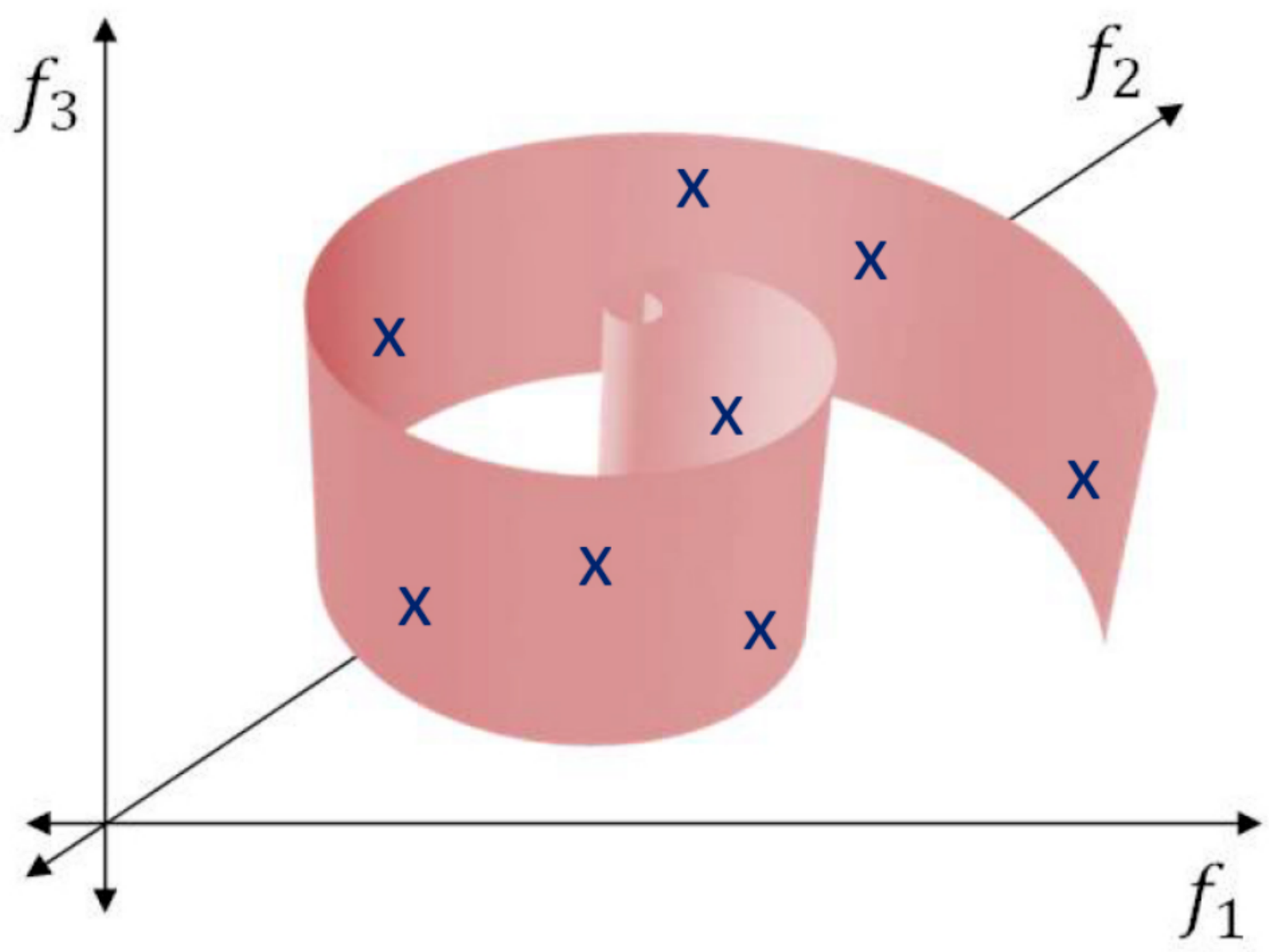}
   }
   \subcaptionbox{Intrinsic. 	 \label{fig:intrinsic_dimensionality_intrinsic}}{
    \includegraphics[width=0.42\linewidth]{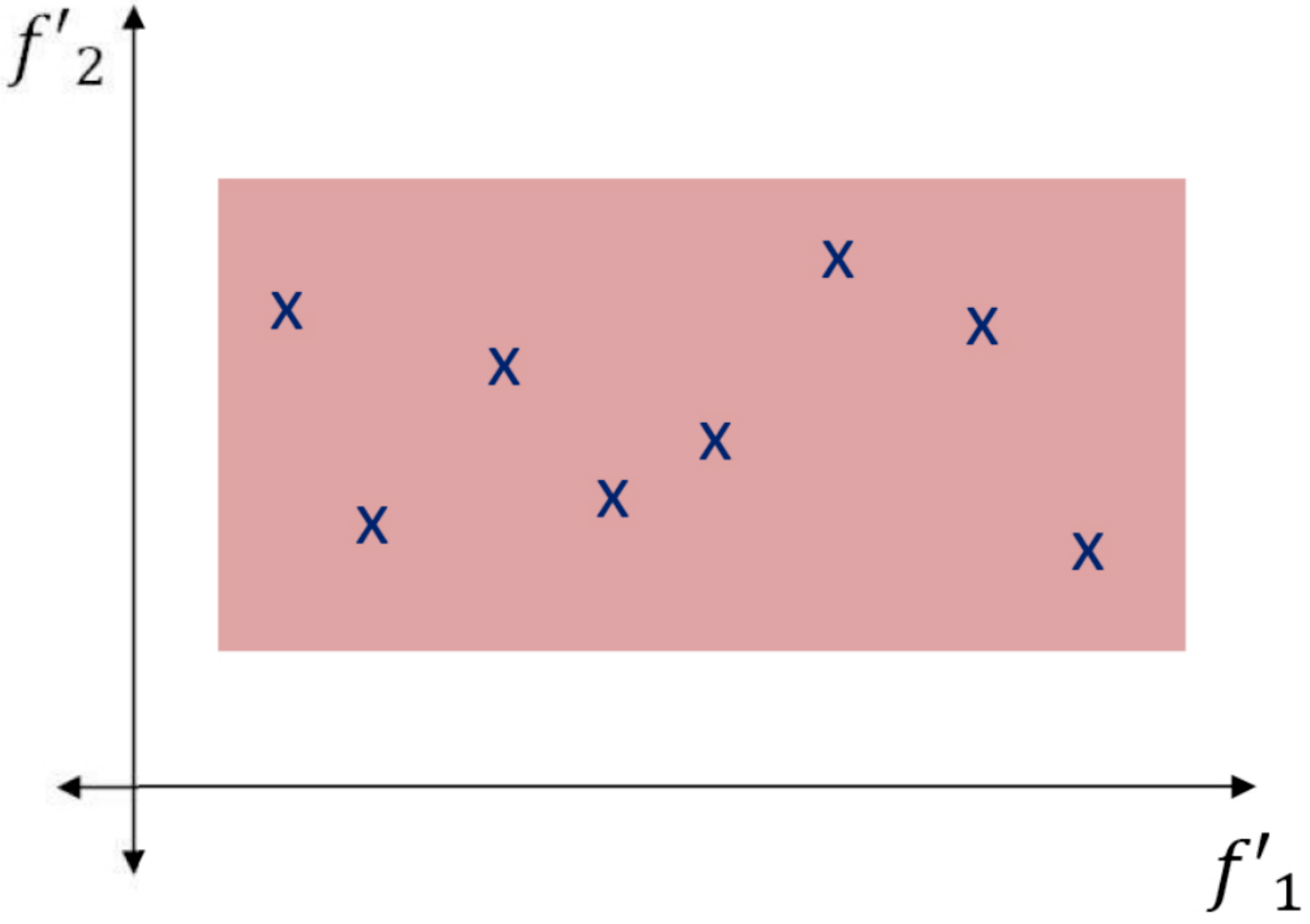}
    }
  \vspace{-0.05in}
  \caption{Actual and intrinsic dimensionality.}
  \label{fig:intrinsic_dimensionality}
  \vspace{-0.1in}
  \end{minipage}
\end{figure*}

\begin{figure*}[h]
  \centering
 % \vspace{-0.1in}
  \includegraphics[width=0.99\linewidth]{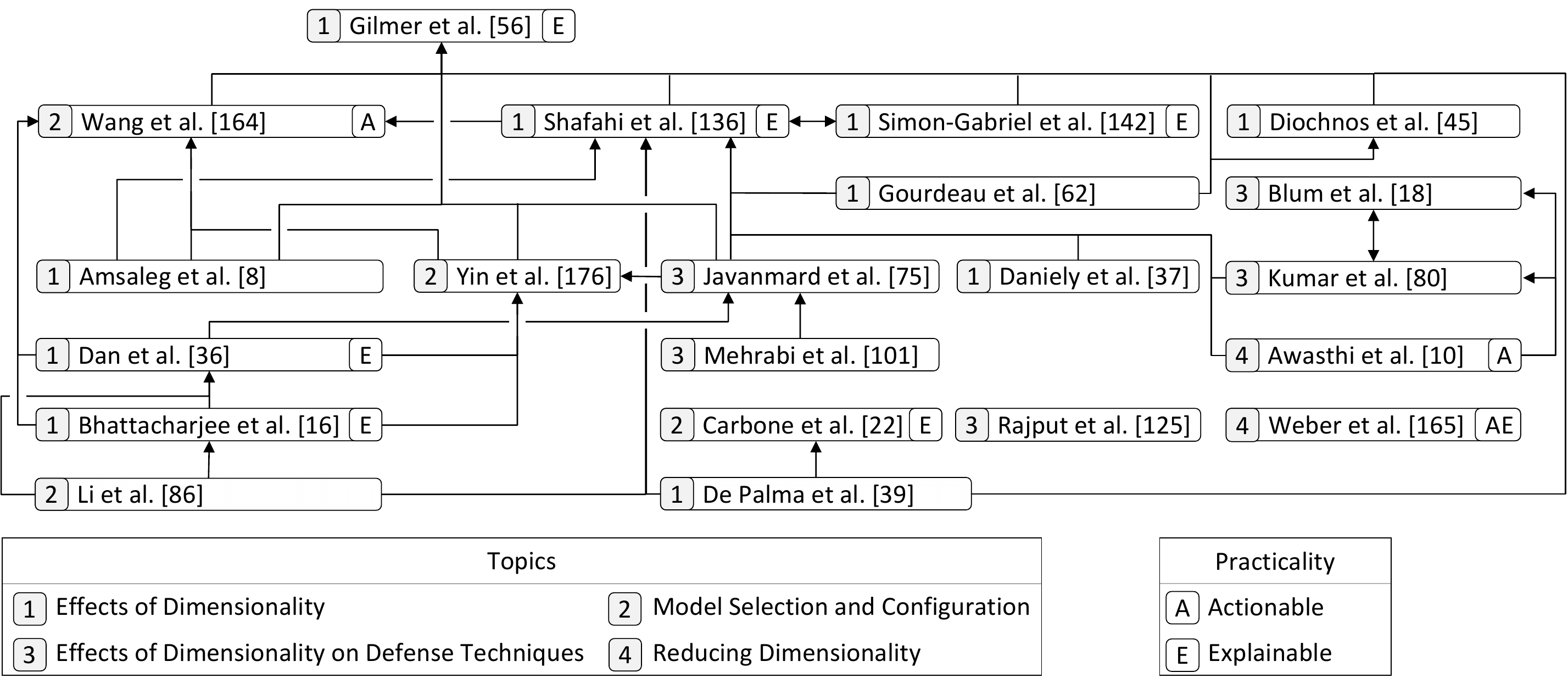}
  \vspace{-0.1in}
  \caption{Papers discussing dimensionality. }
  \label{fig:dimensionality}
  \vspace{-0.1in}
\end{figure*}

\emph{Dimensionality} refers to the number of features used to represent the data, e.g.,
features $f_1$, $f_2$, $f_3$ in Fig.~\ref{fig:dimensionality_illustration}.
For illustration purposes, we show a dataset with a dimensionality of three on the left-hand side of the figure and
a dataset with a dimensionality of one on the right-hand side.
\emph{Intrinsic dimensionality} refers to the number of features used in a minimal representation of the data.
Fig.~\ref{fig:intrinsic_dimensionality} shows an example of a case where the intrinsic dimensionality
is smaller than the actual dimensionality:
the samples in Fig.~\ref{fig:intrinsic_dimensionality}a are lying on a three-dimensional ``swiss roll''.
``Unwrapping'' the roll into a plain sheet, as shown in Fig.~\ref{fig:intrinsic_dimensionality}b,
makes it possible to distinguish between the samples using only two dimensions.

Papers studying the relationship between data dimensionality and adversarial robustness are shown in Fig.~\ref{fig:dimensionality}.
We divide them into papers
\roundrect{1} characterizing the hardness of robust generalization due to high dimensionality, 
\roundrect{2} suggesting robust model types and configurations for high-dimensional data,
\roundrect{3} discussing the impact of high dimensionality on existing defense techniques, and
\roundrect{4} utilizing dimensionality reduction techniques for improving robustness.

\vspace{0.05in}
\noindent
\roundrect{1} {\bf Effects of Dimensionality}.
A number of authors show that adversarial examples are inevitable in high-dimensional space.
Specifically, Gilmer et al.~\cite{Gilmer:Metz:Faghri:Schoenholz:Raghu:Wattenberg:Goodfellow:ICLR:2018} prove
this for a synthetic binary dataset composed of two concentric multi-dimensional spheres
(a.k.a. hyperspheres) in high-dimensional space ($>$100), showing that samples are, on average, closer to their nearest adversarial examples than to each other.
They also prove that the adversarial risk
of a model trained on this dataset only depends on its standard accuracy and dimensionality.
A similar result is shown by Diochnos et al.~\cite{Diochnos:Mahloujifar:Mahmoody:2018},
for a uniformly distributed boolean hypercube dataset, and
Shafahi et al.~\cite{Shafahi:Huang:Studer:Feizi:Goldstein:ICLR:2019}, for unit-hypersphere and unit-hypercube datasets.
De Palma et al.~\cite{DePalma:Kiani:Lloyd:ICML:2021} prove that irrespective of the model architecture, 
for a dataset with dimensionality $d$, the perturbation required to fool a classifier is inversely proportional 
to $\sqrt{d}$. This means that it gets easier to generate adversarial examples with an increase in dimensionality.

Another line of work analyzes the effect of dimensionality on the robustness of specific types of classifiers.
In particular,
Simon-Gabriel et al.~\cite{Simon-Gabriel:Ollivier:Scholkopf:Bottou:Lopez-Paz:ICML:2019} study feedforward neural networks
with ReLU activation functions and He-initialized weights, 
showing that a higher input dimensionality increases the success rate of adversarial attacks, regardless of the topology of the network.
The authors, however, demonstrate that regularizing the gradient norms of the network decreases the impact of the input dimension on adversarial vulnerability, thereby improving model robustness on high-dimensional inputs.
Daniely et al.~\cite{Daniely:Schacham:NeurIPS:2020} study the effect of dimensionality on ReLU networks
with random weights and with layers having decreasing dimensions.
Like Simon-Gabriel et al., the authors prove that the robustness of ReLU networks degrades proportionally to dimensionality.

Amsaleg et al.~\cite{Amsaleg:Bailey:Barbe:Erfani:Furon:Houle:Radovanovic:Nguyen:TIFS:2021}
focus on $k$-NNs and other non-parametric models that base predictions on the proximity of samples.
The authors use the Local Intrinsic Dimensionality metric to represent the intrinsic dimensionality
in the neighborhood of a particular sample $x$.
The authors build up on the observation that when this metric is high, there are more samples in close proximity of $x$
(as, otherwise, a more sparse neighborhood could be encoded in fewer dimensions).
Thus,
it is possible to arbitrarily change the neighborhood ranking of the nearest neighbor of $x$ using a small perturbation.
As predictions of proximity-based models are based on the nearest neighbor ranking, the adversarial risks increase in this setup.

All the aforementioned works are also in agreement with a number of papers discussed
in Section~\ref{sec:results-number-of-samples},
i.e., ~\cite{Dan:Wei:Ravikumar:ICML:2020,Bhattacharjee:Jha:Chaudhuri:PMLR:2021,Gourdeau:Kanade:Kwiatkowska:Worrell:JMLR:2021},
which show, in their respective settings, that sample complexity for robust generalization is proportional to dimensionality.

\vspace{0.05in}
\noindent
\roundrect{2} {\bf Model Selection and Configuration}.
%\jr{Jaskeerat-P41}
Wang et al.~\cite{Wang:Jha:Chaudhuri:ICML:2018} prove that the optimal $k$ for producing robust $k$-NN classifiers depends on the dimensionality $d$ and number of samples $n$ of the given dataset ($k = \Omega(\sqrt{dn \text{ log}(n)})$).
However, they note that for high-dimensional data, the optimal $k$ might be too large to use in practice.
The authors thus focus on improving the robustness of 1-NN algorithms through sample selection,
showing the effectiveness of their approach on 
the \halfmoon, \mnistv, and \abalone datasets.

Yin et al.~\cite{Yin:Kannan:Bartlett:ICML:2019} show that transferring a robust solution found on training data
to test data gets more difficult as the dimensionality of data increases.
However, constraining the classifier weights mitigates this problem.
Specifically, the authors prove that constraining the weights by $L_p$ norm, for $p > 1$, leads to
a performance gap between training and test data that has a polynomial dependence on dimensionality;
when the weights are constrained by $L_1$ norm, the performance gap has no dependence on dimensionality.
Li et al.~\cite{Li:Jin:Zhong:Hopcroft:Wang:NeurIPS:2022} rather focus on model configuration.
The authors show that robust generalization in networks with ReLU activation requires  
the network size to be exponential in original and intrinsic data dimensionalities, 
even in the simplest case when the underlying distribution is linearly separable.

Carbone et al.~\cite{Carbone:Wicker:Laurenti:Patane:Bortolussi:Sanguinetti:NeurIPS:2020} study neural networks,
showing that adversarial vulnerability arises due to the gap between the actual and intrinsic dimensionality,
a.k.a. degeneracy.
The authors show that adversarial example generations in high-dimensional degenerate data can be performed by
using gradient information of a neural network, to move the samples in the direction normal to the data manifold.
As such, example generation exploits the additional dimensions without changing the ``semantics'' of the perturbed sample.
The authors then show that Bayesian Neural Networks are more robust than other neural networks to gradient-based attacks:
due to their randomness, they make gradients less effective for crafting attacks.

\vspace{0.05in}
\noindent
\roundrect{3} {\bf Effects of Dimensionality on Defense Techniques}.
High dimensionality also poses challenges to defense techniques that aim to improve robustness.
Specifically, Blum et al.~\cite{Blum:Dick:Manoj:Zhang:JMLR:2020} focus on randomized smoothing~--
a technique that improves robustness by generating noisy instances of a (possibly perturbed) sample
and then making predictions for the sample based on an aggregation of predictions for its noisy instances.
The authors show that the amount of noise required to defend against $L_{p}$ adversaries, for $p > 2$,
is proportional to dimensionality.
They further demonstrate that, for high-dimensional images, randomized smoothing indeed fails to generate instances that preserve semantic image information.
In a similar line of work, Kumar et al.~\cite{Kumar:Levine:Goldstein:Feizi:ICML:2020} show that the certified radius decreases
as the dimensionality increases when using randomized smoothing for certifying robustness for a given $L_{p}$ radius.

Adversarial training~-- a defense technique that improves model robustness by adaptively training a model
against possible adversarial examples~-- often incurs a trade-off between standard and adversarial accuracy~\cite{Tsipras:Santurkar:Engstrom:Turner:Madry:ICLR:2019, Zhang:Yu:Jiao:Xing:Ghaoui:Jordan:ICML:2019}:
optimizing for high robust accuracy results in a drop in standard accuracy and vice versa.
Mehrabi et al.~\cite{Mehrabi:Javanmard:Rossi:Rao:Mai:ICML:2021} build up on the work of Javanmard et al.~\cite{Javanmard:Soltanolkotabi:Hassani:COLT:2020},
discussed in Section~\ref{sec:results-number-of-samples}, which 
showed that, for a finite number of training samples, 
the trade-off between adversarial and standard accuracy improves as the number of samples per dimension increases.
Mehrabi et al. further extend this result for unlimited training data and computational power, observing that,
for an unlimited number of training samples, the trade-off between adversarial and standard accuracy improves as the dimension of the data decreases.

Data augmentation is another common defense technique that aims to improve robustness of a model by creating
perturbed samples at radius $r$ from a certain subset of original samples in training data.
Rajput et al.~\cite{Rajput:Feng:Charles:Loh:Papailiopoulos:ICML:2019}
prove, for linear and certain nonlinear classifiers, that the number of augmentations required for robust generalization depends on the dimensionality of data, 
i.e., it is at least linearly proportional to dimensionality for any fixed radius $r$.
Thus, data augmentation becomes more expensive for high-dimensional data.

\vspace{0.05in}
\noindent
\roundrect{4} {\bf Reducing Dimensionality}.
Following the idea that the gap between the actual and intrinsic dimensionality contributes to adversarial vulnerability,
Awasthi et al.~\cite{Awasthi:Jain:Rawat:Vijayaraghavan:NeurIPS:2020} propose to use Principal Component Analysis (PCA)~\cite{Jolliffe:2002}
to decrease the dimensionality of data before applying randomized smoothing.
As a result, a larger amount of noise can be injected to perturb samples, thus improving robustness without compromising accuracy.
The authors apply the proposed ideas to image data, showing that the combination of PCA and randomized smoothing is more beneficial than using randomized smoothing alone.
Weber et al.~\cite{Weber:Zaheer:Rawat:Menon:Kumar:NeurIPS:2020} show, for hierarchical data,
that changing the representation from Euclidean to hyperbolic space reduces the dimensionality
without sacrificing semantic information embedded in the input data.
\subsection{Distribution}
\label{sec:results-distribution}

\begin{figure*}[h]
\centering
  \begin{minipage}{0.52\textwidth}
  \centering
  \vspace{0.2in}
  \includegraphics[width=\linewidth]{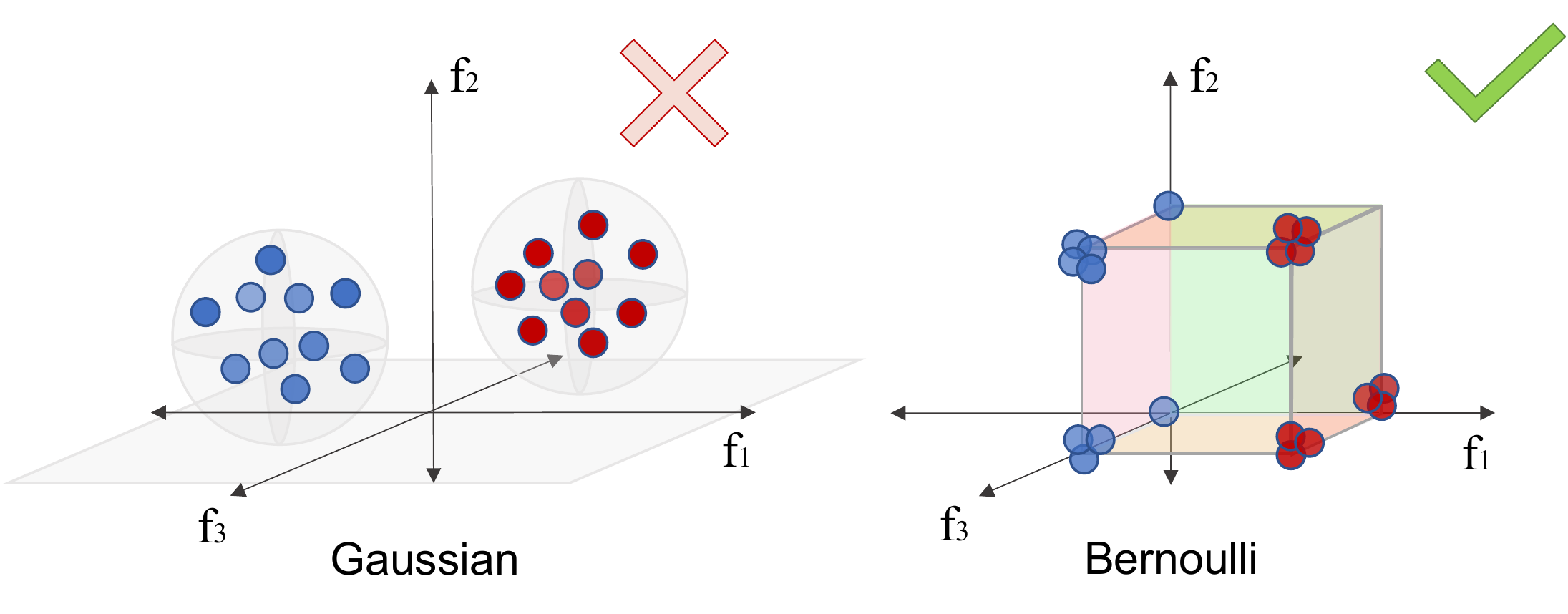}
  \vspace{-0.22in}
  \caption{Distribution illustration.}
  \label{fig:distribution_illustration}
  \vspace{-0.1in}
  \end{minipage}
  \hspace{0.1in}
  \begin{minipage}{0.44\textwidth}
	\centering
	\vspace{-0.1in}
	\includegraphics[width =\textwidth]{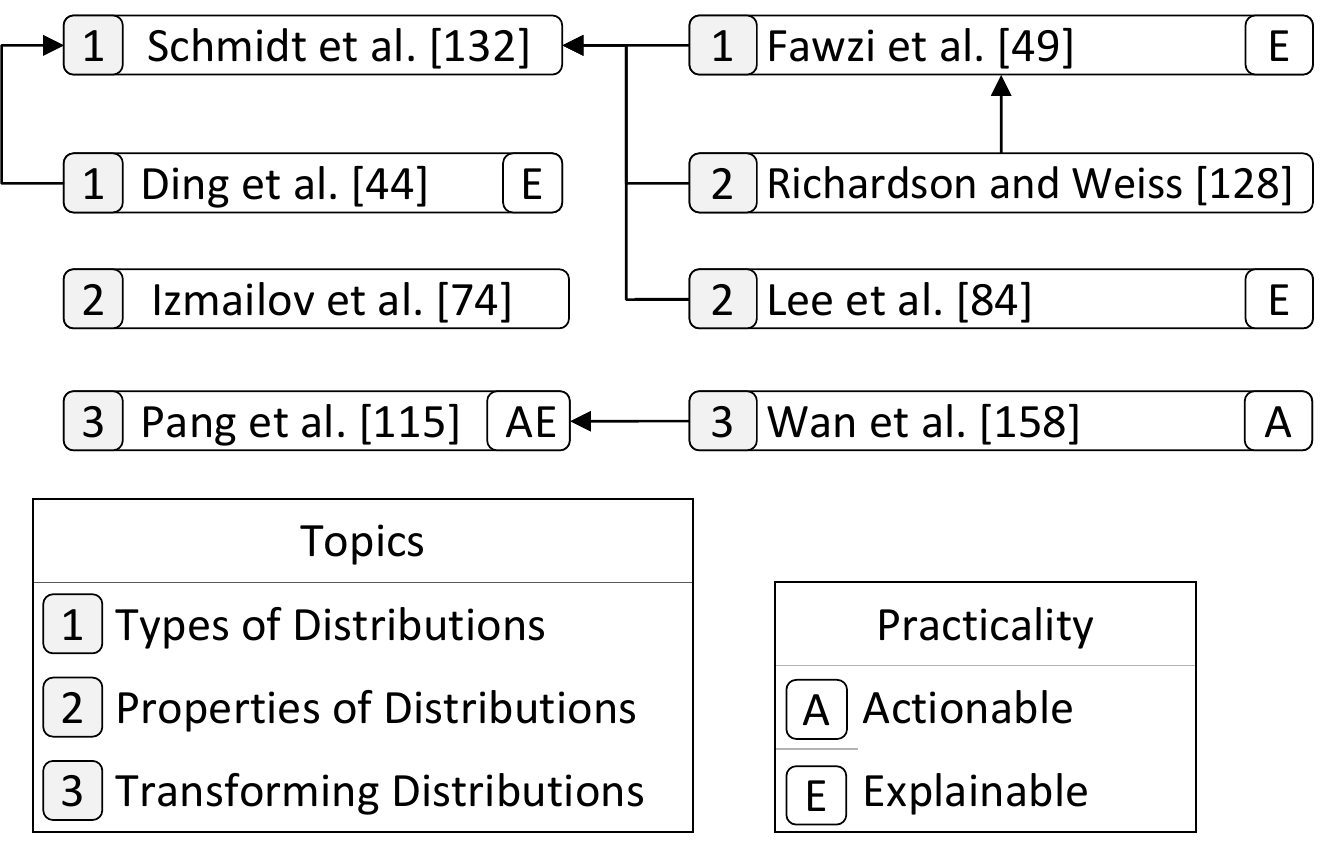}
	\vspace{-0.25in}
    \caption{Papers discussing distribution.}
  \label{fig:distribution}
  \end{minipage}
  \vspace{-0.1in}
\end{figure*}

\emph{Distribution} refers to a function that encodes how samples lie in space, usually by giving the probabilities of their occurrence in particular regions.
Common types of distributions, such as uniform, Bernoulli, and Gaussian
are introduced in Section~\ref{sec:background_distribution}.
Fig.~\ref{fig:distribution_illustration} shows examples of datasets that follow a Gaussian distribution (left) and a Bernoulli distribution (right).
The term \emph{variance} refers to a measure of dispersion that takes into account the spread of all data points in a dataset.
Specifically, the \emph{variance of a distribution} measures the dispersion of samples from the mean;
\emph{feature variance} measures the dispersion of samples over a particular feature only.
We say that a distribution satisfies \emph{symmetry} when distributions on either side of the mean mirror each other.

Papers that discuss how distribution properties, including variance and symmetry, influence models' robustness
are shown in Fig.~\ref{fig:distribution}.
They can be categorized into:
\roundrect{1} papers showing that model robustness depends on the underlying data distribution,
\roundrect{2} papers identifying properties of distributions that improve robustness, and
\roundrect{3} papers introducing techniques to transform distributions into ones that are more optimal for robustness.

\vspace{0.05in}
\noindent
\roundrect{1} {\bf Types of Distributions}.
As discussed in Section~\ref{sec:results-number-of-samples},
Schmidt et al.~\cite{Schmidt:Santurkar:Tsipras:Talwar:Madry:NeurIPS:2018} prove, for nonlinear classifiers,
that a mixture of Gaussian distributions incurs higher sample complexity
for robust generalization than a mixture of Bernoulli distributions.
Likewise, Ding et al.~\cite{Ding:Lui:Jin:Wang:Huang:ICLR:2019} show that
a distribution shift alone can affect robust accuracy while retaining the same standard accuracy.
Specifically, the authors prove that uniform data lying on a unit cube results in more robust models than
uniform data lying on a unit sphere.
They further experiment with \mnist and \cifarten datasets,
applying existing semantically lossless transformations, namely \emph{smoothing} and \emph{saturation},
to cause the distribution shift.
The results of this experiment show that robustness decreases gradually when transforming \mnist from
a unit-cube-like to a unit-sphere-like distribution and increases for \cifarten when going the opposite way;
in both cases, the models retain \mbox{their standard accuracy}.

Fawzi et al.~\cite{Fawzi:Fawzi:Fawzi:NeurIPS:2018} study the robustness of data distributions modeled by
a smooth generative model~-- a type of generative model that maps samples from input space to output
space while preserving their relative distances, e.g., to compress data.
The authors show that smooth generative models with high-dimensional input space
produce data distributions that make any classifier trained on this data inherently vulnerable.
The authors conclude that non-smoothness and low input space dimensionality
are desirable when modeling data with generative models.

\vspace{0.05in}
\noindent
\roundrect{2} {\bf Properties of Distributions}.
Izmailov et al.~\cite{Izmaliov:Sugrim:Chadha:McDaniel:Swami:MILCOM:2018} show that, in a binary classification setting,
features with small variance in both classes and means close to each other
cause adversarial vulnerability.
Moreover, a feature with a small variance in one class can still cause vulnerability
even if the means of this feature in both classes are farther separated
but the second class has a larger feature variance.
Intuitively, that is because models tend to assign non-zero weights to such features,
which can be leveraged by attackers to shift the classification into the wrong class.
That is, even small perturbations in such features can shift data points to another class.
To increase robustness, the authors suggest removing such features, either based on domain knowledge
or based on feature evaluation metrics, such as, mutual information~\cite{Shannon:1949}.

Similarly,
Lee et al.~\cite{Lee:Lee:Yoon:CVPR:2020} prove that decreasing feature variance in individual classes
can increase robustness for Schmidt's Gaussian mixtures.
These mixtures have equivalent feature variances for all classes and separated means.
In such a setting, low feature variance implies that the feature has a strong correlation with the class and
perturbing this feature will unlikely result in a vulnerability 
(i.e., will likely result in a semantically-meaningful change).
However, even when features have low variance, if these features are
non-robust~\cite{Ilyas:Santurkar:Tsipras:Engstrom:Tran:Madry:NeurIPS:2019}, i.e., hold no semantic information,
and have a smaller variance in the training data than
in the underlying true population,
they will still cause adversarial vulnerability as adversarially trained models tend to overfit to them.
As a countermeasure, the authors propose a label-smoothing-based data augmentation technique that uses
continuous instead of discrete values for labels and acts like a regularization method that prevents the model from overfitting to such features.

\begin{wrapfigure}{r}{0.30\textwidth}
  \vspace{-0.35in}
  \begin{center}
    \includegraphics[width=0.26\textwidth]{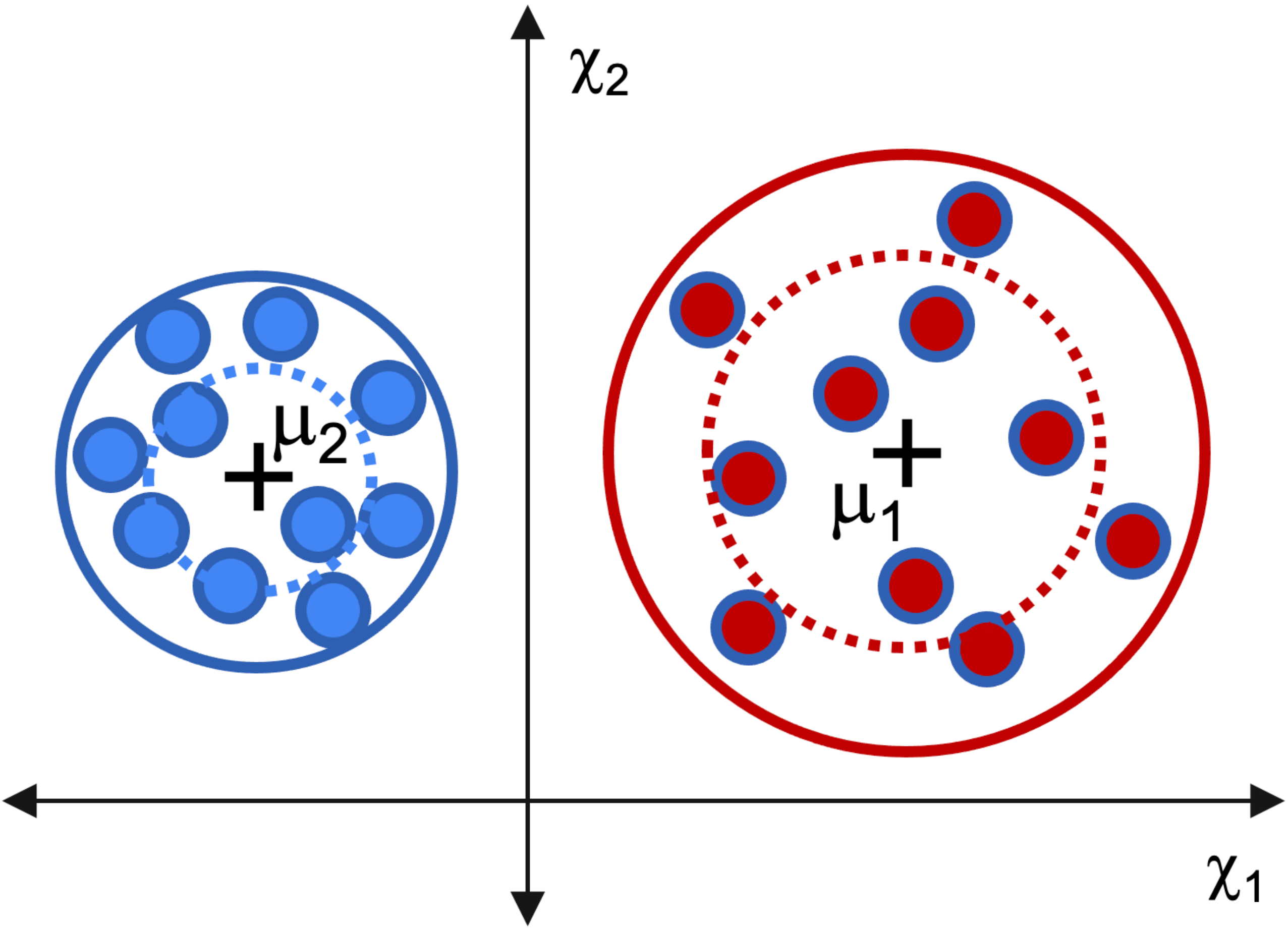}
  \end{center}
    \vspace{-0.15in}
  \caption{Asymmetrical dataset.}
  \label{fig:asymetric_data}
  \vspace{-0.1in}
\end{wrapfigure}
Richardson and Weiss~\cite{Richardson:Weiss:JMLR:2021} claim that adversarial vulnerability can be caused by
sub-optimal data distributions and/or sub-optimal training methods.
The authors define synthetic binary datasets (of images) that use Gaussian distributions with separated means
and say that a dataset is symmetric if and only if classes have the same variance.
They further prove that even an optimal classifier is non-robust when the underlying dataset has strong asymmetry,
as in the
example in Fig.~\ref{fig:asymetric_data}.
If the dataset is symmetric the optimal classifier is provably robust,
even though a sub-optimal training method can still cause vulnerability when trained on this dataset.

\vspace{0.05in}
\noindent
\roundrect{3} {\bf Transforming Distributions}.
Both Pang et al.~\cite{Pang:Du:Zhu:ICML:2018} and Wan et al.~\cite{Wan:Chen:Yu:Wu:Zhong:Yang:TPAMI:2022}
change the latent DNN feature representation to be similar to Gaussian mixtures. 
Specifically, Pang et al. show that,
for Linear Discriminant Analysis (LDA) classifiers trained on Gaussian mixtures, the robustness radius of LDA
is proportional to the distance between the Gaussian centers.
The robustness of LDA is further maximized for symmetric Gaussian mixtures.
The authors thus modify the DNN loss function to create a latent feature representation similar to symmetric Gaussian mixtures
and further replace the last layer of DNN from commonly used Softmax Regression~\cite{Cramer:2002} to LDA.
To achieve the desired robustness radius, the authors compute the coordinates of the desired
Gaussian centers (as a function of the number of classes and the dimensionality of the input data)
and feed this data to the loss function.
Departing from the assumption that symmetric Gaussian mixtures are advantageous for the underlying model robustness,
Wan et al. modify the DNN loss function to compute the centers of the Gaussians directly while generating symmetric \mbox{Gaussian feature distributions}.

\subsection{Density}
\label{sec:results-density}

\begin{figure*}[h]
	\centering
	\vspace{-0.15in}
	\begin{minipage}{0.53\textwidth}
		\centering
		\includegraphics[width=\linewidth]{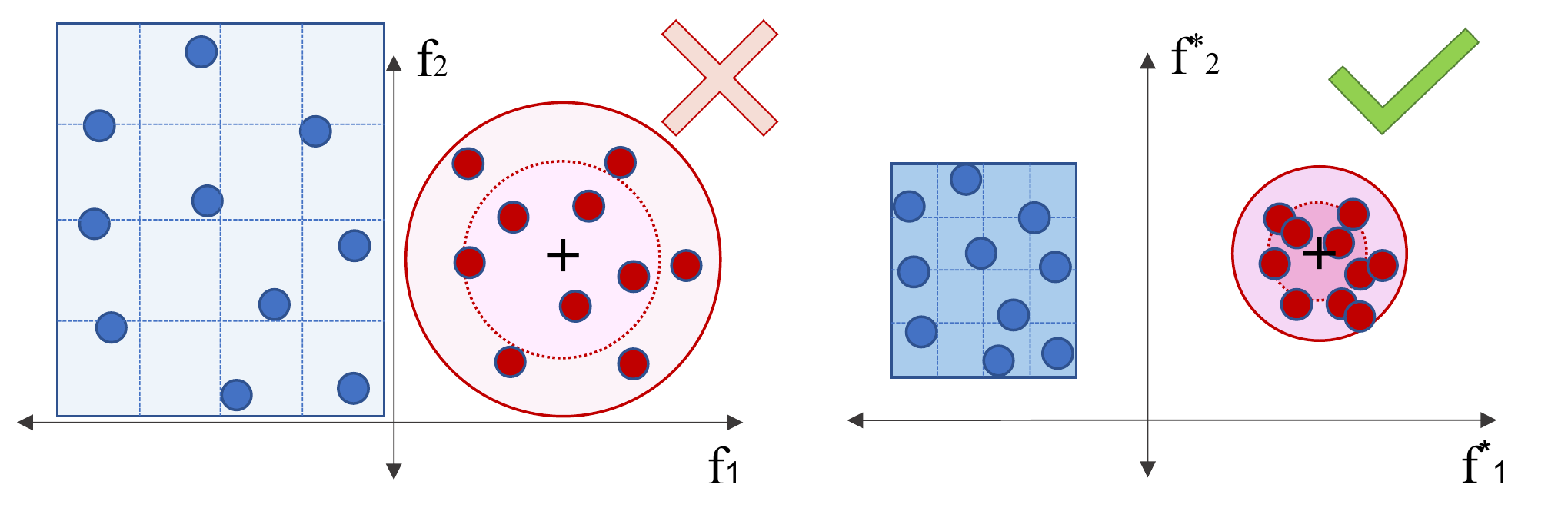}
		\vspace{-0.14in}
		\caption{Density illustration.}
		\label{fig:density_guideline}
		\vspace{-0.1in}
	\end{minipage}
	\hspace{0.1in}
	\begin{minipage}{0.42\textwidth}
		\centering
		\includegraphics[width =\textwidth]{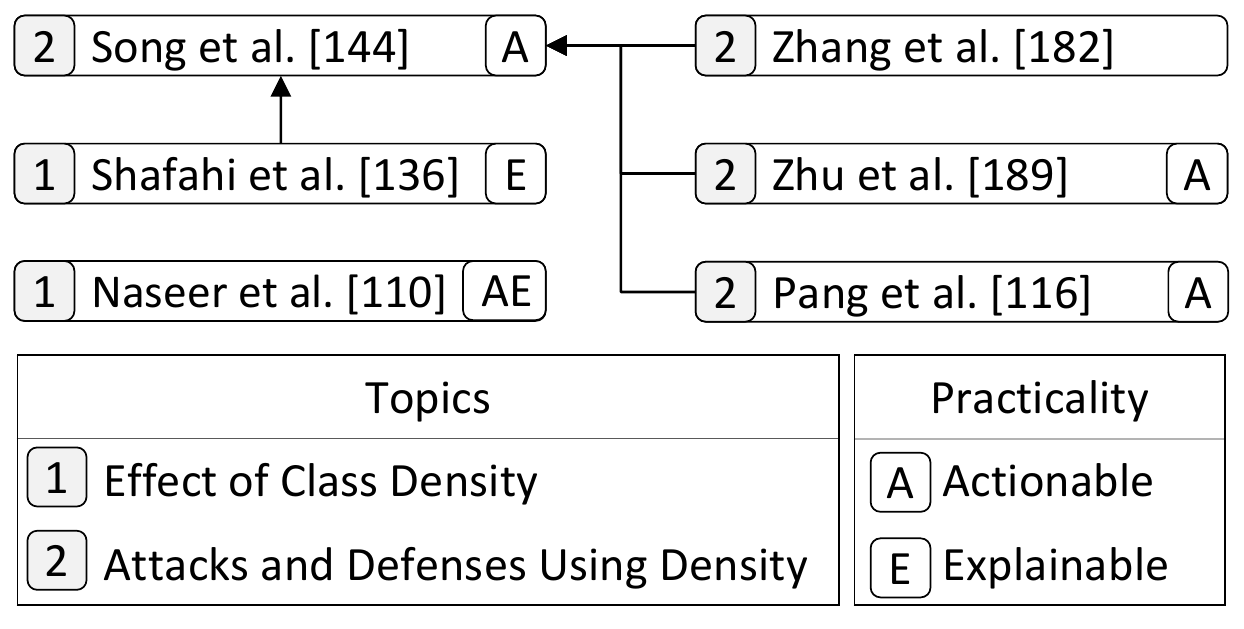}
		\vspace{-0.25in}
		\caption{Papers discussing density. }
		\label{fig:density}
	\end{minipage}
\end{figure*}

\emph{Density} measures the closeness of samples in a particular bounded region.
For continuous data, it is mathematically described by the probability density function,
which gives the probability for a variable to take a certain range of values.
For discrete data, it is described as the probability mass function, which
gives the probability for a variable to take a particular value.
We say that an area is dense when there is a high probability that random samples lie in the same area, i.e.,
close to each other.
For example, the dataset on the right-hand side of Fig.~\ref{fig:density_guideline} contains a larger number of samples in close proximity and, thus, is more dense than the dataset on the left-hand side of the figure.
Furthermore, density can be defined over samples from one class, in which case, it is referred to as \emph{class density}.

Papers that study how density influences adversarial robustness are shown in Fig.~\ref{fig:density}. They can roughly be divided into
\roundrect{1}~papers discussing the effect of class density on robustness and
\roundrect{2}~papers proposing attacks and defenses using density information.

\vspace{0.05in}
\noindent
\roundrect{1} {\bf Effects of Class Density}.
Shafahi et al.~\cite{Shafahi:Huang:Studer:Feizi:Goldstein:ICLR:2019} 
show that datasets with a higher upper bound of class density lead to better robustness.
In particular, for image datasets, the authors show that images of lower complexity,
e.g., with simple objects on plain backgrounds, have a higher correlation among adjacent pixels.
Datasets comprised of such images have a higher density, as pixel values are more frequently repeated, and, thus, lead to better robustness.
The authors confirm this observation by showing that classifiers trained on \mnist, which has a lower image complexity and thus higher density than \cifarten, are more robust than those trained on \cifarten.
Furthermore, the authors state that class density is a better predictor of robustness than dimensionality:
even after up-scaling \mnist to the same dimensionality as \cifarten,
it still has a higher density and thus results in more robust classifiers than \cifarten.

Naseer et al.~\cite{Naseer:Prabakaran:Hasan:Shafique:ML:2023} show that imbalance in class densities 
is a more substantial predictor of robustness bias among classes than the difference in the number of samples.
The authors further propose a two-step strategy to remove this bias through data augmentation.
First, they gradually increase the perturbation size in samples from all classes and identify which classes get misclassified with the smallest perturbation size, treating this as an indication of low density. 
They then generate realistic and diversified samples for these classes, 
to reduce imbalance, which in turn leads to improved robustness.

\vspace{0.05in}
\noindent
\roundrect{2} {\bf Attacks and Defenses Using Density}.
Several works note that adversarial examples are commonly found in low-density regions of the training dataset, as models are unable to learn accurate decision boundaries using a small number of samples from these regions.
Zhang et al.~\cite{Zhang:Chen:Song:Boning:Dhillon:Hsieh:ICLR:2019}
propose an attack strategy that retrieves candidate samples from low-density regions and
perturbs them to generate adversarial examples.
The authors demonstrate that, even after adversarial training, models will not be robust to adversarial attacks that target these low-density regions.

A similar finding by Zhu et al.~\cite{Zhu:Sun:Li:ICLR:2022} suggests that adversarial examples from low-density regions have a higher probability of being transferable between different models trained on the same dataset.
Based on this observation, the authors propose an attack that increases the transferability of adversarial examples by
identifying perturbation directions that maximize both the adversarial risk and
the alignment with the direction of density decrease for the underlying data distribution,
i.e., move samples towards regions with lower density.

Departing from the same idea that low-density regions are prone to adversarial attacks,
Song et al.~\cite{Song:Kim:Nowozin:Ermon:Kushman:ICLR:2017} focus on creating a defense mechanism
that uses generative models to detect if a sample comes from a low-density region when making predictions.
If so, the sample is moved towards a more dense region of the training data as a ``purification'' step.

To harden models directly, Pang et al.~\cite{Pang:Xu:Dong:Du:Chen:Zhu:ICLR:2020} propose a new loss function for DNNs,
to learn dense latent feature representations.
The authors first show that the commonly used Softmax Cross-Entropy loss function induces sparse representations
(i.e., with low class density),
which lead to vulnerable models. This is because a low number of samples in close proximity to each other prevent a model from learning reliable decision boundaries.
They then propose a loss function that explicitly encourages feature representations to concentrate around class centers;
like in their earlier work~\cite{Pang:Du:Zhu:ICML:2018}, the authors compute the coordinates of the desired
class centers (as a function of the number of classes and the dimensionality of the input data)
to maximize the distances between the centers.
The authors demonstrate that the proposed approach improves robustness under both standard and adversarial training.

\subsection{Separation}
\label{sec:results-separation}

\begin{figure*}[h]
	\centering
	\vspace{-0.25in}
	\begin{minipage}{0.43\textwidth}
		\centering
		\vspace{0.25in}
		\includegraphics[width=0.68\linewidth]{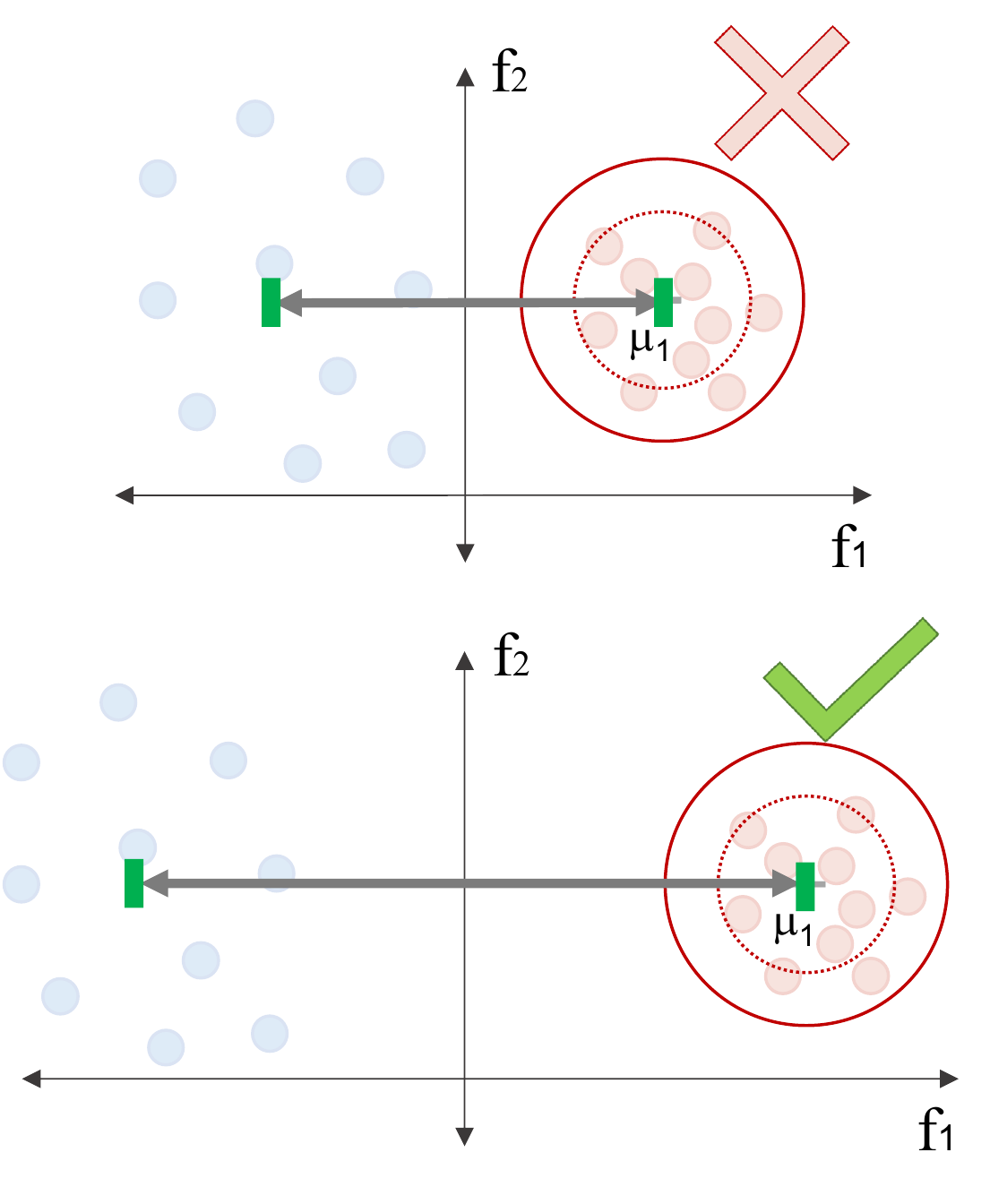}
		\vspace{0.07in}
		\caption{Separation illustration.}
		\label{fig:separation_guideline}
		\vspace{-0.3in}
	\end{minipage}
	\begin{minipage}{0.5\textwidth}
		\vspace*{0.10in}
		\centering
 		\includegraphics[width =\textwidth]{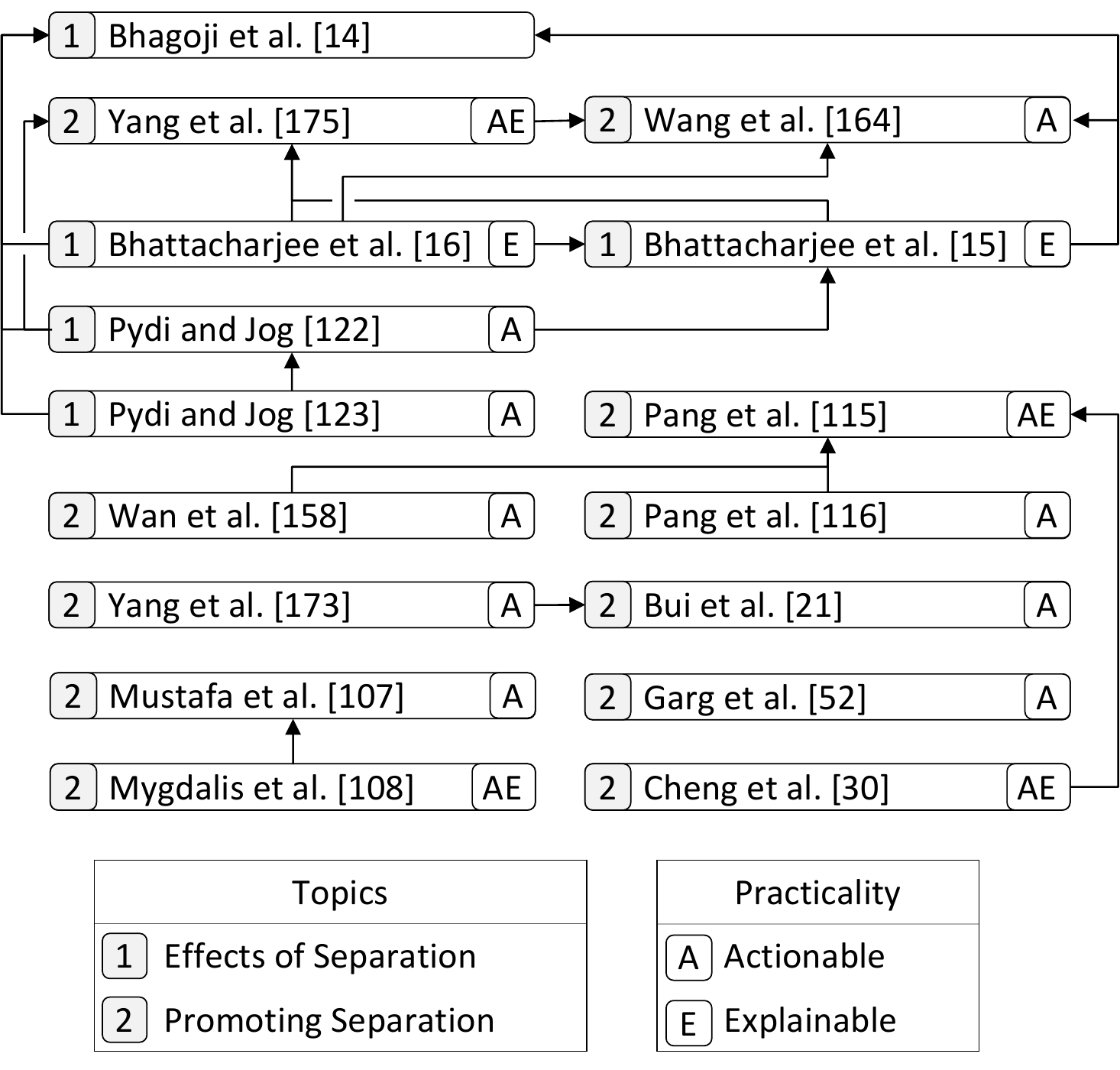}
		\vspace{-0.25in}
		\caption{Papers discussing separation.}
		\label{fig:separation}
	\end{minipage}
    \vspace{-0.1in}
\end{figure*}

Closely related to density, \emph{separation} refers to the distance between classes.
Fig.~\ref{fig:separation_guideline} shows examples of not well-separated (top) and well-separated (bottom) datasets.
Intuitively, learning an accurate classifier is easier when data is well-separated as samples from different classes are farther apart and samples from the same class are closer together.
Different metrics to quantify separation include
the \emph{optimal transport distance},
which computes the minimum distance required to transport samples from one class to another, and
\emph{inter-class distance}, which computes the distance between samples in different classes.

Papers that discuss data separation in relation to adversarial robustness are shown in Fig.~\ref{fig:separation}.
They can roughly be divided into
\roundrect{1} papers showing the effect of separation on robustness and
\roundrect{2} papers proposing techniques to promote separation and, thus, increase robustness.

\vspace{0.05in}
\noindent
\roundrect{1} {\bf Effects of Separation}.
Bhagoji et al.~\cite{Bhagoji:Cullina:Mittal:NeurIPS:2019} calculate lower bounds for adversarial risk 
in a binary classification setting using the optimal transport distance.
The authors show
that the lower bound decreases as the distance between the two class distributions increases,
i.e., a classifier becomes more robust with better separation.
Based on this result, they estimate the minimum adversarial risks for image datasets, like \mnist and \cifarten,
showing that the theoretically calculated risks are lower than the empirical values achieved by the state-of-the-art defense models.
The authors conclude that there is still room for improving existing techniques.

Pydi and Jog~\cite{Pydi:Jog:ICML:2020, Pydi:Jog:NeurIPS:2021} arrive at a similar conclusion~-- 
that robustness improves as separation between classes increases.
The authors further focus on datasets with simple univariate distributions, such as Gaussian and uniform.
They propose a technique to construct classifiers that can achieve the optimal,
lowest possible adversarial risk for a given separation between classes.
The main idea behind this technique is to analyze the optimal way to
transport samples from one class to another 
(which represents the smallest perturbation needed to create adversarial examples) and 
further use this information to identify the decision boundary that induces the maximal distance required to transport 
samples between classes. 
That is, the approach maximizes the distance between samples of each class and the decision boundary, 
resulting in an optimally robust classifier.

Bhattacharjee et al.~\cite{Bhattacharjee:Chaudhuri:ICML:2020}
prove that certain non-parametric models, such as k-NNs, 
are inherently robust when trained on a large number of well-separated samples.
This is because these classifiers make predictions based on neighborhoods and well-separated data ensures that
samples in close proximity to each other share the same labels.
In their later work, discussed in Section~\ref{sec:results-number-of-samples}~\cite{Bhattacharjee:Jha:Chaudhuri:PMLR:2021},
the authors
show that, in well-separated data, robust accuracy is independent of dimensionality and a robust linear classifier can be learned without the need for a large number of training samples.
This result shows that adversarial vulnerability can be efficiently tamed by increasing separation.

\vspace{0.05in}
\noindent
\roundrect{2} {\bf Promoting Separation}.
Yang et al.~\cite{Yang:Rashtchian:Wang:Chaudhuri:AISTATS:2020} propose a sample-selection-based technique to improve the adversarial robustness of non-parametric models by increasing the separation among the training data.
In particular,
as non-parametric models tend to learn complex decision boundaries when the training samples from different classes are close to each other,
the authors propose to remove the smallest subset of samples so that all pairs of differently labeled samples
remain separated even when perturbed by the maximum perturbation size. 
Wang et al.~\cite{Wang:Jha:Chaudhuri:ICML:2018}, already discussed in Section~\ref{sec:results-dimensionality}, 
focus on improving robustness of 1-NN classifiers. 
Such classifiers struggle to take advantage of points close together with opposite labels, resulting in worse robustness.
Hence, the authors propose retaining the largest subset of training samples that are
(i) well-separated and (ii) in high agreement on labels with their nearby samples
(a.k.a., highly confident). 
The authors show that their approach outperforms adversarially trained 1-NNs. 

For non-parametric classifiers, a more effective strategy is to enforce separation in the latent representations. 
Specifically, Mustafa et al.~\cite{Mustafa:Khan:Hayat:Goecke:Shen:Shao:TPAMI:2020}
attribute the cause of adversarial vulnerability to close proximity of classes in latent space.
Hence, they propose a loss function to learn intermediate feature representations that
separate different classes into convex polytopes, i.e., polyhedra in higher dimensions,
that are maximally separated.
Mygdalis et al.~\cite{Mygdalis:Pitas:PR:2022} propose a loss function to separate classes into hyperspheres, 
such that samples in a class have minimum distance from their hypersphere center and 
maximum distance from the remaining hyperspheres. 
The authors demonstrate that their approach outperforms that of
Mustafa et al.~\cite{Mustafa:Khan:Hayat:Goecke:Shen:Shao:TPAMI:2020} and other baselines 
w.r.t. standard and robust accuracy for \cifarten, \cifarhundred and \svhn.

Bui et al.~\cite{Bui:Le:Zhao:Montague:deVel:Abraham:Phung:ECCV:2020} observe that the adversarial vulnerability of DNNs arises from a large difference in intermediate layer values between clean and adversarial data.
They thus propose to modify the loss function so that it results in an intermediate latent representation 
that has high similarity between clean and their corresponding adversarial samples,
while promoting large inter-class and small intra-class distances and 
increased margins from class centers to decision boundaries.
Likewise,
Pang et al.~\cite{Pang:Du:Zhu:ICML:2018} and Wan et al.~\cite{Wan:Chen:Yu:Wu:Zhong:Yang:TPAMI:2022} discussed in Section~\ref{sec:results-distribution},
as well as Pang et al.~\cite{Pang:Xu:Dong:Du:Chen:Zhu:ICLR:2020} discussed in Section~\ref{sec:results-density},
improve DNN robustness by separating centers of the produced latent distributions, which, in turn, 
increases the separation \mbox{between classes}.
Similarly, Cheng et al.~\cite{Cheng:Zhu:Zhang:Liu:PR:2023} propose improving separation by 
enforcing equal variance in all directions for all classes (Distribution Normalization) and 
maximizing the minimum margin between any two classes (Margin Balance).

Yang et al.~\cite{Yang:Feng:Du:Du:Xu:ICDM:2021} propose a representation-learning technique
to learn feature representations that
bring samples of class $C$ and adversarial examples generated for $C$ into close proximity
while separating the samples of $C$ from both
(i) adversarial examples generated for other classes and misclassified as class $C$ and
(ii) samples from other classes.
These separations are enforced by the loss function proposed by the authors.
The authors show that their approach improves the resulting model robustness compared with standard DNNs.

Garg et al.~\cite{Garg:Sharan:Zhang:Hu:NeurIPS:2018} propose an approach to generate well-separated features for a dataset using graph theory.
Specifically, they convert the input dataset into a graph, where vertices correspond to the input data points and
edges represent the similarity between the data points (e.g., calculated using Euclidean distance).
The authors prove that features extracted using the eigenvectors of the Laplacian matrix capturing the structure of the graph
will have significant variation across the data points,
while being robust to small perturbations. These qualities make them good candidates for robust features.
The authors then demonstrate that a linear model trained on the \mnist dataset with 20 features generated using their approach is more robust to $L_2$-norm-based transfer attacks than a fully connected neural network trained on the full pixel values of the \mnist dataset.

\subsection{Concentration}
\label{sec:results-concentration}

\emph{Concentration} of a dataset refers to the ``concentration of measure'' phenomenon from measure theory~\cite{Talagrand:1996:AnnalsProbability}.
In a nutshell, concentration is the minimum value of a measured function over all valid measurable sets,
after an $\epsilon$-expansion.  
More formally,
for a metric probability space $(\mathcal{X}, \mu, d)$ with instance space $\mathcal{X}$, probability measure $\mu$, and distance metric $d$,
the concentration function $h$ is defined as: $h(\mu, \alpha, \epsilon) = $ inf$_{A \subseteq \mathcal{X}} \{\mu(A_\epsilon) : \mu(A) \geq \alpha\}$ for any $\alpha \in (0,1)$ and $\epsilon \geq 0$~\cite{Mahloujifar:Zhang:Mahmoody:Evans:NeurIPS:2019}.
Here $A_\epsilon$ refers to the $\epsilon$-expansion of set $A$, defined as $A_\epsilon = \{ x : d(x, A) \leq \epsilon\}$.

\begin{figure*}[h]
  \centering
  \begin{minipage}{0.40\textwidth}
  \centering
  \vspace{-0.08in}
  \includegraphics[width =0.82\textwidth]{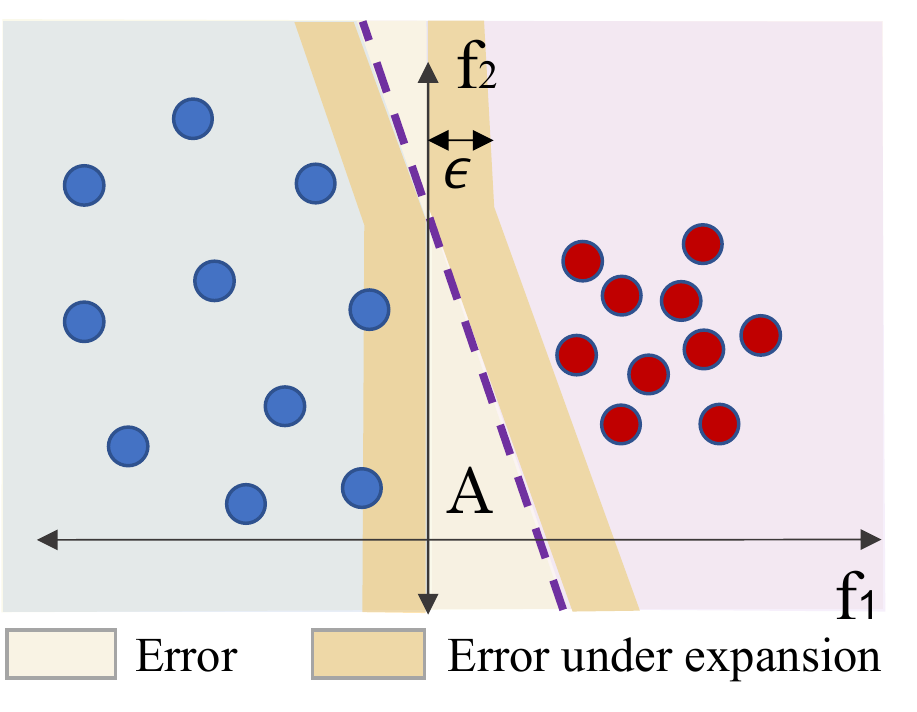}
  \vspace{-0.12in}
  \caption{Concentration illustration.}
  \label{fig:concentrationFig}
  \end{minipage}
  \begin{minipage}{0.56\textwidth}
  \centering
  \includegraphics[width =0.9\textwidth]{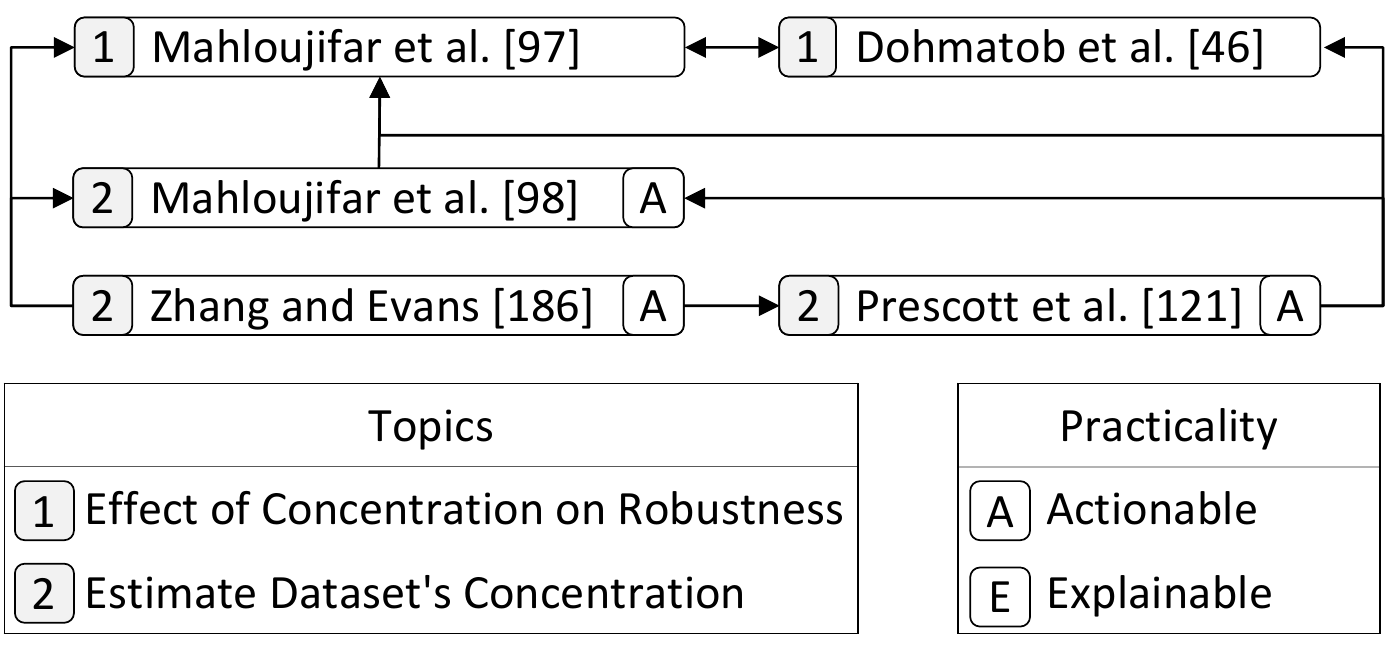} 
  \vspace{-0.1in}
  \caption{Papers discussing concentration. }
  \label{fig:concentration}
  \end{minipage}
  \vspace{-0.1in}
\end{figure*}

Fig.~\ref{fig:concentrationFig} shows how the concentration of measure phenomenon can be used to determine the classification error after adversarial perturbation.
By modeling the classification error set as measurable set $A$
and adversarial errors from perturbation budget $\epsilon$ as $A_{\epsilon}$,
one can relate the concentration of the data to the minimum adversarial risk for any imperfect classifier with error rate $\mu(A) \geq \alpha$.
Using this formulation, a dataset being highly concentrated implies that, for some non-zero initial error, the minimum adversarial risk from an $\epsilon$-expansion on the error set is very large.
We refer to such datasets as datasets with low \emph{intrinsic robustness}~--
a measure that represents the maximal achievable robustness for any classifier on a dataset.

Fig.~\ref{fig:concentration} shows the papers that relate data concentration to adversarial robustness.
They can roughly be divided into:
\roundrect{1} papers discussing the effect of concentration on robustness and
\roundrect{2} papers proposing techniques to estimate robustness through calculating concentrations.

\vspace{0.05in}
\noindent
\roundrect{1} {\bf Effects of Concentration}.
A number of papers prove the inevitability of adversarial examples
using the concentration of measure phenomenon. 
In particular, Dohmatob~\cite{Dohmatob:ICML:2019} investigates datasets
conforming to uniform, Gaussian, and several other distributions
that satisfy $W_2$ transportation-cost inequality~\cite{Talagrand:1996}.
The author proves that data distributions satisfying such inequality have high concentration,
which results in a rapid robustness decrease, beyond a critical perturbation size~--
a value that depends on the standard error of the classifier and the natural noise level of the dataset, which, in turn,
is defined as the largest variance in the case of Gaussian distribution.
Even though \mnist might not satisfy the $W_2$ transportation-cost inequality,
the author experiments with this dataset, observing a sudden drop in robustness
as the perturbation size increases.
As such, the author suggests that the \mnist dataset may also have high concentration and be
governed by the concentration of measure phenomena.

Mahloujifar et al.~\cite{Mahloujifar:Diochnos:Mahmoody:AAAI:2019} focus on a collection of data distributions with high concentration called L\'evy families~\cite{Levy:1951}, which include unit sphere, unit cube, and  isotropic n-Gaussian
(i.e., Gaussian with independent variables with the same variance).
The authors prove that classifiers trained on such highly-concentrated data distributions admit adversarial examples with perturbation
$\mathcal{O}(\sqrt{d})$  for dimensionality $d$.
This implies that a relatively small perturbation can mislead models trained on these data distributions with high dimensional inputs.

\vspace{0.05in}
\noindent
\roundrect{2} {\bf Estimating Robustness Through Concentration}.
Several approaches utilize the connection between concentration and adversarial risk to estimate 
the intrinsic robustness of datasets by calculating their concentrations.
Mahloujifar et al.~\cite{Mahloujifar:Zhang:Mahmoody:Evans:NeurIPS:2019} are the first to propose an approach for estimating
dataset concentration using subsets of samples.
Specifically, the authors propose a technique that searches for the minimum expansion set based on a collection of subsets carefully chosen according to the perturbation norm (e.g., a union of balls for $L_2$ norm).
They prove that
the estimated concentration value converges to the true value for the underlying distribution as the sample size and the quality/representativeness of the chosen subsets increase.
The authors apply their approach for estimating the maximum achievable robustness for the \mnist and \cifarten datasets,
observing a gap between the derived theoretical values and values observed empirically by the state-of-the-art models.

In follow-up work, Prescott et al.~\cite{Prescott:Zhang:Evans:ICLR:2021} propose
an alternative approach to estimate concentration based on half space expansion using \emph{Gaussian Isoperimetric Inequality}
for the $L_2$ norm~\cite{Borell:InvMath:1975}.
The authors further generalize their results to $L_p$ norms, where $p \geq 2$.
Compared with Mahloujifar et al.~\cite{Mahloujifar:Zhang:Mahmoody:Evans:NeurIPS:2019},
their approach yields higher achievable robustness on \mnist and \cifarten, revealing a larger gap between the theoretical robustness and the state-of-the-art.
As the theoretically achievable robustness derived from a concentration perspective is shown to be high,
the authors suggest that factors other than concentration may contribute to this gap.

Zhang and Evans~\cite{Zhang:Evans:ICLR:2022} assume access to information about label uncertainty,
i.e., a function that assigns the level of label uncertainties for any data point.
Such a function can use, e.g., labeling results from multiple human annotators or confidence scores from an ML classifier.
The authors suggest that considering regions with high label uncertainty can guide the concentration estimation
as these are the regions where a classifier is more likely to make mistakes and be vulnerable to attacks.
They thus propose an approach to estimate concentration by identifying the smallest set after
$\epsilon$-expansion with an average uncertainty level greater than a pre-set value. 
The evaluation results show that the maximum achievable robustness estimated with their approach
is closer to the robustness values observed for CNN models on the \cifarten dataset than in
any of the aforementioned works, implying that the
room for improvement is smaller than assumed earlier.

\subsection{Label Quality}
\label{sec:results-label-quality}

\begin{figure*}[h]
	\vspace{-0.12in}
	\begin{minipage}[t]{0.55\textwidth}
		\includegraphics[valign=t, width=0.95\linewidth]{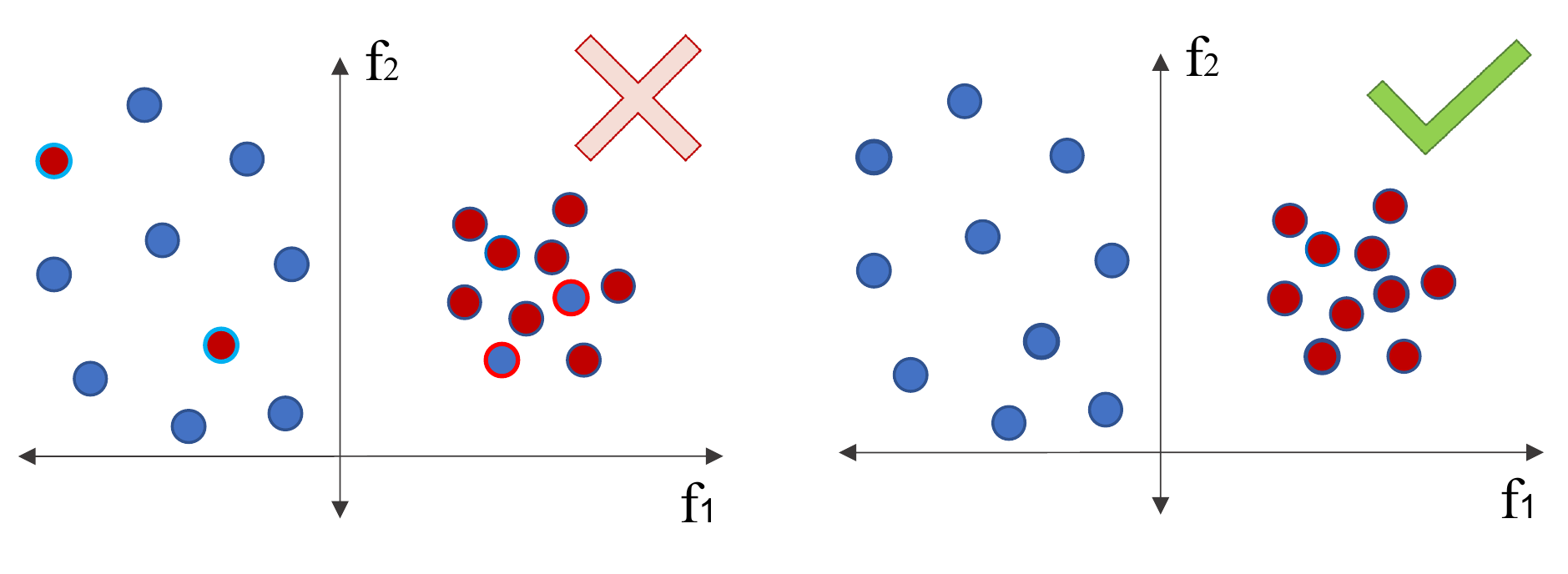}
		\vspace{-0.18in}
		\caption{Label noise illustration.}
		\label{fig:label_quality_guideline}
	\end{minipage}
	\begin{minipage}[t]{0.4\textwidth}
		\centering
		\includegraphics[valign=t, width=0.9\textwidth]{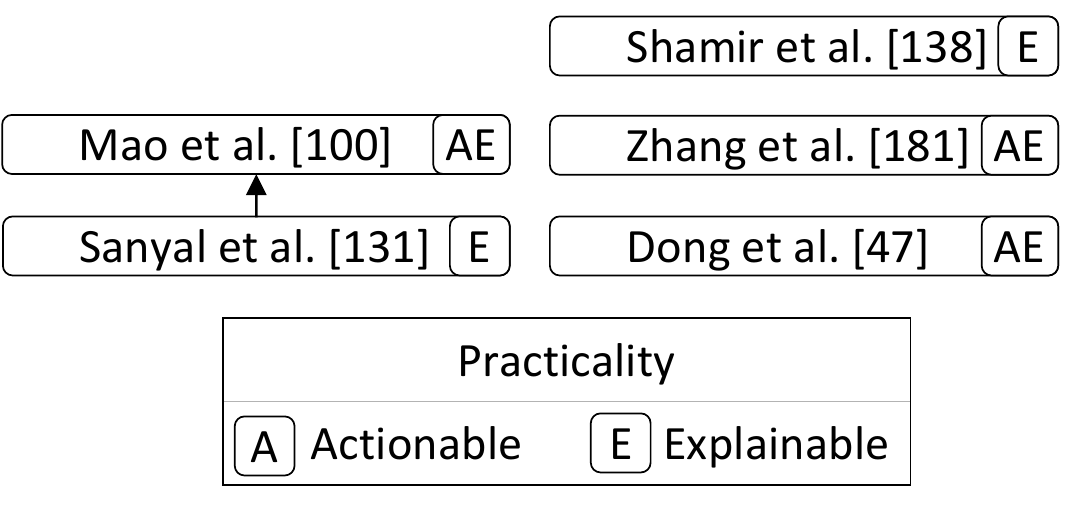} 
		\vspace{-0.05 in}
		\caption{Papers discussing label quality.}
		\label{fig:label}
	\end{minipage}
	\vspace{-0.1in}%
\end{figure*}

\emph{Label quality} refers to the correctness and informativeness of the set of labels assigned to a training dataset.
Label correctness or, inversely, the presence of inaccurate labels (shown as the highlighted dots on the left-hand side of Fig.~\ref{fig:label_quality_guideline}) is typically referred to as \emph{label noise}.
The granularity of the labels is typically referred to as \emph{label informativeness}.
Papers that  discuss the relationship between label quality and model robustness are outlined in Fig.~\ref{fig:label}.

Mao et al.~\cite{Mao:Gupta:Nitin:Ray:Song:Yang:Vondrick:ECCV:2020} show that training a model
simultaneously for multiple tasks, e.g., to simultaneously locate and estimate the distance of objects in images
(an approach also referred to as \emph{multi-task learning}),
improves robustness.
This is because in multi-task learning, a model learns a shared feature representation by training on data
with labels from several tasks.
As a result, perturbations required to attack multiple tasks at the same time, e.g.,
to sabotage an autonomous driving system by misleading the model in both object identification and distance estimation,
cancel each other out.
While the authors prove that the model robustness to adversarial attacks is proportional to the number of tasks
that it is trained on,
the benefits of multi-task learning disappear when the concurrently trained tasks are highly correlated with each other,
as it reduces the chances for the perturbations to cancel each other.
The authors further show that training with multiple tasks also improves model robustness against single-task attacks.

\begin{wrapfigure}{r}{0.4\textwidth}
  \vspace{-0.25in}
    \begin{minipage}[t]{0.45\textwidth}
	 \centering
   	 \subcaptionbox{Overfit}{
     \includegraphics[width=0.4\linewidth]{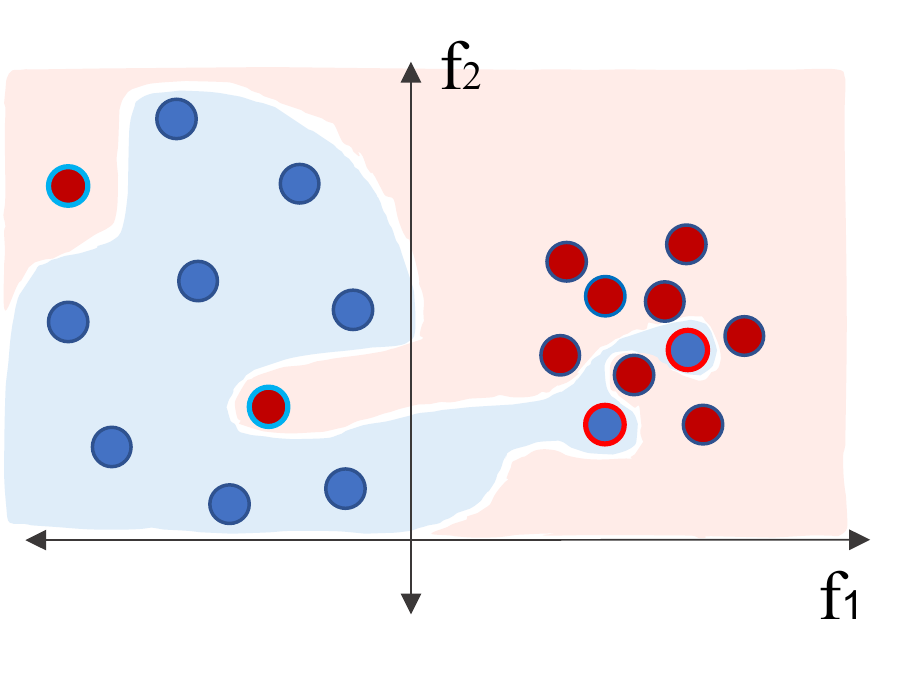}
     \vspace{-0.15in}
   	 }
     \subcaptionbox{Not overfit.}{
     \includegraphics[width=0.4\linewidth]{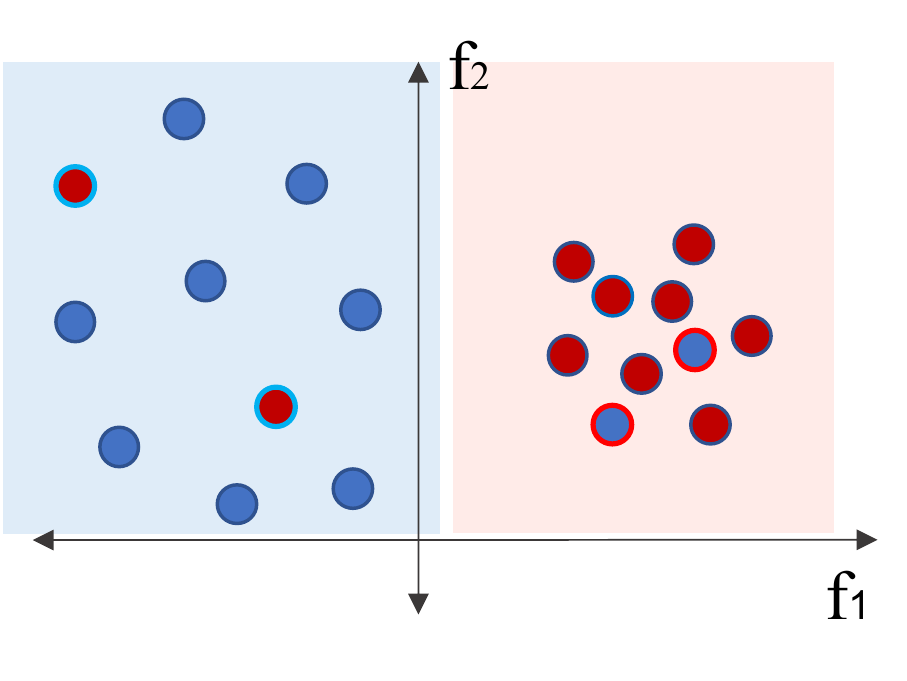}
     \vspace{-0.15in}
     }
     \vspace{-0.15in}
     \caption{Influence of overfitting to label noise.}
     \label{fig:label_noise_influence}
    \end{minipage}
    \vspace{-0.15in}
\end{wrapfigure}
Sanyal et al.~\cite{Sanyal:Dokania:Kanade:Torr:ICLR:2021} hypothesize that label noise and coarse labels are the reasons for adversarial vulnerability.
The authors prove that, given a large training set with random label noise, any classifier that overfits to that set is likely to be vulnerable to adversarial attacks.
This is because overfitting leads to overly complex decision boundaries that leave more room for attacks, as illustrated in  Fig.~\ref{fig:label_noise_influence}.

The authors also demonstrate that adversarial risk increases as the level of label noise increases.
Defense mechanisms, such as early stopping and adversarial training, enhance robustness by preventing models from overfitting to noisy samples.
In the absence of label noise, using coarse labels
(e.g., one label for the entire class of dogs rather than labels for each individual dog breed)
results in ``sub-optimal'' latent feature representations and also contributes to the adversarial vulnerability.
A similar result was also shown by Shamir et al.~\cite{Shamir:Safran:Ronen:Dunkelman:ArXiv:2019} 
for high-dimensional data and piecewise linear classifiers, such as DNN with ReLU activation: 
the number of perturbations required to generate successful adversarial examples is proportional to the number of classes, 
making models trained on data with fine-grained labels more robust. 

Dong et al.~\cite{Dong:Liu:Shang:NeurIPS:2022} show that label noise is an inherent part of adversarial training 
as labels assigned to the adversarial examples may not always match their ``correct'' labels. 
The authors show that the amount of label noise introduced in adversarial training is proportional 
to the perturbation radius and the confidence of the model prediction for an adversarial example. 
They further propose to mitigate this issue by filtering low-confidence examples generated with a high perturbation radius and 
demonstrate that their approach can achieve higher robust accuracy than in standard adversarial training for 
\cifarten, \cifarhundred, and \tinyset. 

Zhang et al.~\cite{Zhang:Jiang:Hou:Wei:Han:Li:Cheng:TIP:2021} utilize soft labels, 
i.e., labels that capture the probability of a data point belonging to a certain class, 
to learn the relationship between classes. This encourages a model 
to learn representations that group similar samples together, thus 
increasing intra-class density and, as a result, increasing robustness.

\subsection{Domain-Specific}
\label{sec:results-domain-specific}

\begin{figure*}[b]
  \centering
  \vspace{0.05in}
  \begin{minipage}{0.37\textwidth}
	  \centering
	  \vspace{0.05in}
	  \includegraphics[valign=t, width =\textwidth]{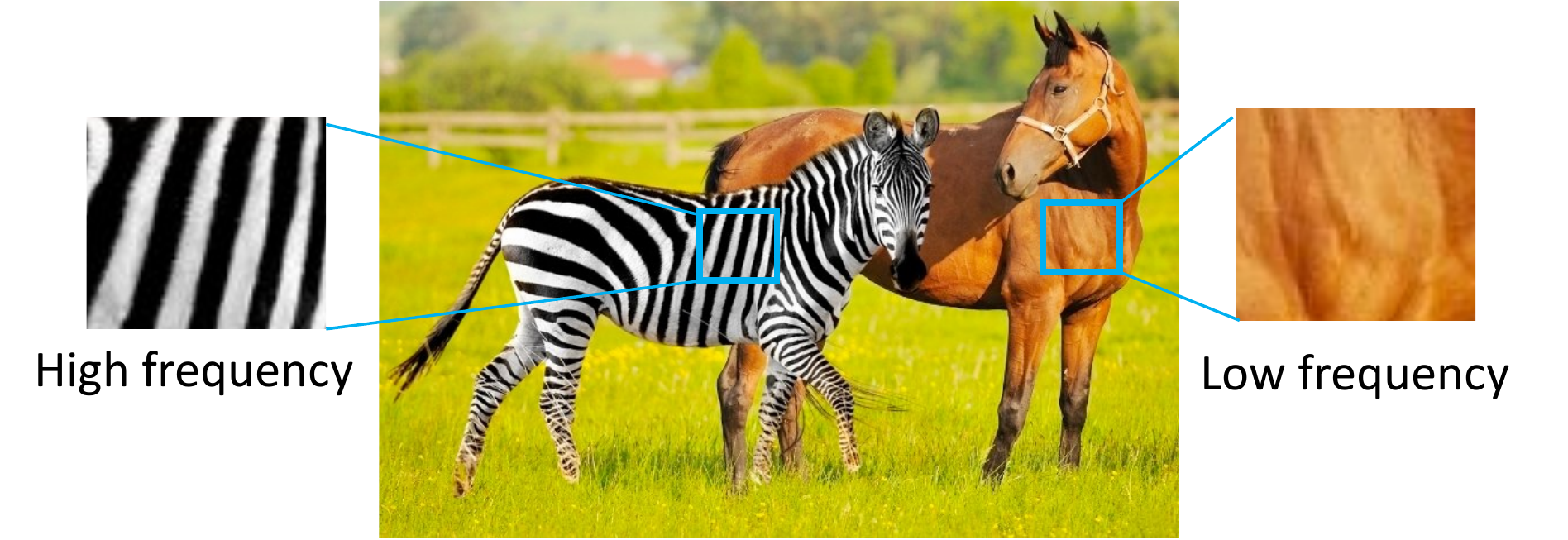}
	  \caption{Image frequency.}
 	  \vspace{-0.13in}
	  \label{fig:image_freq}
  \end{minipage}
  \begin{minipage}{0.62\textwidth}
		\centering
		  \vspace{-0.2in}
		\includegraphics[valign=t, width =\textwidth]{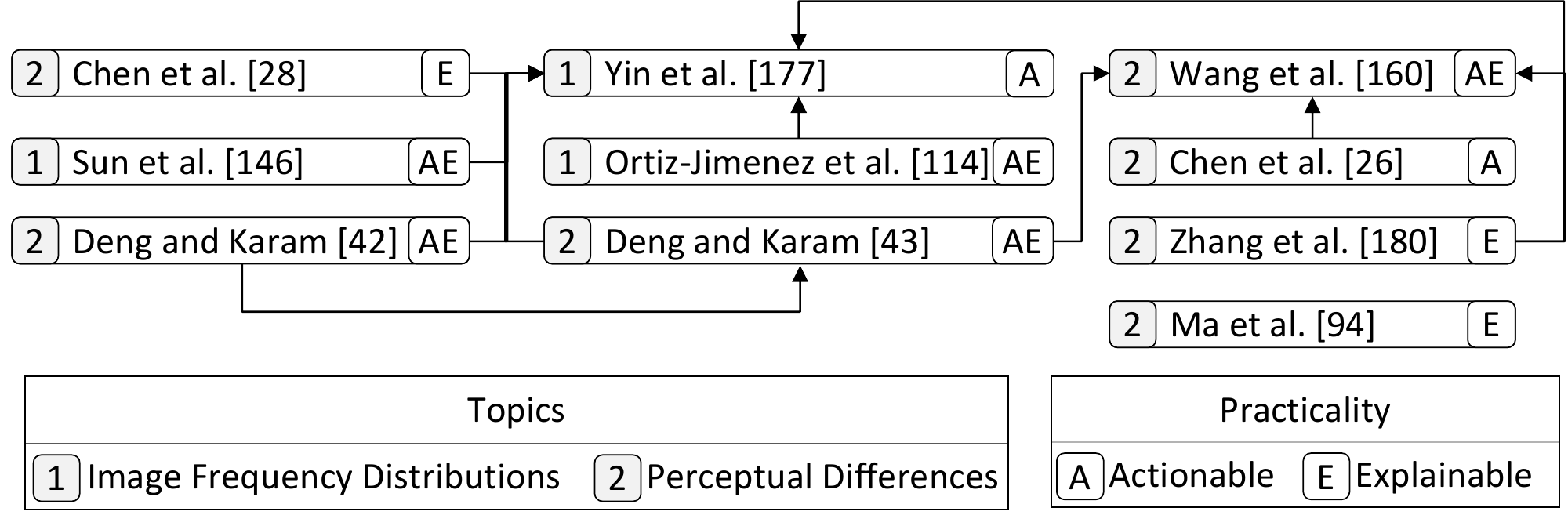} 
		\vspace{-0.07in} 
		\caption{Papers discussing domain-specific properties. }
		\label{fig:domain}
  \end{minipage}
  \vspace{-0.05in}
\end{figure*}

Papers in this category provide insights into the correlation between domain-specific data properties and adversarial robustness.
Among our collected papers, all the domain-specific studies focused on the same topic: understanding the adversarial vulnerabilities of image classifiers based on image \emph{frequency}~--
how fast the intensity of pixel values changes with respect to space (i.e., images with intensive color changes have high-frequency).
As shown in Fig.~\ref{fig:image_freq}, the skin of a zebra has higher image frequency than a horse, because of the black-and-white stripes.
Papers studying image frequency are listed in Fig.~\ref{fig:domain}.
They can be roughly divided into:
\roundrect{1} papers discussing the influence of frequency distribution on the model adversarial robustness,
and~\roundrect{2}~papers explaining adversarial vulnerabilities using perceptual differences between humans and models. 

\vspace{0.05in}
\noindent
\roundrect{1} {\bf Effects of Image Frequency}.
Wang et al.~\cite{Wang:Wu:Huang:Xing:CVPR:2020} attribute adversarial examples to the perceptual differences between humans and ML models in frequency ranges: 
while high-frequency components are not visible to the human eye and 
humans mostly classify images based on low-frequency components, 
models can utilize both 
ranges, allowing them to create adversarial examples imperceptible to humans in high-frequency ranges.
The authors propose to use smoother convolutional filters to reduce a model's attention to high-frequency components, 
thus improving robust generalization.

Deng and Karam~\cite{Deng:Karam:ECCVWorkshops:2020, Deng:Karam:TIP:2022} also acknowledge and 
experimentally demonstrate the difference between the human and model perspectives. 
The authors focus on universal adversarial perturbations, 
i.e., input-agnostic perturbations that cause label change for most inputs sampled from the data distribution,
noting that such perturbations can be created by exploiting human sensitivity to different frequency ranges.
They use this observation to generate high-success adversarial example by utilizing the Just-Noticeable-Difference~\cite{Albert:Heidi:HVVPD:1997,Hontsch:Karam:TIP:2002} metric, 
which approximates the maximal imperceptible perturbation along different frequency ranges.
Through their experiments, the authors demonstrate that humans are generally more sensitive 
to changes in low-frequency rather than high-frequency components.

Zhang et al.~\cite{Zhang:Benz:Karjauv:Kweon:AAAI:2021} 
observe that successful universal adversarial perturbations are dominated by perturbations in high-frequency components. 
The authors also show that such perturbations cause more ``distinct'' changes in  images with more low-frequency components,
making them more susceptible to universal adversarial perturbations.
The same observation is supported by 
Chen et al.~\cite{Chen:Ren:Yan:NeurIPS:2022}, who use an explainability technique to attribute predictions to data:  
as the adversarial training process is focusing on high-frequency components, 
adversarially trained models rely more heavily on low-frequency components while 
standard models utilize both. 
Yet, the authors show that using high-frequency components is necessary to accurately predict 
certain classes. As such, when the model learns to prioritize low-frequency components through adversarial training, 
the accuracy for high-frequency images is compromised.

Yin et al.~\cite{Yin:Lopes:Shlens:Cubuk:Gilmer:NeurIPS:2019} outline another disadvantage 
of techniques that generate perturbations with high-frequency components, 
such as adversarial training and Gaussian data augmentation techniques:
they result in models more vulnerable to low-frequency perturbations.
Similarly, Ortiz-Jimenez et al.~\cite{Ortiz-Jimenez:Modas:Moosavi-Dezfooli:Frossard:NeurIPS:2020}
show that CNN models tend to have smaller margins along low-frequency vs. high-frequency ranges and 
that adversarial training results in significantly larger margins against high-frequency perturbations, 
making the model more vulnerable to low-frequency perturbations.

Sun et al.~\cite{Sun:Mehra:Kailkhura:Chen:Hendrycks:Hamm:Mao:ECCV:2022} also observe the issue 
of low robustness against low-frequency perturbations. Yet, in this case, it is caused by 
using an inference-time technique (randomized smoothing) rather than by adversarial training. 
The aforementioned works propose to mitigate these issues by increasing the diversity of frequency distribution in training data.

\vspace{0.05in}
\noindent
\roundrect{2} {\bf Effects of Other Image Properties}.
Unlike the work above that focuses on image frequency,
Chen et al.~\cite{Chen:Peng:Ma:Li:Du:Tian:ICCV:2021} posit that the adversarial vulnerability of CNNs
results from their over-reliance on amplitude information of images~--
the magnitude of the different frequencies in the image.
The authors show that replacing the amplitude information of an image with information from another image can successfully mislead CNNs but not humans, who rather rely on phase information~-- the locations of the features, to recognize objects.
Based on this observation, the authors propose to strengthen CNNs' attention to phase information through a data augmentation technique that fuzzes amplitude while preserving the same phase information.

Focusing on medical image classification, Ma et al.~\cite{Ma:Niu:Gu:Wang:Zhao:Bailey:Lu:PR:2021} show that image classifiers pre-trained on natural image datasets, such as ImageNet, have lower adversarial robustness when applied to medical images. 
The authors attribute this degradation to the distinct biological textures and relative simplicity of medical images compared to natural ones. 
They argue that models designed for natural images have a high likelihood of over-fitting to 
noise present in non-lesion areas and,
consequently, a higher susceptibility to adversarial perturbations in those regions.

\subsection{Summary of Results}
\label{sec:results-summary}

Overall, the surveyed papers are mostly in agreement on how each of the identified data properties influences adversarial robustness. The main findings are given below. 

\vspace{0.05in}
\noindent {\bf Number of Samples.} More training samples are needed for robust than for standard generalization.
For a variety of training setups (i.e., different types of classifiers and data distributions), the number of training samples required to achieve robust generalization is proportional to the dimensionality of the training data. Unlabeled samples or generated data can be used to fulfill the need for more samples needed for robust generalization, i.e., to close the sample complexity gap.
Class imbalance, i.e., having an imbalanced number of samples across different classes, hurts robust generalization due to the model bias towards over-represented samples. 
Re-weighted loss functions can be used to mitigate this model bias and thereby improve robustness.

\vspace{0.05in}
\noindent
{\bf Dimensionality.} Dimensionality captures the size of the feature set.
Higher dimensionality correlates with
higher adversarial risk, worse standard-to-adversarial risk trade-off, difficulty in robustness certification, and difficulty in applying common defense techniques.
This is because adversarial attacks can exploit the excess dimensions to construct adversarial examples. 

\vspace{0.05in}
\noindent
{\bf Distribution.} Some data distributions are more robust than others, e.g., mixtures of Bernoulli distributions are more robust than mixtures of Gaussian distributions.
Learning feature representations that resemble robust distributions can improve robustness.

\vspace{0.05in}
\noindent
{\bf Density.} Density reflects the closeness of samples in a particular bounded region (intra-class distance).
Adversarial examples are commonly found in low-density regions of data, where samples are far apart from each other.
This is because models cannot accurately learn decision boundaries near low-density regions due to the small number of samples available. As such, high data density for each class correlates with lower adversarial risk.

\vspace{0.05in}
\noindent
{\bf Separation.} Separation characterizes the distance of samples from different classes to each other (inter-class distance).
Greater separation between classes decreases adversarial risk as it is harder to generate perturbations that 
will cross the decision boundaries between classes.
Most papers that provide techniques for improving separations do so by modifying the loss function to learn a latent data representation with higher separation.
They also ensure that this increase in separation does not come at the expense of decreasing density, as these two concepts are closely related.

\vspace{0.05in}
\noindent
{\bf Concentration.} Given a function defined over a non-empty set,
concentration (from the phenomenon of concentration of measure~\cite{Talagrand:1996:AnnalsProbability})
is the minimum value of the function after expanding the input set by $\epsilon$ in all dimensions.
For example, expanding the set of misclassified samples by a certain $\epsilon$ gives a set of possible samples that can be
misclassified with an $\epsilon$-size perturbation (candidate adversarial examples).
Concentration, in this case, measures the minimal possible size of this set, which provides the upper bound of the achievable model robustness.
As some datasets tend to exhibit inherently high concentration,
e.g., datasets that lie on unit hypersphere~\cite{Mahloujifar:Diochnos:Mahmoody:AAAI:2019},
achieving high robust generalization is harder for these datasets.
The impact of high concentration on adversarial robustness is further magnified for high-dimensional data.

\vspace{0.05in}
\noindent
{\bf Label Quality.} High label noise correlates with higher adversarial risk.
Label noise can be automatically generated as part of adversarial training 
because perturbation can change the semantics of the perturbed data samples. 
Labels thus have to be carefully examined and rectified.
More specific labels, e.g., ``cat'' and ``dog'' instead of ``animal'', are more robust than coarse labels, as they allow the model to extract more distinct features.
Learning for different tasks concurrently, e.g., to simultaneously locate and estimate the distance of objects in images,improves the robustness of the learned models, as the model can utilize the information from multiple sources of data.

\vspace{0.05in}
\noindent
{\bf Domain-Specific.} Image frequency, i.e., the rate of change in pixel value, is shown to be correlated with robustness.
As humans perceive differences in images by focusing on low-frequency components 
whereas models consider both low and high-frequency components,  
high-frequency perturbations can ``fool'' models while being imperceptible to humans. 
Adversarial training tends to focus on high-frequency components and, as a result,
adversarially trained models rely more heavily on low-frequency components. 
While this helps ensure that models focus on low-frequency components as well, it introduces new issues
related to clean and robust accuracy. 
Encouraging a diverse distribution of frequencies in training data results in lower adversarial risk.

\section{Implications and Future Work}
\label{sec:discussion}

We now summarize the practical implications of our findings, 
extracting guidelines and techniques for improving adversarial robustness by 
manipulating data and learning procedures, 
as well as techniques for estimating best-case robustness for particular data (Section~\ref{sec:discussion-practical}). 
We then outline the gaps in research and possible directions for future work (Section~\ref{sec:discussion-future}).

\subsection{Practical Implications}
\label{sec:discussion-practical}

\noindent {\bf Improving Robustness by Manipulating Data.}
We identified two main directions for improving robustness through data manipulation techniques:

\vspace{0.02in} 
\emph{1. Increasing datasets size and diversity.}
Collecting a large number of real labeled samples required to train adversely robust models could be challenging. 
Our survey shows that one could employ cheaper data collection alternatives, such as unlabeled or generated samples, 
and then use semi-supervised learning techniques, such as pseudo-labeling, to learn from such samples~\cite{Uesato:Alayrac:Huang:Stanforth:Fawzi:Kohli:NeurIPS:2019,Carmon:Raghunathan:Schmidt:Duchi:Liang:NeurIPS:2019,
	Najafi:Maeda:Koyama:Miyato:NeurIPS:2019,Gowal:Rebuffi:Wiles:Stimberg:Calian:Mann:NeurIPS:2021}.	
Moreover, data dimensionality can be used to guide the number of augmented/generated samples 
required to improve robustness~\cite{Rajput:Feng:Charles:Loh:Papailiopoulos:ICML:2019}.
Our survey also shows that using fine labels can lead to more robust models~\cite{Sanyal:Dokania:Kanade:Torr:ICLR:2021, Mao:Gupta:Nitin:Ray:Song:Yang:Vondrick:ECCV:2020}.
	
When collecting data, increasing the diversity of samples, 
e.g., including samples from both low and high frequency ranges, in case of images~\cite{Yin:Lopes:Shlens:Cubuk:Gilmer:NeurIPS:2019,Sun:Mehra:Kailkhura:Chen:Hendrycks:Hamm:Mao:ECCV:2022}, is necessary.  
Furthermore, data properties, such as intrinsic dimensionality and separation, are defined over the 
underlying data manifold, assuming the knowledge of the entire data distribution; 
yet, for practical purposes, these properties are estimated and enforced for concrete datasets. 
This can lead to incorrect expectations / estimations in cases where training samples do not 
correctly approximate the underlying distribution.
For example, producing a well-separated subset of training data by selecting samples can
improve robustness in non-parametric models~\cite{Yang:Rashtchian:Wang:Chaudhuri:AISTATS:2020,Wang:Jha:Chaudhuri:ICML:2018},
but this strategy does not work for more complex classifiers, 
as sample selection does not make the underlying distribution more separated.
Paying close attention to the representativeness of the training dataset can help mitigate this issue.

\vspace{0.02in}
\emph{2. Cleaning and transforming data.}
Besides collecting large and diverse datasets, there are also multiple ways to improve the quality of existing data.
For example, one can use domain knowledge and/or 
automated feature selection techniques, such as mutual information gain~\cite{Shannon:1949}, 
to remove features with low variance, which helps reduce dimensionality and 
improve robustness~\cite{Izmaliov:Sugrim:Chadha:McDaniel:Swami:MILCOM:2018}.
Dimensionality reduction techniques, which project samples onto lower-dimensional spaces, e.g., PCA~\cite{Jolliffe:2002}, 
also reduce adversarial vulnerability~\cite{Awasthi:Jain:Rawat:Vijayaraghavan:NeurIPS:2020}.
Another approach, 
which relies on using graph theory~\cite{Fan:1997} to transform datasets into graphs that use pair-wise 
distances as edges, 
can help generate 
features that maximize the distance between samples from different classes, resulting in 
well-separated feature representations~\cite{Garg:Sharan:Zhang:Hu:NeurIPS:2018}. 

Numerous learning-based techniques use loss functions engineered to optimize for more complex data properties. 
While common loss functions, such as Softmax Cross-Entropy loss~\cite{Pang:Xu:Dong:Du:Chen:Zhu:ICLR:2020},
can result in sparse representations (i.e., with low class density), 
more advanced loss functions can utilize inter-class or intra-class distances to learn feature representations that cluster points around class centers and/or promote separation between classes~\cite{Pang:Xu:Dong:Du:Chen:Zhu:ICLR:2020,Mustafa:Khan:Hayat:Goecke:Shen:Shao:ICCV:2019,Mygdalis:Pitas:PR:2022,Bui:Le:Zhao:Montague:deVel:Abraham:Phung:ECCV:2020,Pang:Du:Zhu:ICML:2018,Cheng:Zhu:Zhang:Liu:PR:2023,Yang:Feng:Du:Du:Xu:ICDM:2021}, resulting in increased robustness. 
Producing latent representations that resemble symmetric Gaussian distributions, 
which have been shown to increase robustness~\cite{Pang:Du:Zhu:ICML:2018,Wan:Chen:Yu:Wu:Zhong:Yang:TPAMI:2022}, 
could also be beneficial.

\vspace{0.02in}
\noindent 
{\bf Improving Robustness by Manipulating Learning Procedures.}
Knowledge about the properties of a dataset can further help improve robustness 
by manipulating the learning procedure:

\vspace{0.02in}
\emph{1. Model selection and configuration.}
When deciding on a model to train for a particular dataset, probabilistic models, 
such as Bayesian Neural Networks~\cite{Carbone:Wicker:Laurenti:Patane:Bortolussi:Sanguinetti:NeurIPS:2020}, 
are shown to be more robust given ``undesired'' data properties, such as high dimensionality~\cite{Carbone:Wicker:Laurenti:Patane:Bortolussi:Sanguinetti:NeurIPS:2020}.
For datasets suffering from class imbalance, 
scale-invariant classifiers~\cite{Wang:Wang:Zhou:Ji:Gong:Zhou:Li:Liu:CVPR:2018,Pang:Yang:Dong:Xu:Zhu:Su:NeurIPS:2020}
can be used to reduce the robustness bias among the classes~\cite{Wu:Liu:Huang:Wang:Lin:CVPR:2021}.

Data properties are also useful in determining the configuration of models. 
For example, data dimensionality can guide the number of parameters necessary to train 
robust parametric models, such as DNNs, 
and the type of regularization used to constrain their weights (e.g., $L_1$ norm instead of the commonly used $L_2$ or $L_\infty$ norms)~\cite{Li:Jin:Zhong:Hopcroft:Wang:NeurIPS:2022, Yin:Kannan:Bartlett:ICML:2019}.
Dimensionality can also be useful in determining hyper-parameters in non-parametric models, 
e.g., $k$ in $k$-NNs~\cite{Wang:Jha:Chaudhuri:ICML:2018}.

\vspace{0.02in}
\emph{2. Training and inference procedures.}
During training, learning to perform different tasks concurrently can lead to more robust models~\cite{Sanyal:Dokania:Kanade:Torr:ICLR:2021, Mao:Gupta:Nitin:Ray:Song:Yang:Vondrick:ECCV:2020}.
When faced with imbalanced datasets, loss functions that assign different weights to samples from different classes can be used to mitigate the influence of class imbalance on robustness~\cite{Wu:Liu:Huang:Wang:Lin:CVPR:2021}.

At inference time, sample purification methods, such as randomized smoothing~\cite{Cohen:Rosenfeld:Kolter:ICML:2019}, 
can be used to defend against adversarial attacks by adding noise to cancel the effects of carefully-crafted perturbations.
Purification can also be achieved by using generative models to move samples from low-density to high-density regions, 
i.e., from low-confidence to high-confidence regions, 
before making predictions~\cite{Song:Kim:Nowozin:Ermon:Kushman:ICLR:2017}. 

\vspace{0.02in}
\noindent 
{\bf Estimating Robustness.}
Data-related metrics can be used to evaluate inherent robustness limitations of datasets.
In particular, several approaches~\cite{Mahloujifar:Zhang:Mahmoody:Evans:NeurIPS:2019,Prescott:Zhang:Evans:ICLR:2021,Zhang:Evans:ICLR:2022} empirically calculate concentration and use it to obtain the minimum adversarial risk.
Similarly, the optimal transport distance among classes can be used to estimate the risk for 
concrete datasets~\cite{Bhagoji:Cullina:Mittal:NeurIPS:2019, Pydi:Jog:ICML:2020, Pydi:Jog:NeurIPS:2021}.
Estimating the minimum adversarial risk serves as an indication of achievable robustness, which can 
inspire trying out alternative data collection, processing, and learning strategies, in cases when 
the observed robustness is significantly less than the achievable value.

%==============================================================================

\subsection{Knowledge Gaps and Future Research Directions}
\label{sec:discussion-future}

\vspace{0.02in}
\noindent
{\bf Empirical Evaluation.}
All but the ten papers outlined in Section~\ref{sec:results-domain-specific} study domain-agnostic data properties.
Yet, the majority of the papers we surveyed conduct experimental evaluations on image datasets only.
Applicability of the findings and of the proposed approaches to other domains, with different forms of data,
may need further investigation.
For example, for datasets with binary features, which are commonly used in malware detection, one cannot arbitrarily
change feature values to reduce the distance between samples.
This further implies that common distance metrics used to model adversaries in the image domain, such as $L_2$ and $L_\infty$, fail to accurately capture the adversarial threat level in such domains.
Hence, future work applying, adapting, and evaluating the proposed metrics and techniques in other domains and data types is needed.

\vspace{0.02in}
\noindent
{\bf Simplified Problem Setup.}
Several studies use a simplified problem setup, 
e.g., pure Gaussian data distributions, to provide formal proofs related to the studied phenomenon.
While such work helps advance knowledge and our understanding of the effects of data on adversarial robustness,
additional work that investigates the generalizability of the findings on realistic datasets used in practice is needed.
For example, assuming uniform data properties, e.g., same distribution, density, and level of label noise,
for all classes on the training data greatly simplifies the proofs, but is not common in reality.
Likewise, considering only binary classification simplifies calculations of data separation, which can be calculated by measuring the distance between the two classes. Yet, in a multi-class setting, one needs to consider the proximity of data points from multiple classes.

Furthermore, most papers only consider a white-box attack setting, which might not be realistic in many practical scenarios.
Even though a white-box setting makes it possible to model the worst-case adversary and to provide better robustness guarantees, it may result in overly pessimistic findings, i.e., some data transformations may be robust against black-box attacks while still being vulnerable to white-box attacks.
Thus, future works might look into the impact of data properties on the different types of attack scenarios.

\vspace{0.02in}
\noindent
{\bf Data and Model Interplay.}
Even when the training data is optimal for robustness, a sub-optimal training method can lead to adversarial vulnerability~\cite{Richardson:Weiss:JMLR:2021}.
For example, adversarial vulnerability may arise when the complexity of the classifier does not match the complexity of the data, e.g., CNNs may achieve lower robustness due to their complexity than simpler models, such as Kernel-SVMs,
on symmetrical data with well-separated means and similar variances~\cite{Richardson:Weiss:JMLR:2021}.
To alleviate such problems, a few papers propose to select, improve, or optimize classifiers based
on the dimensionality of data~\cite{Wang:Jha:Chaudhuri:ICML:2018,
Yin:Kannan:Bartlett:ICML:2019,
Carbone:Wicker:Laurenti:Patane:Bortolussi:Sanguinetti:NeurIPS:2020}.
Similar work that looks at other properties of data, such as separation and density, could be of value.
Future works could also explore strategies for determining whether the input data (vs. the model itself) is the dominant cause \mbox{of adversarial vulnerability}.

\vspace{0.02in}
\noindent
{\bf Training Dataset vs. Data Distribution.}
As discussed in Section~\ref{sec:discussion-practical}, representative training data is necessary to truly estimate 
the properties of the entire underlying data manifold. 
Yet, in practice, collecting representative datasets is challenging and data properties, such as separation and density,
are often evaluated on concrete datasets.  
More research on how to assess the gap between the observed and the ``real'' data properties could be valuable. 

\vspace{0.02in}
\noindent
{\bf Efficient Selection of Unlabeled Samples.}
The need for large-scale training data is growing, aided by the increase in compute power and efficient algorithms.
However, manually labeling all samples remains impractical in several domains.
A number of works~\cite{Uesato:Alayrac:Huang:Stanforth:Fawzi:Kohli:NeurIPS:2019, Carmon:Raghunathan:Schmidt:Duchi:Liang:NeurIPS:2019, Najafi:Maeda:Koyama:Miyato:NeurIPS:2019} showed that 
using unlabeled and generated samples can help to fill the sample complexity gap for developing more robust models.
Xing et al.~\cite{Xing:Song:Cheng:NeurIPS:2022} assess the quality of generated samples 
w.r.t. ones from an ideal generator (a.k.a. real unlabeled samples).
However, as the quality of unlabeled samples is undetermined 
(e.g., they can contain a cat and a dog in the same picture when performing a cat vs. dog binary classification),
research on the effects of unlabeled sample quality on model robustness in needed.
Such research can further facilitate the development of techniques that prioritize particular 
unlabeled samples in robust model training.

\vspace{0.02in}
\noindent
{\bf Interdependence of Properties.}
Only a few works in our collected literature consider multiple data properties simultaneously or establish interdependence of data properties.
For example, Wang et al.~\cite{Wang:Jha:Chaudhuri:ICML:2018} and
Rajput et al.~\cite{Rajput:Feng:Charles:Loh:Papailiopoulos:ICML:2019}
find that the number of samples and dimensionality collectively influence the performance of the resulting model.
Sanyal et al.~\cite{Sanyal:Dokania:Kanade:Torr:ICLR:2021} study the tolerable amount of label noise as a function of the dataset density.
Such works are very valuable as adversarial robustness is indeed a result of compounding properties.
Yet, optimizing for multiple properties simultaneously is not always possible.
A productive direction of future work could be to investigate correlations between different data properties, e.g.,
the effects of feature dimensionality reduction approaches on class density and separation.

\vspace{0.02in}
\noindent
{\bf Additional Data Properties.}
Existing research on the effects of data on
\emph{standard generalization}~\cite{Lopez:Fernandez:Garcia:Palade:Herrera:InfSci:2013,
Lorena:Garcia:Lehmann:Souto:Ho:CSUR:2020,
Santos:Henriques:Pedro:Japkowicz:Fernandez:Soares:Wilk:Santos:AIR:2022} identified several data properties
not discussed in the papers related to robust generalization that we reviewed.
These include the presence of (i) outliers, i.e., samples that drastically differ from most observed samples in a dataset,
(ii) overlapping samples, i.e., different samples of the dataset having the same feature representation, and
(iii) small disjuncts, i.e., training samples from the same class forming small disjoint clusters dispersed throughout the input space
(more details are in Section~\ref{sec:relatedwork}).
Investigating the effect of such data properties on the model's adversarial robustness could be yet another direction
for possible future work.

\vspace{0.02in}
\noindent
{\bf Quantitative Measure.}
Literature shows that the lower/upper bound of adversarial robustness can be determined by the properties of the underlying data~\cite{Bhagoji:Cullina:Mittal:NeurIPS:2019,Mahloujifar:Zhang:Mahmoody:Evans:NeurIPS:2019}.
Modifying certain properties of the data can also change the robustness of the resulting classifier.
Hence, the ability to quantitatively measure such data properties is very valuable.
However, some data properties discussed in this survey, such as, type of distributions and label noise, lack any reliable estimation techniques.
Current work mostly relies on informal comparative analysis, e.g., that the \mnist dataset is closer to a Bernoulli mixture data than a Gaussian mixture because the pixels are concentrated towards black or white.
Quantitatively measuring the degree of similarity between distributions, although difficult, may be necessary in order to make more accurate conclusions.

Interestingly, other data properties have multiple, often inconsistent, measurement techniques, e.g., concentration~\cite{Mahloujifar:Zhang:Mahmoody:Evans:NeurIPS:2019,Prescott:Zhang:Evans:ICLR:2021,Zhang:Evans:ICLR:2022}, density~\cite{Song:Kim:Nowozin:Ermon:Kushman:ICLR:2017,Pang:Xu:Dong:Du:Chen:Zhu:ICLR:2020},
intrinsic dimensionality~\cite{Amsaleg:Bailey:Barbe:Erfani:Furon:Houle:Radovanovic:Nguyen:TIFS:2021,Lorena:Garcia:Lehmann:Souto:Ho:CSUR:2020}, and inter-class distances~\cite{Ding:Lui:Jin:Wang:Huang:ICLR:2019, Bhagoji:Cullina:Mittal:NeurIPS:2019, Pydi:Jog:ICML:2020}.
For example, the inter-class distances can be calculated as the total distance required to move the samples from one class to
another~\cite{Bhagoji:Cullina:Mittal:NeurIPS:2019, Pydi:Jog:ICML:2020}.
It can also be calculated as the pairwise distances between a pre-defined portion of samples from different classes,
e.g., 10\% from each class, that are the closest to each other~\cite{Ding:Lui:Jin:Wang:Huang:ICLR:2019}.
While the inter-class distance derived through the first approach is more computationally expensive, the second approach
is more susceptible to outliers as it relies only on a subset of samples close to each other.
Moreover, these metrics might not necessarily correlate with each other.
We believe future research can provide more insights about appropriate application scenarios for each of the proposed metrics.

\vspace{-0.1in}
\subsection{Summary}
Practical methods for augmenting data and learning procedures to improve robustness include 
increasing dataset size and diversity through the selection of natural or generated samples. 
One can also change the underlying data representation to optimize for certain properties, 
such as reduced dimensionality and better separation, or to project data into a more desirable distribution. 
This can be achieved through statistical and learning-based techniques. 
Selecting appropriate models, as well as configuring models to consider the underlying data properties,
is also shown to be useful. 
Finally, techniques to estimate achievable robustness can be used to gauge 
how much the robustness observed for a particular dataset can be improved.

As for future research directions, assessing the effects of data on robustness in domains beyond images, such as structured text, software, and more, could be valuable. 
Moreover, investigating ways to assess, evaluate, and utilize properties of realistic datasets, 
independently and in conjunction with commonly used types of models, could have large practical significance.
There is also a lack of work on the interplay of different data properties: 
it may be infeasible to modify a data-property without affecting others.
Studying additional properties of data that were shown to benefit standard generalization, 
such as overlapping and outlier samples, could also help better understand both robust generalization and the accuracy-robustness trade-off.
Finally, the lack of consistent metrics for some data properties may also make empirical measurements less reliable. 

\section{Related Work}
\label{sec:relatedwork}

To the best of our knowledge, our survey is the first to explicitly focus on properties of training data 
in the context of model robustness under evasion attacks. 
We review other surveys that focus on evasion attacks more broadly, while including some discussion 
on data, in Section~\ref{sec:relatedwork-surveys-data}.
Section~\ref{sec:relatedwork-standard} reviews literature that examines how data properties affect standard generalization. 
Finally, in Section~\ref{sec:relatedwork-attacks}, we review techniques related to evasion attacks and defenses.
Additional work that studies non-data-related reasons for evasion attacks, 
as well as non-evasion attacks, such as poisoning and backdoor, is discussed in our online appendix~\cite{appendix}.
 
\vspace{-0.1in}
\subsection{Surveys on Evasion Attacks that Discuss Data}
\label{sec:relatedwork-surveys-data}
Numerous existing surveys review the literature on evasion attacks.
Most of these works do not focus specifically on properties of data but discuss attack and defense mechanisms, non-data-related reasons for adversarial vulnerability, 
and the different threat models. 
Only a few of these works mention data-related reasons for the existence of adversarial examples~\cite{Serban:Poll:Visser:CSUR:2020, Machado:Silva:Goldschmidt:CSUR:2021, Akhtar:Mian:Kardan:Shah:IEEEAccess:2021, Akhtar:Mian:IEEEAccess:2018}.
Specifically, Serban et al.~\cite{Serban:Poll:Visser:CSUR:2020} observe that adversarial vulnerability can be caused by an insufficient training sample size 
and high data dimensionality. 
Similarly, Machado et al.~\cite{Machado:Silva:Goldschmidt:CSUR:2021} mention that the lack of sufficient training data, high dimensionality, 
and high concentration contribute to adversarial vulnerability.
Akhtar et al.~\cite{Akhtar:Mian:IEEEAccess:2018, Akhtar:Mian:Kardan:Shah:IEEEAccess:2021} also mention high dimensionality, along with other non-data-related reasons, 
as a source of adversarial examples.

A concurrent work by Han et al.~\cite{Han:Lin:Shen:Wang:Guan:CSUR:2023}
studies the origins of adversarial vulnerability in deep learning w.r.t. the model, data, and other perspectives.
The authors mention high dimensionality, distributions with high concentration, a small number of output classes, data imbalance, and the perceptual difference in image frequencies as potential sources of adversarial examples.
However, as (a) the focus of that survey is not on data-related properties in particular, 
(b) its paper search was conducted in 2021, and 
(c) it focuses on deep learning models only, 
our work was able to identify more than 50 additional relevant papers which focus on other types of models, 
e.g., non-parametric and linear classifiers, 
and/or discuss additional types of data-related properties, 
such as, types of distribution, class density, separation, and label quality.

In summary, by explicitly focusing on the effects of data properties on evasion attacks in our survey and 
including more than 50 papers not covered in prior work, we were able to 
identify additional relevant properties, practical suggestions, and future research directions in this area.
 
\vspace{-0.1in}
\subsection{Effects of Training Data on Standard Generalization}
\label{sec:relatedwork-standard}
A number of surveys investigate the influence of data properties on standard
rather than robust generalization.
One of the earliest is probably the work of Raudys and Jain~\cite{Raudys:Jain:TPAMI:1991},
who review studies related to the influence of sample size on binary classifiers, showing that
a limited sample size usually leads to sub-optimal generalization.
Bansal et al.~\cite{Bansal:Sharma:Kathuria:CSUR:2021} and
Bayer et al.~\cite{Bayer:Kaufhold:Reuter:CSUR:2022} also survey papers addressing the data scarcity problem.
Their results show that augmentation techniques
can help improve a model's generalization by reducing the problem of model overfitting.

Label noise is another aspect of data that influences both standard and robust generalization.
Most works on this topic find that noisy labels increase the need for a greater number of training samples and may result in unnecessarily complex decision boundaries~\cite{Frenay:Verleysen:TNNLS:2014,Song:Kim:Park:Shin:Lee:TNNLS:2022}. 
These works also show that adversarial training can improve model clean accuracy in the presence of 
label noise.

Lorena et al.~\cite{Lorena:Garcia:Lehmann:Souto:Ho:CSUR:2020} identify 26 quantitative metrics that can be used to 
estimate the difficulty of performing classification on a given dataset.
While some of these metrics, such as high dimensionality and low class separation, 
were also studied in the context of robust generalization, 
other metrics, such as ambiguity of classes and complexity of their separation boundaries, 
were not yet explored, according to our survey.

A number of authors~\cite{He:Garcia:TKDE:2009,Lopez:Fernandez:Garcia:Palade:Herrera:InfSci:2013,Santos:Henriques:Pedro:Japkowicz:Fernandez:Soares:Wilk:Santos:AIR:2022,Yang:Jiang:Song:Guo:IJCV:2022} focus on the imbalance learning problem. 
They show that several data properties, such as low density, data noise, 
data shifts, and~-- most importantly, data overlap between classes~--
further complicated learning from imbalanced data. 
Similarly, Moreno-Torres et al.~\cite{MorenoTorres:Raeder:Rodrigues:Chawla:Herrera:PR:2012} study 
data shift, showing that it negatively affects clean accuracy. 

The aforementioned works show that 
some of the properties discussed in our survey, such as 
the number of samples, dimensionality, density, and label quality, also affect clean accuracy. 
There are also additional data properties that are covered exclusively by these or by our work, e.g., data shifts and data distributions, respectively. 
Studying these additional properties, as well as the interplay between data properties for clean and robust accuracy, 
is an interesting research direction, which could be facilitated by our work.

\vspace{-0.05in}
\subsection{Evasion Attacks and Defenses}
\label{sec:relatedwork-attacks}
A number of works focus on techniques for generating evasion attacks, countermeasures against these attacks, 
and defining the notion of the attack itself.

\vspace{0.02in}
\noindent
{\bf Attacks and Defense.}
Several works~\cite{Maiorca:Biggio:Giorgio:CSUR:2019,
Zhang:Sheng:Alhazmi:Li:ACMTIST:2020,
Demetrio:Coull:Biggio:Lagorio:Armando:Roli:ACMTPS:2021,
Rosenberg:Shabtai:Elovici:Rokach:CSUR:2021,
Li:Li:Ye:Xu:CSUR:2021,
Liu:Tantithamthavorn:Li:Liu:CSUR:2022,
Liu:Nogueria:Fernandes:Kantarci:IEEECST:2022,
Sun:Dou:Yang:Zhang:Wang:Philip:He:Li:TKDE:2022} 
survey adversarial attacks and defenses, observing that 
new attacks constantly bypass defenses, which gives rise to new defenses being proposed, only to be broken again 
(a.k.a. the `cat and mouse race' or the `arms race'). 
They also observe that research in this field studies attacks / defenses at a feature-level, which restricts 
the practicality of the developed techniques by the feasibility of perturbing the corresponding features in real life. 

More recently, researchers have started to investigate the 
susceptibility of newer models to adversarial evasion attacks. 
For example, several studies~\cite{Wang:Pan:Hu:Duan:Pan:IJSWIS:2022, 
Shi:Han:Tan:Kuang:NeurIPS:2022,Yin:Lin:Sun:Wei:Chen:TIFS:2023,Wang:Xie:Microsoft:ChatGPT:ArXiv:2023} 
propose attack techniques against contemporary models, 
such as Graph Neural Networks, Generative Pre-trained Transformers (GPT), and Vision Transformers. 
These studies showed that adversarial examples persist even in newer models, some of which are 
trained with large volumes of data. 
As all these works focus on attack and defense mechanisms rather than 
the effects of data on adversarial robustness, our work extends and complements this research.

\vspace{0.02in}
\noindent
{\bf Adversarial Examples.}
Adversarial examples are generally defined as inputs constructed by perturbing a correctly classified sample in a way that makes the change imperceptible to a human. 
However, as `imperceptible to a human' is hard to define, existing research on adversarial examples approximates imperceptibility with a small perturbation measured through $L_p$ norms.
A line of research~\cite{Gilmer:Adams:Goodfellow:Anderson:Dahl:ArXiv:2018,Sharif:Bauer:Reiter:CVPRW:2018,Fezza:Bakhti:Hamidouche:Deforges:QoMEX:2019, Mezher:Deng:Karam:EUVIP:2022} 
investigates the validity of this assumption. 
This work shows that perturbations generated by $L_p$ norms do not entirely align with human perceptions, 
i.e., some changes with a small $L_p$ norm can be apparent to humans. 
In addition, adversarial examples with the minimum $L_p$ perturbation may be less effective and transferable than 
higher perturbation~\cite{Biggio:Roli:PR:2018,Rosenberg:Shabtai:Elovici:Rokach:CSUR:2021}. 
Hence, a number of approaches explore metrics for imperceptibility 
in computer vision and NLP domains~\cite{Fezza:Bakhti:Hamidouche:Deforges:QoMEX:2019,Mezher:Deng:Karam:EUVIP:2022, Zhang:Sheng:Alhazmi:Li:ACMTIST:2020}. 
Yet another issue with $L_p$ norms is that they cannot be used reliably in domains other than images. 
For example, in the case of software/malware, simply generating adversarial examples with $L_p$ norms 
may result in feature representations that are not possible in 
the problem space~\cite{Rosenberg:Shabtai:Elovici:Rokach:CSUR:2021,Pierazzi:Pendlebury:Cortellazz:Cavallaro:2020}. 
While all these works focus on the properties of adversarial examples, 
they are orthogonal to the topic of our survey, as we rather focus on how properties of the training data 
affect the success of adversarial examples.

\section{Conclusion}
\label{sec:conclusion}
In this survey, we systematically collected, analyzed, and described papers that discuss how data properties affect adversarial robustness in machine learning models.
By analyzing 77 research papers from top scientific venues in Machine Learning, Computer Vision, Computational Linguistics, and Security,
we identified seven domain-agnostic data properties and one image-specific data property that are correlated with adversarial robustness. 

While several of the guidelines for constructing high-quality data that we identified
are similar to those recommended for training accurate models,
producing robust models is more sensitive to the characteristics of the data and requires more effort,
e.g., a larger number of samples, better label qualities, etc. There are also additional data properties important for building
robust models that are not extensively discussed in non-adversarial settings, e.g., concentration of measure.
In a sense, robust generalization is a stronger form of standard generalization.

We identified possible next steps towards improving the understanding of how the data affects a model's adversarial robustness.
These include
studying interactions between different properties of data,
considering the effect of additional properties that improve standard generalization on robust model generalization,
devising quantitative metrics for different aspects of the data, and
extending the studies and their empirical evaluation beyond the images domain.
We hope our survey will help researchers and ML practitioners to better understand adversarial vulnerability
and will spark further research to address the identified knowledge gaps.

\bibliographystyle{ACM-Reference-Format}
\bibliography{00}

\end{document}